\documentclass[journal]{IEEEtran}

\usepackage{cite}
\usepackage[T1]{fontenc}

\usepackage{float}
\usepackage{graphicx}
\usepackage{subfig}
\usepackage{overpic}

\usepackage{amsthm,amsmath,amssymb}
\usepackage{bm}

\usepackage[ruled,vlined,linesnumbered]{algorithm2e}
\SetAlgorithmName{Alg.}{}{}
\usepackage{array}
\usepackage{booktabs}
\usepackage{enumerate}

\usepackage{url}
\usepackage{color}

\begin{document}
%
\title{Parallel Self-assembly for a Multi-USV System on Water Surface with Obstacles}
%
%
%
\author{Lianxin~Zhang, Yihan~Huang, Zhongzhong Cao, Yang Jiao,
        and~Huihuan~Qian 
\thanks{This paper is partially supported by Project 2022A1515240063 from Guangdong Basic and Applied Basic Research Foundation, University Stability Support Program from Shenzhen Science and Technology Innovation Commission, and Project AC01202101105 from Shenzhen Institute of Artificial Intelligence and Robotics for Society, China.}
\thanks{Lianxin Zhang and Huihuan Qian are with Shenzhen Institute of Artificial Intelligence and Robotics for Society, The Chinese University of Hong Kong, Shenzhen, Shenzhen, Guangdong, 518129, China. }
\thanks{Lianxin Zhang, Yihan Huang, Zhongzhong Cao, Yang Jiao, and Huihuan Qian are also with School of Science and Engineering, The Chinese University of Hong Kong, Shenzhen, Shenzhen, Guangdong, 518172, China.}
\thanks{\emph{*Corresponding author is Huihuan Qian (e-mail: hhqian@cuhk.edu.cn).}}
}


\maketitle

\begin{abstract}
Parallel self-assembly is an efficient approach to accelerate the assembly process for modular robots. 
However, these approaches cannot accommodate complicated environments with obstacles, which restricts their applications.
We in previous work consider the surrounding stationary obstacles and propose a parallel self-assembly planning algorithm.
With this algorithm, modular robots can avoid immovable obstacles when performing docking actions, which adapts the parallel self-assembly process to complex scenes.
The algorithm was simulated in 25 distinct maps with different obstacle configurations and shows a significantly higher success rate, which is more than $80\%$, compared to the existing parallel self-assembly algorithms.
For verification in real-world applications, we in this paper develop a multi-agent hardware testbed system. The algorithm is successfully deployed on four omnidirectional unmanned surface vehicles, CuBoats. The navigation strategy that translates the high-level discrete plan to the continuous controller on the CuBoats is presented.
The algorithm's feasibility and flexibility were demonstrated through successful self-assembly experiments on 5 maps with varying obstacle configurations.

\emph{Note to Practitioners-}This paper addresses deploying of self-assembly technologies for modular robots in practical environments with obstacles to facilitate overwater construction tasks or collective transportation systems.
Stationary obstacles may severely influence the assembly planning and robot routing processes.
Moreover, efficient task coordination, robot navigation, and structure formation are required for large-scale assembly tasks.
The algorithm in this work allows all participating robots to navigate online and connect simultaneously to promote efficiency.
The strategy presented here endows the robots' assembly with obstacle-avoidance capability in dense environments.
This work will interest those pursuing efficient assembly in scenes with surrounding obstacles. 
Our hardware experiments demonstrate a concept system and verify the real-time performance of the algorithm under limited computing power.
The approach introduced here is not applicable to robots with heterogeneous shapes, three-dimensional target structures, or overcrowded environments with too many obstacles. 
\end{abstract}

\begin{IEEEkeywords}
	Self-assembly planning, unmanned surface vehicle, autonomous docking, collision avoidance
\end{IEEEkeywords}

%
\IEEEpeerreviewmaketitle

\section{Introduction}
\label{sect:intro}

\begin{figure}
	\centering
	\includegraphics[width=0.9\linewidth]{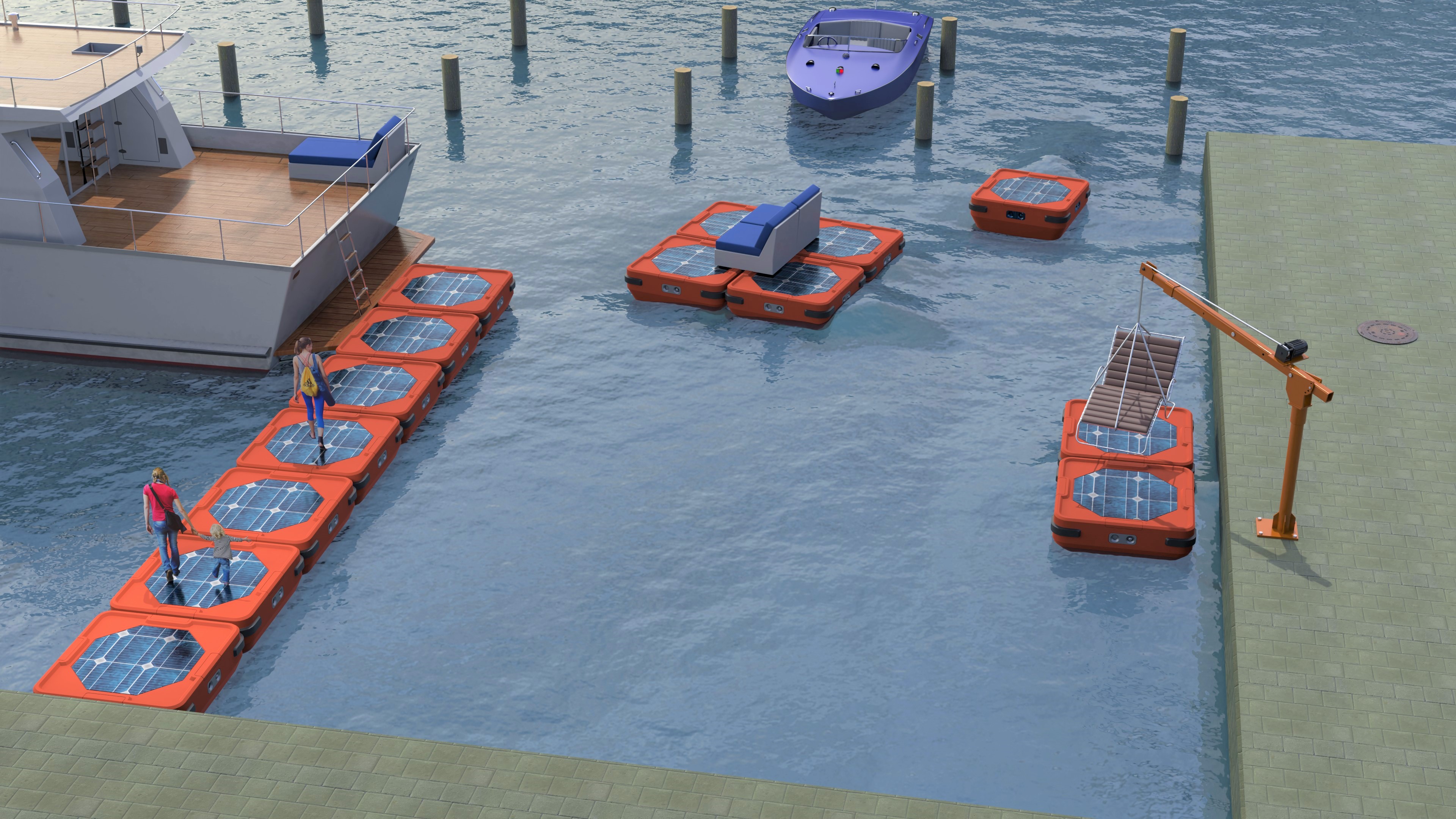} 
	\caption{A fleet of modular USVs assembles on water surfaces with obstacles. The USVs can construct a floating bridge to connect a yacht with the shore or form a large platform to transport large and various-sized cargo.}
	\label{fig:design_rendering}
\end{figure}

Modular self-assembly unveils a promising prospect for collaborative tasks of many robots, as it endows modular robots with reconfigurability and adaptability \cite{jin2010morphogenetic} by scaling \cite{petersen2019review} and shapeshifting \cite{gheneti2019trajectory}. 
As portrayed in Fig. \ref{fig:design_rendering}, the self-assembly technique enhances swarm robots' adaptability to the environment and capability of accomplishing complex missions that overwhelm one single robot, such as collaboration of underwater robots \cite{ganesan2016stochastic}, modular quadrotors \cite{gandhi2020self}, modular self-reconfigurable robots (MSRRs) \cite{daudelin2018integrated, liang2020freebot, ozkan2021self} and unmanned surface vehicles (USVs) \cite{wang2020distributed}, and thus motivates research on robotic self-assembling systems.

Robotic self-assembly planning (SAP) has attracted accelerative attention recently, as it is guaranteed to generate collision-free paths and the assembly sequence without undesired docking.
The SAP problem is similar to the combined target-assignment and path-finding (TAPF) \cite{ma2017overview} in certain aspects,
e.g., the target assignment and path planning before the robot docking.
Nevertheless, collision avoidance is always a must in the TAPF, while robots in the SAP dock together (regarded as collision) and move as a whole at multiple moments during moving.
The SAP problem, generally NP-hard \cite{lv2010assembly}, is intractable for the optimal solution, for which a lot of algorithms have been proposed, and come in two categories: serial and parallel approach. 
In serial self-assembly, the individual or small groups of robots are sequentially connected to the growing structure, e.g., the seed-initiated rule-based methods \cite{baldassarre2007self, rubenstein2014programmable}, the chain forming approach \cite{RN1066}, and the collective robotic construction \cite{werfel2008three}.
However, these methods are not efficient because of the linearly-increasing time step with the number of robots.

Some parallel-docking approaches parallelize the aggregation of multiple robots to one connected component based on the existing serial methods, e.g., growing multiple branches from one seed \cite{liu2023smores} and setting up several seeds \cite{jilek2022self}.
However, these methods are not fully parallel, since the number of branches will not increase with more robots.
A concurrent assembly process is proposed in \cite{klavins2006grammatical} where multiple assembly rule applications simultaneously occur. 
Similarly, a centroidal Voronoi tessellation-based algorithm is presented in \cite{wei2017centroidal} where robots move synchronously by ignoring the collisions and obstacles.
Further, to address challenges including collision avoidance and undesired attachments, reference \cite{saldana2017decentralized} proposes a decentralized fully-parallel self-assembly algorithm with a binary assembly tree, named PAA.
Nevertheless, studies on obstacle avoidance in parallel approaches are still deficient.

Therefore, we in \cite{zhang2021efficient} presented a parallel SAP algorithm with avoidance of immovable obstacles for modular square robots, based on the PAA algorithm.
The novelty of our algorithm, as seen through the comparison with existing algorithms in Table \ref{Table:SAP_alg}, primarily lies in the improvement of the obstacle avoidance capability of the parallel approaches.
In this paper, we first present the same SAP problem and algorithm in Section II, III and IV as \cite{zhang2021efficient}. We then further discussed its advantages/limitations and extended the study by developing a multi-USV hardware testbed system to simulate real-world applications.
The design motivation originates from the great application potentiality of the self-assembly and robotic construction technologies on water surfaces, for instance, intelligent waterway transportation \cite{park2020social}, rescue USV \cite{paulos2015automated}, and floating cities \cite{UN-Habitat}.
Some projects have been done with modular boat design, e.g., TEMP \cite{seo2016assembly}, Roboats \cite{wang2019roboat}, and Modboat \cite{knizhnik2021docking}.
However, their deficiencies include high manufacturing costs, limitations for massive deployment, and unique configurations adverse to running generalized SAP algorithms.
The main contributions of our previous work are proposing a virtual extension procedure and a $Pair$ module in the PAA-based SAP algorithm and simulating it on 25 distinct maps.
The original contributions of this study are itemized below.
\begin{itemize}
	\item A multi-USV hardware testbed system is developed based on an omnidirectional USV with magnetic docking systems named CuBoat, which can perform autonomous docking. 
	\item For each boat, the navigation strategy that translates the discrete high-level planning outputs to the continuous low-level controller is presented.
	\item Experiments on 5 distinct maps with different obstacle configurations are conducted to reveal the rationality and applications of the whole system.
\end{itemize}

\begin{table}[tbp]\scriptsize
	\centering
	\caption{COMPARISON OF TYPICAL SAP METHODS} \label{Table:SAP_alg}
	\addtolength{\tabcolsep}{0pt}
	\begin{tabular}{lcc p{0.78cm}<{\centering} cc}
		\hline
		Ref. & Parallelism & Architecture & NO. of Robots & Obstacle & Applications \\ \hline
		\cite{RN1066}   & Serial & \textbf{Decentralized} & 49   & $\bullet$  & Hollow Shape  \\
		\cite{werfel2008three} &  Serial   &  \textbf{Decentralized}   &  \textbf{512}   &  $\times$   & 3D Construction    \\
		\cite{liu2023smores}  &  \textbf{Parallel}   &  Centralized   &  7   &   $\times$  & Tree Topologies    \\
		\cite{wei2017centroidal} &  \textbf{Parallel}   &  Centralized   &  41   &  $\times$   &  Legged Robots   \\
		\cite{saldana2017decentralized} &  \textbf{Parallel}   &  \textbf{Decentralized}   &  62   &  $\times$   & Bridges, etc.  \\
		\textbf{\cite{zhang2021efficient}} &  \textbf{Parallel}   &  Centralized   &  16   &  $\bullet$   &  Bridges, etc.  \\ \hline
	\end{tabular}
\end{table}

The paper is structured below. 
Section \ref{sect:problem} and \ref{sect:SAPOAalgorithm} present the concerned self-assembly problem and the detailed steps of the proposed SAP algorithm, respectively.
In Section \ref{sect:evaluation}, a series of simulations are performed on 25 maps with different obstacle configurations to verify their feasibility and performance.
The presented SAP algorithm and simulations are identical to those in \cite{zhang2021efficient}.
To further demonstrate the applications, the SAP algorithm is deployed on a multi-USV testbed system. 
The components and control system of the testbed are presented in Section \ref{sect:deployment}, and in Section \ref{sect:experiment} the result of the validation experiments is discussed.
Finally, Section \ref{sect:conclusion} summarizes the paper.

\begin{figure*} [htb]
	\centering
	\includegraphics[width=1\linewidth]{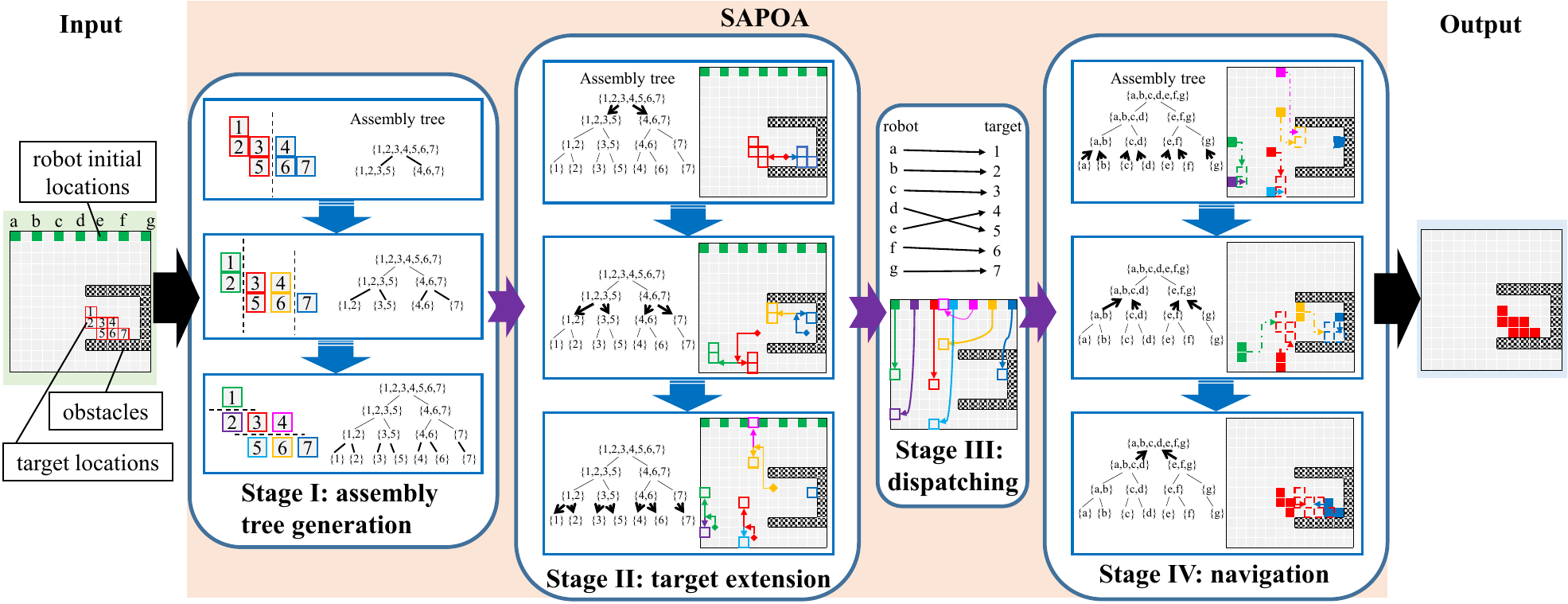}
	\caption{Overview of the SAPOA. A seven-robot self-assembly process is plotted to vividly clarify the four stages. The subplots in panels of stages II and IV depict the recording and tracing processes of the landmark points, respectively.}
	\label{fig:pipeline}
\end{figure*}

\section{Problem Formulation}
\label{sect:problem}

We express the problem in the Euclidean space $\mathbb{R}^2$, with all robots having an identical length $w$.
It is worth noting that the problem definition and algorithm in the following sections are entirely derived from \cite{zhang2021efficient}.

\newtheorem*{preliminariy}{Preliminaries and Notations} \label{thm:preliminariy}
\begin{preliminariy}
	We have $N$ square holonomic robots, each equipped with isomorphic docking systems installed on all four sides.
	Their locations and unspecified targets are represented by sets $\mathbf{A}=\left\{ \boldsymbol{a}_1,\boldsymbol{a}_2,\cdots ,\boldsymbol{a}_N \right\}$, where the location of each robot $i$ is denoted by $\boldsymbol{a}_i=\left[ \begin{matrix} x&		y\\ \end{matrix} \right] ^T\in \mathbb{R}^2$, and $\mathbf{G}=\left\{ \boldsymbol{g}_1,\boldsymbol{g}_2,\cdots ,\boldsymbol{g}_N \right\}$, where the location of each target $j$ is denoted by $ \boldsymbol{g}_j\in \mathbb{R}^2$, respectively.
	The control input $\boldsymbol{v}_i\in \mathbb{R}^2$ of each robot follows first-order kinematics, namely, $\dot{\bm{a}}_i=\bm{v}_i$.
	The assignment from robots to targets is represented by a transformation of set $\mathbf{A}$, namely, $T\left(\mathbf{A}\right)$.
	$\lVert \cdot \rVert_2 $ denotes the Euclidean norm, while $\left| \cdot \right|_c$ calculates the number of connected components of a point set.
	Obstacles are immovable and are represented as $\mathbf{B}=\left\{ \boldsymbol{b}_1,\boldsymbol{b}_2,\cdots ,\boldsymbol{b}_M \right\}$, with each obstacle point $\boldsymbol{b}_k\in \mathbb{R}^2$ having the same size as robots ($w$).
\end{preliminariy}

\newtheorem*{Given}{Given Information} \label{thm:Given}
\begin{Given}
	The initial robot positions ($\mathbf{A}_0$), target positions ($\mathbf{G}$), and the map with obstacles ($\mathbf{B}$) are provided at the outset. At the initial time ($t_0$), the positions of the robots are set to $\mathbf{A}\left( t_0 \right) = \mathbf{A}_0$. An instance is shown in the input panel of Fig. \ref{fig:pipeline}.
\end{Given}

\newtheorem*{constraints}{Constraints} \label{thm:constraints}
\begin{constraints}
		(1) At the end time $t_e$, the constructed structure must be connected, i.e., $\left| \mathbf{A}\left( t_e \right) \right|_c=1$.
		(2) The robots must avoid collisions at all times $t$ during their movement, meaning that $\lVert \bm{a}_i\left( t \right) -\bm{a}_j\left( t \right) \rVert _{2}^{2}\geqslant w,\lVert \bm{a}_i\left( t \right) -\bm{b}_k \rVert _{2}^{2}\geqslant w,\forall i,j\in N,i\ne j,k\in M$.
\end{constraints}

\newtheorem*{assumption}{Assumptions} \label{thm:assumption}
\begin{assumption}
	Considering the challenges of docking on the water surface, we have made the following assumptions.
	\begin{enumerate}
		\item Each robot only moves forward/backward/leftward/ rightward at a uniform speed of 1 robot-length/step, with no rotation, namely, $\boldsymbol{v}_i\in \left\{ \left[ \begin{matrix}
			0&		0\\
			\end{matrix} \right] ^T,\left[ \begin{matrix}
			\pm w&		0\\
			\end{matrix} \right] ^T,\left[ \begin{matrix}
			0&		\pm w\\
			\end{matrix} \right] ^T \right\}$.
		\item Once the robots are adjacent, they will remain connected by the docking process until the end. The equipped docking systems are passive actuators, with examples including magnetic docking mechanisms in \cite{saldana2018modquad, saldana2019design}. \label{ass:connection}
		\item The docking action only occurs sequentially between two groups of robots owing to environmental disturbance.
	\end{enumerate}
\end{assumption}

Our task is to develop an SAP algorithm that leads a swarm of modular robots to reach a set of target locations and form the desired structure while navigating through obstacles. 
The objective of the algorithm is to minimize the total time, namely, the overall moving steps of paths that move robots from their initial positions to the desired formation.
Overall, the SAP problem can be formulated as
\begin{equation} \label{eq:sap_problem}
\begin{aligned} 
&\underset{T\left( \mathbf{A} \right) ,\boldsymbol{v}_i}{\min}\,\,\int_{t_0}^{t_e}{1dt}
\\
\text{s}.\text{t}. \quad & \mathbf{A}\left( t_0 \right)=\mathbf{A}_0, \quad
\left| \mathbf{A}\left( t_e \right) \right|_c=1,
\\
&\sum_{\tilde{\boldsymbol{a}}\in T\left( \mathbf{A} \right)}{\lVert \tilde{\boldsymbol{a}}\left( t_e \right)-\boldsymbol{g}_{\tilde{\boldsymbol{a}}} \rVert _{2}^{2}}=0, \quad
\dot{\boldsymbol{a}}_{i}=\boldsymbol{v}_i, 
\\
&\lVert \boldsymbol{a}_i\left( t \right) -\boldsymbol{a}_j\left( t \right) \rVert _{2}^{2}\geqslant w, \quad
\lVert \boldsymbol{a}_i\left( t \right)-\boldsymbol{b}_{j} \rVert _{2}^{2}\geqslant w,
\\
&\forall i,j\in N,i\ne j,k\in M,
\end{aligned} 
\end{equation}
where $\boldsymbol{g}_{\tilde{\boldsymbol{a}}}\in \mathbf{G}$ is the target location corresponding to the assigned robot $\tilde{\boldsymbol{a}}$.
Without loss of generality, we will analyze this problem in a two-dimensional grid map where each cell has the same dimension as a modular robot, i.e., $w=1$.
To accomplish the objective, we have proposed an SAP algorithm with Obstacle Avoidance, called SAPOA.

\section{Algorithm Statement}
\label{sect:SAPOAalgorithm}

\subsection{The Overview and Terms}
\label{sect:pipeline}

Fig. \ref{fig:pipeline} outlines the proposed SAPOA algorithm \cite{zhang2021efficient}, whose idea stems from the assembly-by-disassembly technology in manufacturing \cite{hoffman1989automated}.
Specifically, all the robots will firstly move to the expanded targets, and then build the desired formation step by step. 
This search-record-reconstruction pattern endows our algorithm with collision avoidance capability.
The algorithm is divided into four stages: assembly tree generation, target extension, dispatching, and navigation. 
Moreover, we have also supplemented this section with an analysis of the algorithm's advantages, limitations, and complexity.

Compared to existing methods, our algorithm has at least the following advantages.
1) The algorithm accommodates the self-assembly process to the surrounding obstacles without prior requirements for their distribution.
2) To minimize the overall moving steps, we employed the following three approaches, namely,
parallelizing the assembly actions, minimizing the overall distances of the robot-target matching during the dispatching stage, and planning the shortest paths by A* algorithm \cite{hart1968formal} for the movements of all robots.
Meanwhile, it is undeniable that the SAPOA has limitations.
1) As discussed in the subsequent section, the algorithm may fail in some circumstances due to its intrinsic randomness.
2) Our scheme can neither guarantee the minimum of the planned overall steps nor quantify the gap with the optimal solution.
To the authors' best knowledge, no published work can guarantee the solution optimality, considering the complexity and variability of the SAP problems.

To facilitate the clarification, some prerequisite terms are defined.
Specifically, the $group$ denotes the attached target locations or robots.
Hereafter, the $pair$ consists of two subgroups to be parted or attached, which are $partner$s mutually.

\subsection{Stage I: Assembly Tree Generation}
\label{sect:assemblytree}

\begin{algorithm} [h]
	\caption{AssemblyTreeGeneration($S$)}
	\label{alg:assemblytree}
	\KwIn{$S$ containing all target locations }
	\KwOut{assembly tree $Tree$}
	
	/*When the group $S$ to be divided contains one point, ends.*/ \\
	\If{$\lvert S \rvert =$ 1}{
		return;
	}
	/*Find all possible partitions by lines.*/\\
	$P \leftarrow$ AllDivisions($S$);\\
	/*Solve Eq. \ref{eq:division} for the most balanced division.*/\\
	$(S_1,S_2) \leftarrow $  BestDivision($P$, $f$);\\
	/*Create new node to save the two $S_1,S_2$.*/\\
	$S.lChild \leftarrow $ NewNode($S_1$);\\
	$S.rChild \leftarrow $ NewNode($S_2$);\\
	AssemblyTreeGeneration($S.lChild$);\\
	AssemblyTreeGeneration($S.rChild$);\\
\end{algorithm}

A recursive algorithm to generate the assembly tree is leveraged here, which is enlightened by \cite{saldana2017decentralized}.
Let $S$ symbolize the set of all the target points.
Alg. \ref{alg:assemblytree} shows that $S$ is firstly divided into two groups $S_i$ and $S_j$ by a straight line horizontally or vertically, and then both $S_i$ and $S_j$ are further split into two.
This process is recursively executed until only one point for each group.
In each recursion, the balanced division is achieved by solving an optimization problem,
\begin{equation} \label{eq:division}
\begin{aligned} 
&\underset{S_i,S_j}{\max}f\left( S_i,S_j \right) 
\\
\text{s}.\text{t}. \quad &\left| S_i \right|+\left| S_j \right|=\left| S \right|,
\end{aligned} 
\end{equation}
where $f\left( S_i,S_j \right)=\left| S_i \right|\left| S_j \right|$ is a factor to evaluate each division, with $\left| \cdot \right|$ counting the number of points in the set. 
This algorithm resembles the Alg. 1 in \cite{saldana2017decentralized}, so the time complexity is $O(m^3 \log m)$, where $m$ is the number of given targets.

\subsection{Stage II: Target Extension}
\label{sect:extension}

Following the top-down level order of the generated assembly tree, as Alg. \ref{alg:extension} describes, the desired structure is expanded in this stage.
A $pair$ with two subgroups is introduced to perform the extension.
The root node is a $pair$ module at this level, which contains two subgroups.
The extension algorithm aims at separating these two subgroups to a user-defined distance, which is at least 4 units. 
The reason is that based on the assumption (\ref{ass:connection}), for robots passing by each other, no less than 1 cell (2 units) needs sparing among them.
At each level, two target groups in a pair are detached in opposite directions as long as not stuck by obstacles or other groups.
Otherwise, they, as a pair, will explore the surrounding until finding space for separation.
Finally, all the pairs are separated with the expanded target points stored as landmarks.

\begin{algorithm} [h]
	\caption{TargetExtension($Tree$, $Map$)}
	\label{alg:extension}
	\KwIn{assembly tree $Tree$, map with obstacles $Map$}
	\KwOut{a set of landmarks $E$}
	
	\ForEach{ \emph{level} l \emph{of} $Tree$\emph{, from root to leaves}}{ 
		
		$TargetPairs \leftarrow \varnothing$; /*To save all target pairs.*/\\
		
		\ForEach{Node \emph{in level} l}{
			$Pair \leftarrow$ NewPair($Node$.lChild, $Node$.rChild); \\
			Save2Pair($Pair$, $TargetPairs$); /*Save new $Pair$ to $TargetPairs$.*/\\
		}
		
		\While{\emph{not all} $TargetPairs$ \emph{separated}}{
			\ForEach{$Pair \in TargetPairs$} {
				\If{distance of $Pair$'s subgroups $<$ 4}{
					$Pair$.Separation($Map$); 	\label{line:separation}\\
				}
			}
			/*Explore when $Separation$ is stuck.*/ \\
			\If{not all $TargetPairs$ \emph{separated}}{
				\ForEach{$Pair \in TargetPairs$} {
					$Pair$.Exploration($Map$); \label{line:exploration}\\
				}
			}
			\If{loop infinite}{
				return $Fail$; \\
			}
		}
		Save2Landmark($Pair$,$E$); /*Save as landmarks.*/
	}
	return $E$;
\end{algorithm}

Two pivotal functions are leveraged to facilitate the extension, i.e., $Separation(\cdot)$ in line \ref{line:separation} and $Exploration(\cdot)$ in line \ref{line:exploration}. The details are explained below.

\begin{itemize}
	\item $Separation(\cdot)$: $Separation(\cdot)$ intends to separate the two subgroups of a target pair in the opposite directions. 
	Without collision, the separation distance between partners or from other target groups is no less than 4 units. 
	If impeded by obstacles or other groups, the separation actions will pause till the next attempt. 
	
	\item $Exploration(\cdot)$: All the target pairs move away from each other, as well as the obstacles, in a randomly-selected and unblocked direction.
	During the exploration, the target pairs always move at least 4 units away from each other, unless the distances in all four directions are greater than 4.
\end{itemize}

Regarding the balanced assembly tree with $m$ leaf nodes, the time complexity for a single traversal is $O(m\log m)$. 
However, the number of iteration in lines 6-15, denoted by $w$, cannot be determined precisely due to the heuristics, which is dependent on both $m$ and the map difficulty.
Consequently, the complexity for Alg. \ref{alg:extension} is $O(mw\log m)$.

\begin{figure} [htbp]  
	\centering
	\subfloat[]{\label{fig:collisionAvoidi}         
		\begin{minipage}[b]{0.3\linewidth}      
			\centering      
			\includegraphics[width=1\linewidth]{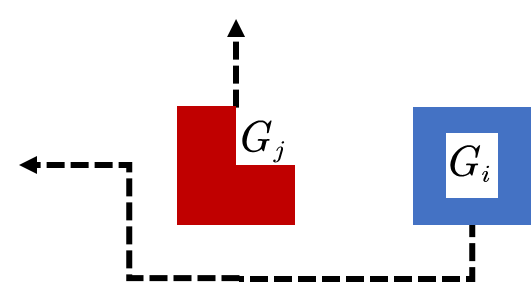}      
		\end{minipage} 
	}
	\subfloat[]{\label{fig:collisionAvoidii}
		\begin{minipage}[b]{0.3\linewidth}      
			\centering      
			\includegraphics[width=1\linewidth]{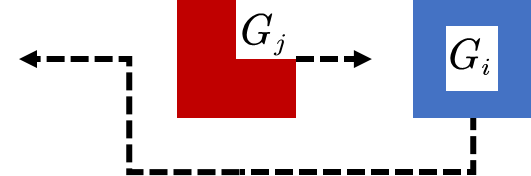}      
		\end{minipage}      
	} 
	\caption{Scenario (i) in panel (a) and (ii) in panel (b) of the rule for collision avoidance.}
	\label{fig:collisionAvoid}
\end{figure}

\subsection{Stage III: Dispatching Robots to the Expanded Targets}
\label{sect:dispatching}

The Hungarian algorithm \cite{kuhn1955hungarian} is used for the allocation of the expanded target points to robots in this stage.
A bipartite graph portrayed in the stage III of Fig. \ref{fig:pipeline} is constituted by the paths from the initial locations of robots to the expanded targets whose lengths are the graph weights.
The bipartite graph can be transformed into an adjacent matrix of which the elements in each row are the lengths of the shortest paths from one robot to all targets computed by A* algorithm.
Inputting this matrix into the Hungarian function yield the assignment vector, of which each component contains the target and each index corresponds to the dispatched robot.
Similar to \cite{paulos2015automated}, the time complexity is $O(m^3)$.
The overall path length of the robot-target coupling is minimized by this dispatching result.

\newcommand{\gridmapwidth}{0.17}
\begin{figure*} [htbp]
	\centering
	\subfloat[]{\label{fig:simuMaps0}         
		\begin{minipage}{\gridmapwidth\linewidth}      
			\centering      
			\includegraphics[width=1\linewidth]{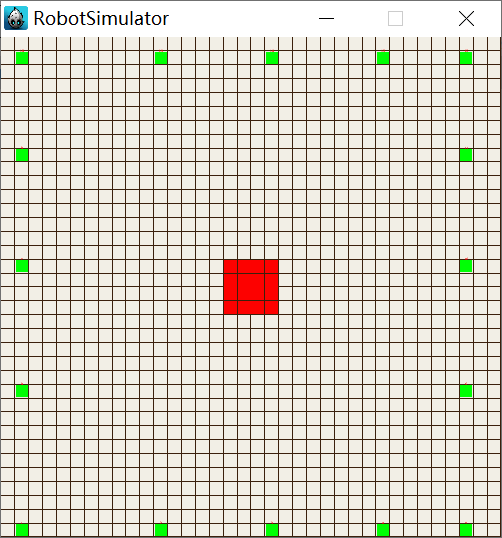}      
		\end{minipage} 
	}\subfloat[]{\label{fig:simuMaps1}
		\begin{minipage}{\gridmapwidth\linewidth}      
			\centering           
			\includegraphics[width=1\linewidth]{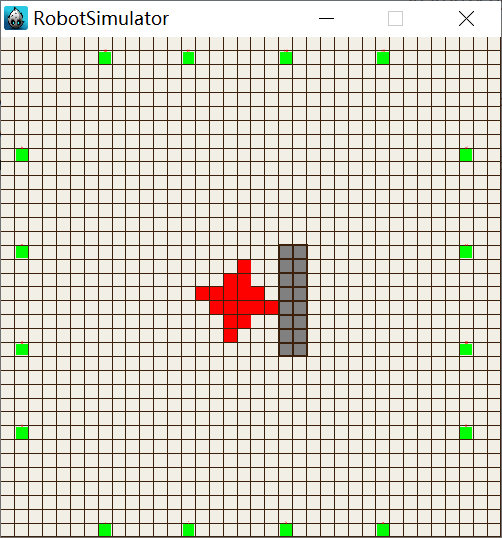}      
		\end{minipage}      
	}\subfloat[]{\label{fig:simuMaps2}
		\begin{minipage}{\gridmapwidth\linewidth}      
			\centering            
			\includegraphics[width=1\linewidth]{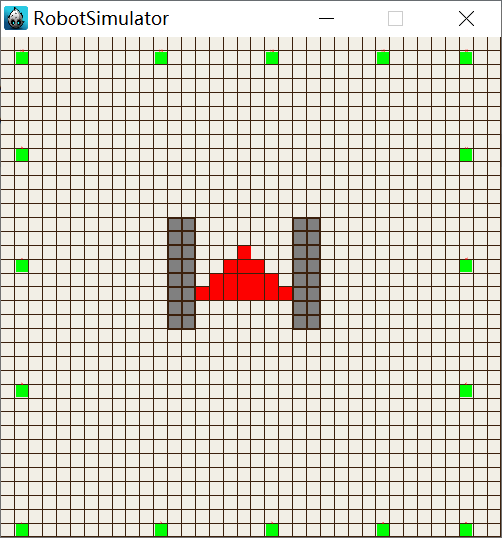}      
		\end{minipage}      
	}\subfloat[]{\label{fig:simuMaps3}
		\begin{minipage}{\gridmapwidth\linewidth}      
			\centering           
			\includegraphics[width=1\linewidth]{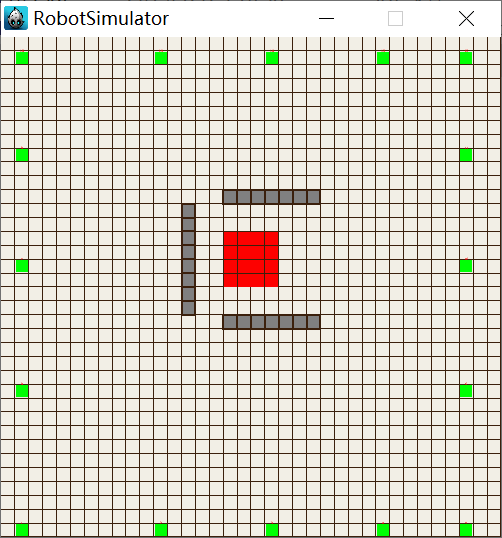}      
		\end{minipage}      
	}\subfloat[]{\label{fig:simuMaps4}
		\begin{minipage}{\gridmapwidth\linewidth}      
			\centering           
			\includegraphics[width=1\linewidth]{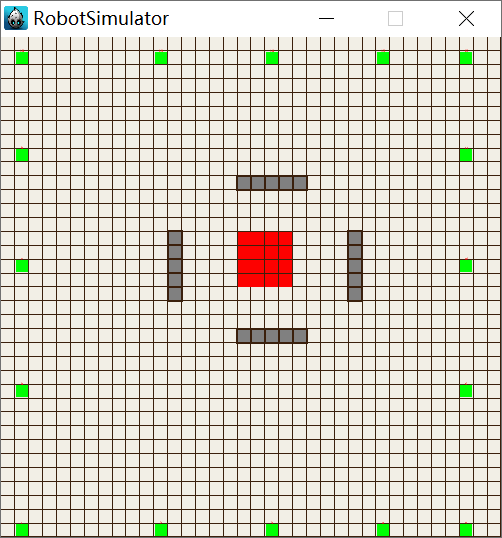}      
		\end{minipage}
	}   
	\caption{Typical examples for the 5 categories of the simulation maps. Cells in green, red, and gray denote robots, targets, and obstacles, respectively. Maps in Cat. 1 have no obstacles, while in Cat. 2, they contain one side of obstacles, and so forth.}
	\label{fig:simuMaps}
\end{figure*}

\subsection{Stage IV: Robot Navigation}
\label{sect:navigation}

Now the robots can plan their trajectories with A* algorithm in $O(v\log v)$ time ($v$ is the number of map grids) and move in a distributed manner. The most time-consuming aspect in this stage is the robot movement, not computation.
In a bottom-up order of the assembly tree, the robots navigate at each level after receiving the landmarks, the obstacles, and the local information.
The algorithm details can be found in \cite{zhang2021efficient}.
During the robot movement, the rules for collision avoidance and docking actions take effect.

\begin{itemize}
	\item \emph{The rule for collision avoidance}: Fig. \ref{fig:collisionAvoid} exhibits two collision scenarios. 
	(i) The group $G_i$ is stopped by another group $G_j$ during movement. 
	Then, $G_i$ will move around $G_j$ by involving it in the path re-planning, while $G_j$ will keep on.
	(ii) The group $G_i$ and $G_j$ mutually block each other with $G_i$ owning lower  predetermined priority. 
	$G_i$ will move around $G_j$ by re-planning the path, while $G_j$ will wait for preset steps before continuation.
	
	\item \emph{The rule for docking actions}: 
	Except docking, a robot group $G_i$ keeps a distance from others by at least 1 cell (i.e., 2 units), even from its partner. 
	Groups will conduct the docking actions as soon as the following two conditions are simultaneously satisfied.
	(i) $G_i$ is only one step left from completely occupying its target locations.
	(ii) Another robot group stays in its partner's target region. 
\end{itemize}

\begin{figure} [htbp] 
	\centering
	\includegraphics[width=0.5\linewidth]{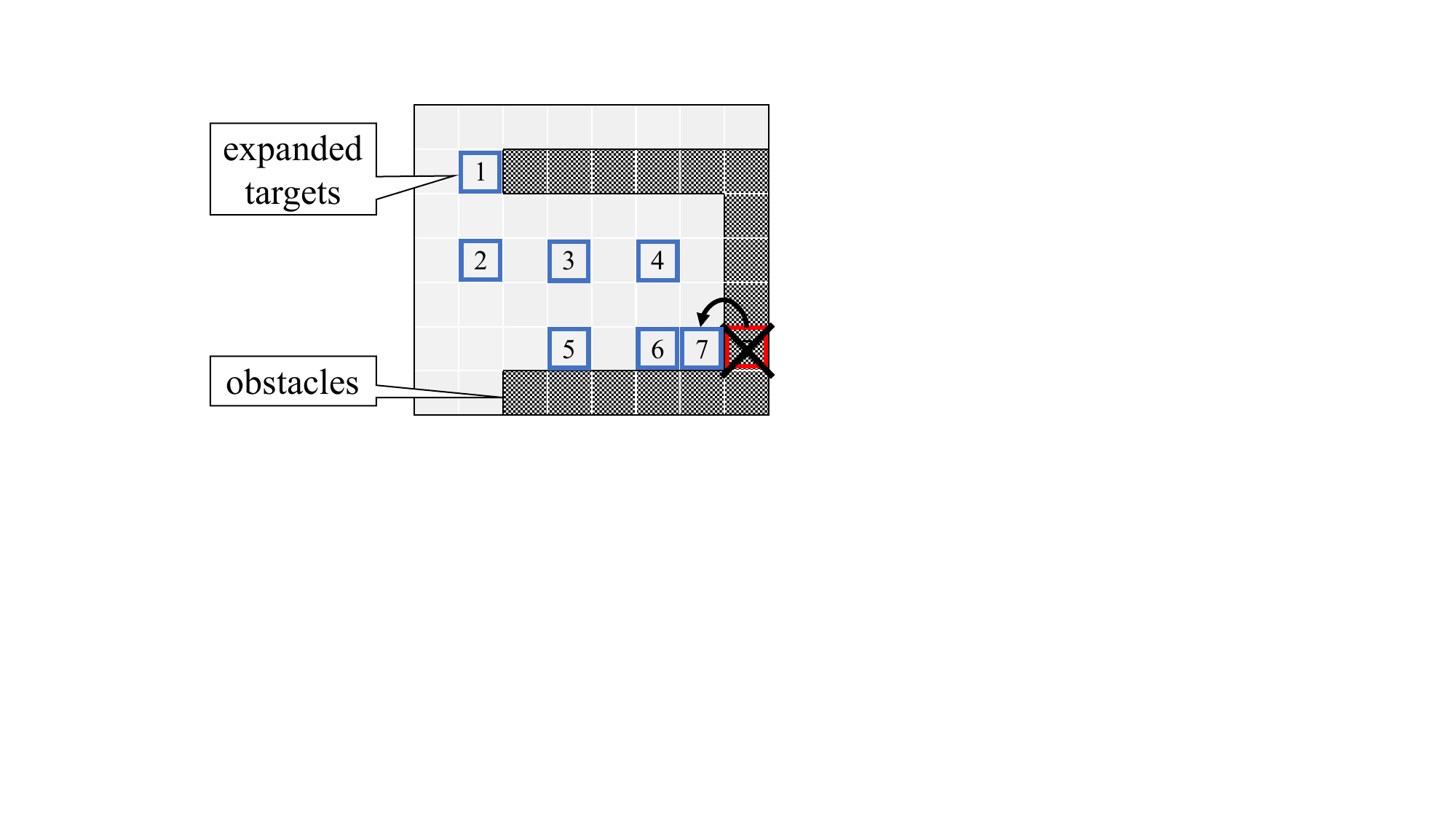}
	\caption{The APAA algorithm adjusts the expanded targets on obstacles. }
	\label{fig:APAA}
\end{figure}

\subsection{Work Example}
\label{sect:example}

A detailed example is presented in Fig. \ref{fig:pipeline}.
In stage I, the target structure undergoes recursive balanced partitions.
For instance, the root $\left\{1,2,3,4,5,6,7\right\}$ is divided into $\left\{1,2,3,5\right\}$ and $\left\{4,6,7\right\}$ until each node contains a single point.
Moving to stage II, the targets are expanded level by level following the assembly tree, as depicted in lines 6-15 of Alg. \ref{alg:extension}. 
For example, the $pair \left\{1,2\right\}-\left\{3,5\right\}$ initially encounters an obstruction in its expansion, prompting an exploration of the map (lines 12-13) to find suitable space. 
Eventually, it expands at the bottom left corner, as demonstrated in the second panel of Stage II.
The expanded target locations are recorded in line 16.
This iterative process continues until all points are separated and adequately spaced apart.
After being dispatched to all targets in stage III, the robots can achieve assembly in stage IV by tracking these recorded landmarks.

\section{Simulation Evaluation}
\label{sect:evaluation}

\subsection{Simulation Configuration} 
\label{sect:simulationConfiguration}

As an open-source project, the algorithm source code in C++ is publicly available at this link: https://github.com/LiamxZhang/Multi-agent-docking.
To verify its generality, we designed 25 distinct maps with a uniform scale of $36 \times 36$, which were evenly classified into 5 categories based on the number of directions with obstacles near the targets.
Typical examples are listed in Fig. \ref{fig:simuMaps}.
For a specific set of target locations, the more directions with obstacles exist, the more challenging it is for the algorithm to succeed.
Therefore, the map difficulty broadly increases from Cat. 1 to 5.
Due to the non-determinacy of target extension in stage II, our algorithm will fail in some cases during the assembly process.
For instance, the target extension will fall in an endless loop or deadlock due to being trapped by obstacles. 
As simulated in \cite{zhang2021efficient}, the algorithm was run 20 times on each map, yielding the success rates and the steps of robot movements.
In the simulation results, we further presented and discussed the extending steps of targets.

For verification, we compare the proposed SAPOA with two types of algorithms.
One is proposed in other works, viz. Naive and APAA, to validate the performance.
The other is the variants of our algorithm, viz. SAPOAnop and SAPOAads, to evaluate its modules.
 
\begin{enumerate}
	
	\item \textbf{Naive algorithm (Naive).} An analogous algorithm is proposed in \cite{wei2017centroidal}. Without target extension process, all the robots are  allocated to the targets by the Hungarian algorithm.
	Then the robots directly aggregate and move to their desired locations based on path planning with local information involved.
	They will assemble as long as next to each other.
	\label{alg:naive}
	
	\item \textbf{Adapted PAA algorithm (APAA).} The original PAA algorithm is proposed in \cite{saldana2017decentralized}, which does not concern the obstacles at the stage of target extension and dispatching.
	A one-step mapping utilizing a linear function is applied to determine the expanded target configuration.
	In the simulation, to adapt the algorithm to the maps with obstacles, we adjust the expanded target points on obstacles to the closest free locations, as depicted in Fig. \ref{fig:APAA}.
	\label{alg:APAA}
	
	\item \textbf{SAPOA without pairs (SAPOAnop).} We dissolve the $pair$ setting in the SAPOA algorithm, which means no $Separation$ action employed in stage II.
	Moreover, during $Exploration$, each robot group moves independently rather than following its partner, as in in Fig. \ref{fig:SAPOAnop}.
	It is used as an ablation test to verify the $Pair$ module.
	\label{alg:SAPOAnop}

	\begin{figure} [htbp]  
		\centering
		\subfloat[]{\label{fig:pair}         
			\begin{minipage}[b]{0.3\linewidth}      
				\centering      
				\includegraphics[width=1\linewidth]{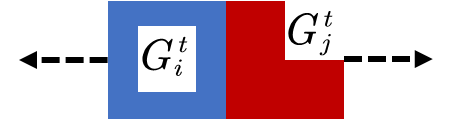}      
			\end{minipage} 
   		}
		\subfloat[]{\label{fig:nopair}
			\begin{minipage}[b]{0.22\linewidth}      
				\centering      
				\includegraphics[width=1\linewidth]{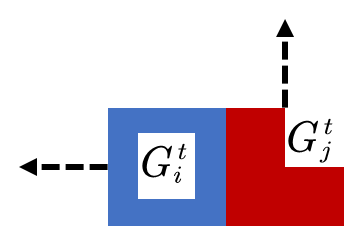}      
			\end{minipage}      
		} 
		\caption{(a) With the $pair$ module, two subgroups $G_i^t$ and $G_j^t$ in a pair separate in the opposite directions. (b) Without the $pair$ module, all groups move individually.}
		\label{fig:SAPOAnop}
	\end{figure}
	
	\item \textbf{SAPOA with active docking systems (SAPOAads).} The proposed SAPOA can also be deployed on the robots with active docking systems.
	These robots with docking flexibility can perform or cease the docking actions at any time \cite{daudelin2018integrated, seo2016assembly}, theoretically yielding a higher success rate.
	Therefore, the robots can stand a closer distance from each other, such as 1 unit.
	However, the active docking systems are usually more complex to implement than the passive ones \cite{murata2002m}, so the latter is still assumed in this paper.
	\label{alg:SAPOAads}
\end{enumerate}

\subsection{Simulation Results} 
\label{sect:simulationResults}

\begin{figure}[htbp]      
	\centering      
	\subfloat[]{\label{fig:successRate}    
		\begin{minipage}{0.9\linewidth}      
			\centering      
			\includegraphics[width=1\linewidth]{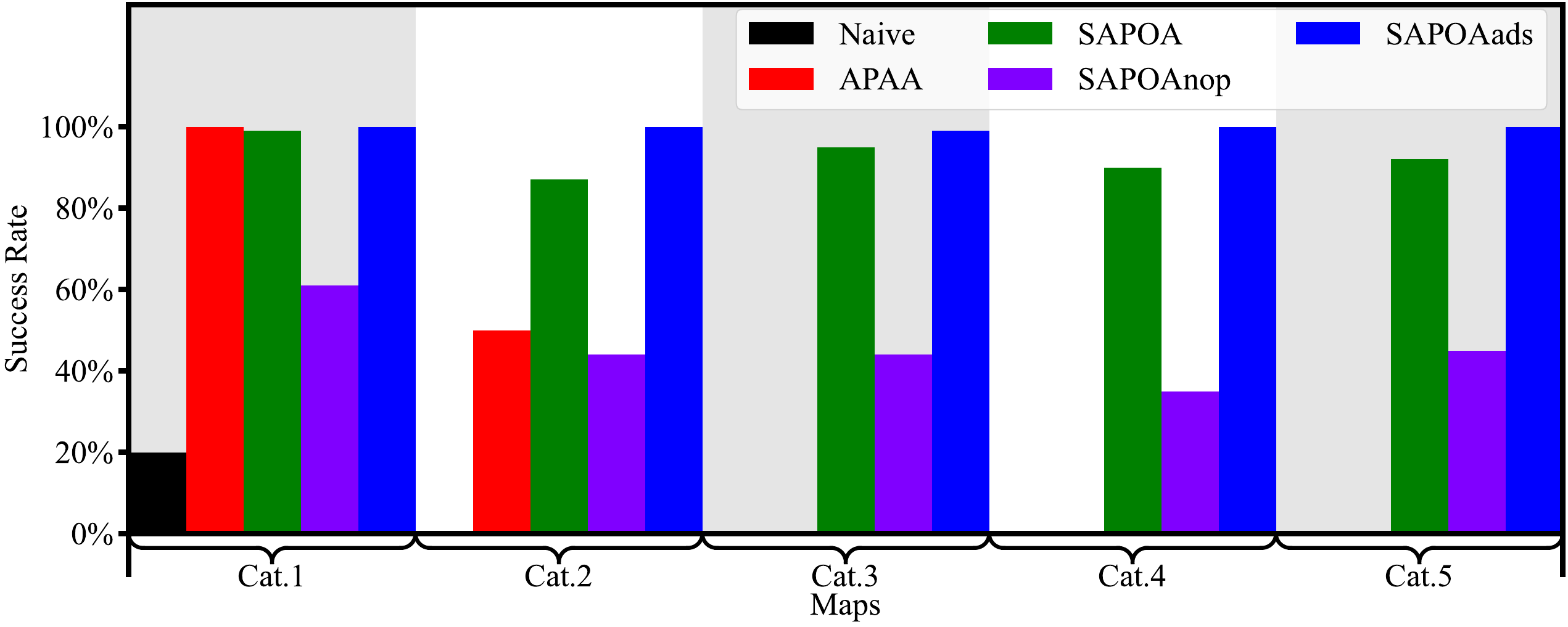}      
		\end{minipage}      
	}

	\subfloat[]{\label{fig:targetStep}       
		\begin{minipage}{0.9\linewidth}      
			\centering      
			\includegraphics[width=1\linewidth]{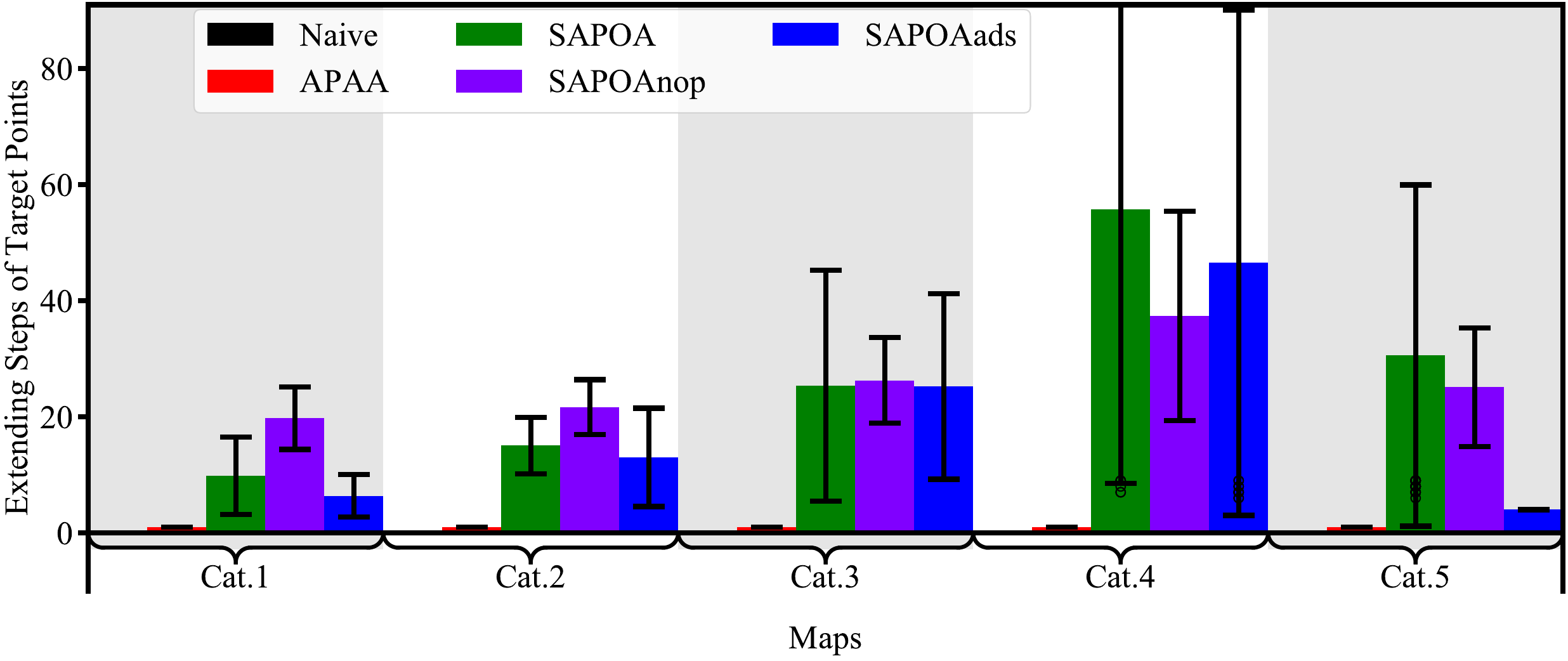}      
		\end{minipage}      
	}

	\subfloat[]{\label{fig:robotStep}       
		\begin{minipage}{0.9\linewidth}      
			\centering      
			\includegraphics[width=1\linewidth]{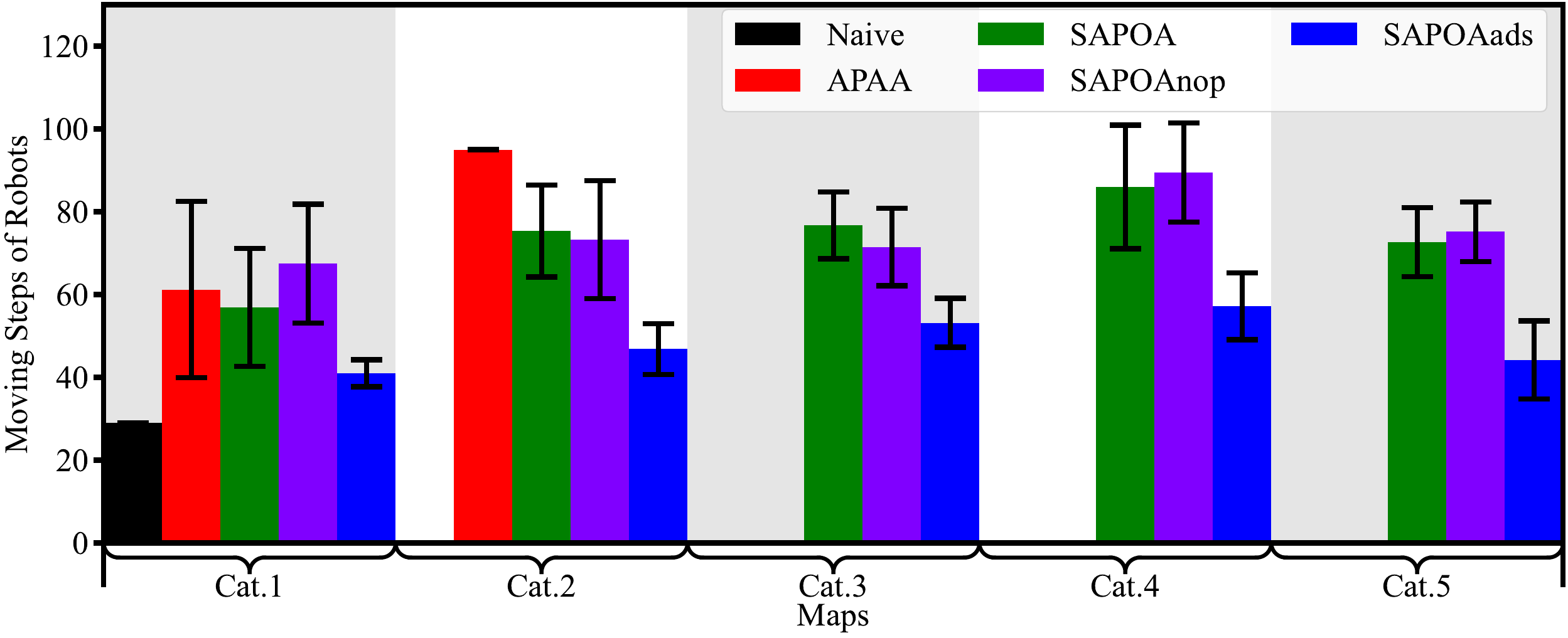}      
		\end{minipage}      
	}      
	\caption{All the SAP algorithms are simulated in the 25 maps of 5 categories for comparison. The statistical averages of (a) the success rates, (b) the target extending steps and (c) the robot moving steps are displayed by category. Here the steps of the failed simulations are not counted.}
	\label{fig:simulationResults}  
\end{figure}

Fig. \ref{fig:successRate} shows a downtrend of success rates for all algorithms with the difficulty of the maps.
Three points can be concluded by contrast.
(1) It can be found that the SAPOA and SAPOAads have the highest success rates which are more than $80\%$.
As a comparison, the naive algorithm merely works in the simplest map and the APAA can only succeed in maps with few obstacles.
(2) One observation is that the success rates improve significantly from the SAPOAnop to the SAPOA,
(3) Unsurprisingly, by replacing the passive docking systems with active ones, the SAPOAads realizes high success rates (nearly $100\%$) in all maps.
Hence, it is meaningful to design active docking systems for the self-assembly of modular robots.
The results reveal a limitation of our algorithm that, despite its impressive success rate, a solitary run is insufficient to ensure success when applied in an unfamiliar environment. Multiple runs are necessary for reliability.

Fig. \ref{fig:targetStep} depicts the extending steps of target points for the successful simulations. 
Given that only SAPOA, SAPOAnop, and SAPOAads algorithms include an extension process, we will exclusively analyze these three.
First, as map difficulty rises, all three algorithms exhibit an expected increase in extending steps.
Second, SAPOAnop shows fewer extending steps in complex maps compared to SAPOA, potentially leading to its lower success rate.
Last, among the three, SAPOAads demonstrates superior performance with the fewest extending steps.

Fig. \ref{fig:robotStep} illustrates the moving steps of the robots simply counted for the successful simulations of each map.
We can see that only a few points of the Naive and the APAA are plotted since these two algorithms fail in most maps.
Regarding their success rate is at most $20\%$ on maps with obstacles, we do not compare them.
Besides, two aspects can be drawn from the comparison result.
(1) We can observe an increasing macro trend of time step in the SAPOA and the SAPOAnop, which reveals that the map complexity gradually rises.
Also, slightly fewer time steps can be noticed for the SAPOA to move the robots in most maps, which means the SAPOA achieves higher success rates more efficiently.
(2) Without exception, the SAPOAads surpasses other algorithms with the minimum time steps, which is average less than 60 in each map category. Therefore, it is a promising direction to design active docking systems for self-assembly.

\begin{figure} [htbp] 
	\centering
	\includegraphics[width=0.8\linewidth]{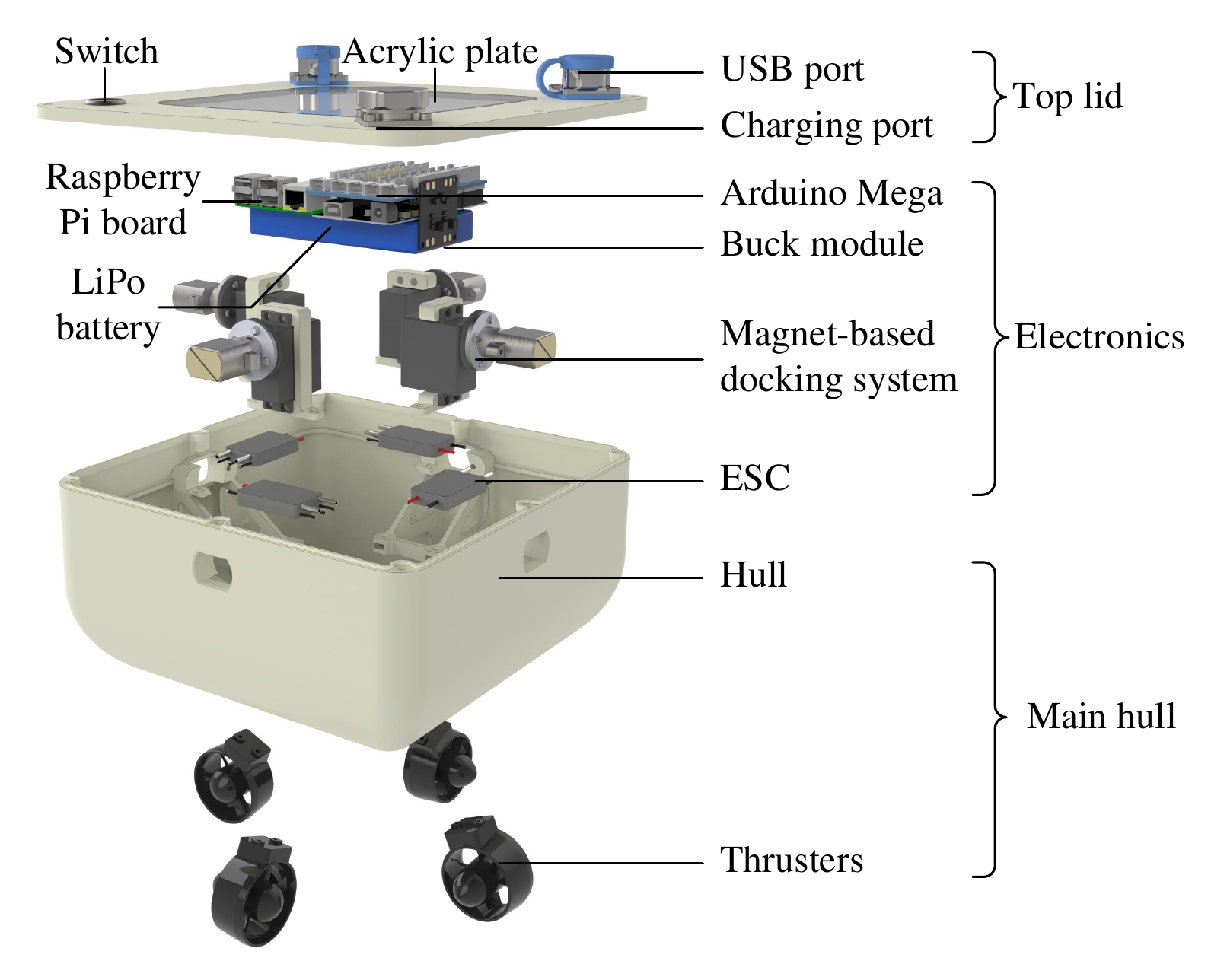}
	\caption{System overview of the CuBoat. }
	\label{fig:cuboat}
\end{figure}

\section{Multi-USV System Design and Control}
\label{sect:deployment}

As the SAPOA algorithm design was established and its effectiveness and efficiency were evaluated, we approached the experimental tests to validate the feasibility. 
In light of the application prospect of the self-assembly on water surfaces, although with various uncertainties (e.g., inaccurate position control), we developed a multi-USV testbed system for the algorithm deployment.
Notably, this testbed system is one of the novel contributions that extend our prior research.
This section describes the main components of the multi-USV system which incorporates four omnidirectional USVs, named CuBoats, autonomously performing the self-assembly tasks in different maps, and the control method for each CuBoat.

\subsection{USV Testbed} 
\label{sect:USV}
Our testbed system consists of four CuBoats, an indoor pool with water dimension of $ 6\ {\rm m} \times 6\ {\rm m} \times 0.4\ {\rm m}$, and an OptiTrack motion capture system for indoor localization.
Fig. \ref{fig:cuboat} exhibits the main components of the CuBoat design, including the hull structure, the electronics, the propulsion system, and the docking system, each of which is detailed described in this section.

\subsubsection{Hull and Propulsion System} 

\begin{figure}[htbp]      
	\centering           
	\subfloat[]{\label{fig:triangle}    
		\begin{minipage}[t]{0.3\linewidth}      
			\includegraphics[width=1\linewidth]{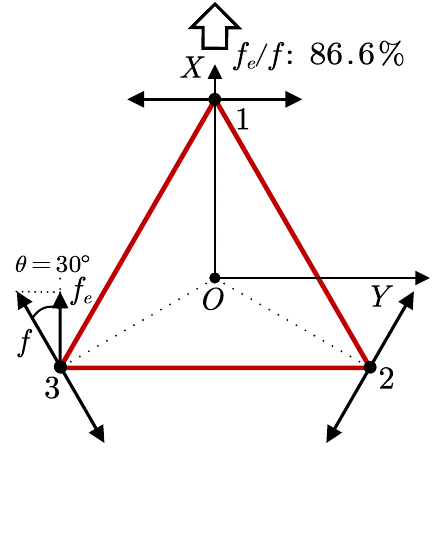}  
		\end{minipage}      
	}
	\subfloat[]{\label{fig:cross}       
		\begin{minipage}[t]{0.3\linewidth}            
			\includegraphics[width=1\linewidth]{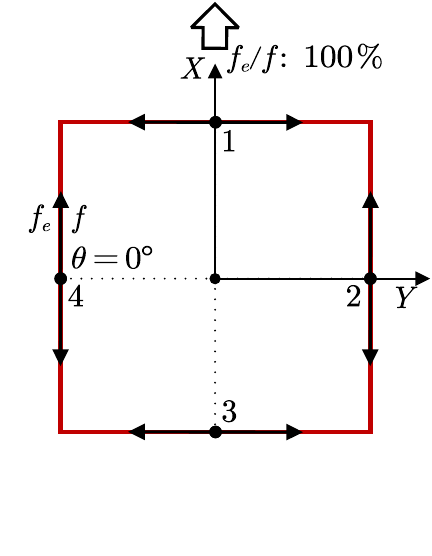} 
		\end{minipage}      
	}
	\subfloat[]{\label{fig:x}       
		\begin{minipage}[t]{0.3\linewidth}      
			\centering     
			\includegraphics[width=1\linewidth]{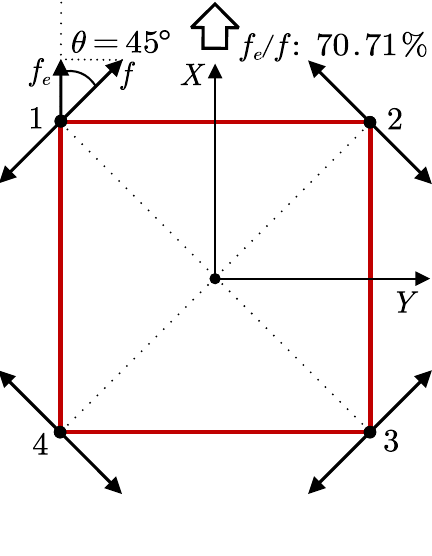}      
		\end{minipage}      
	}    
	\caption{Comparison among three different propulsion configuration: (a) three thruster, (b) "+" shaped and (c) "x" shaped configurations. The propulsion efficiency $f_e/f$ is calculated.}   
	\label{fig:propulsion_configuration}  
\end{figure}

\begin{figure} [htbp] 
	\centering
	\begin{minipage}[b]{.8\linewidth}
		\centering
		\subfloat[]{
			\includegraphics[width=1\linewidth]{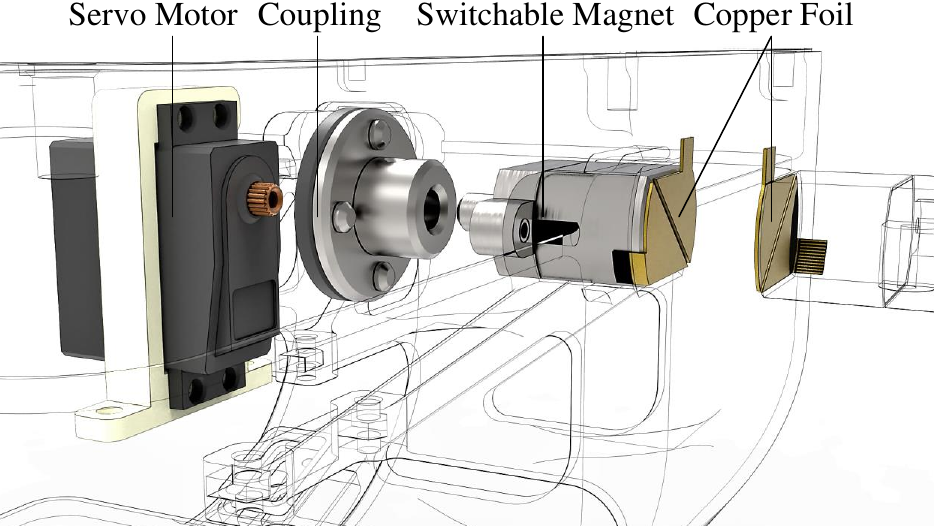}
		}
	\end{minipage}
	
	\begin{minipage}[b]{.9\linewidth}
		\centering
		\subfloat[]{
			\begin{overpic}[width=0.5\textwidth]{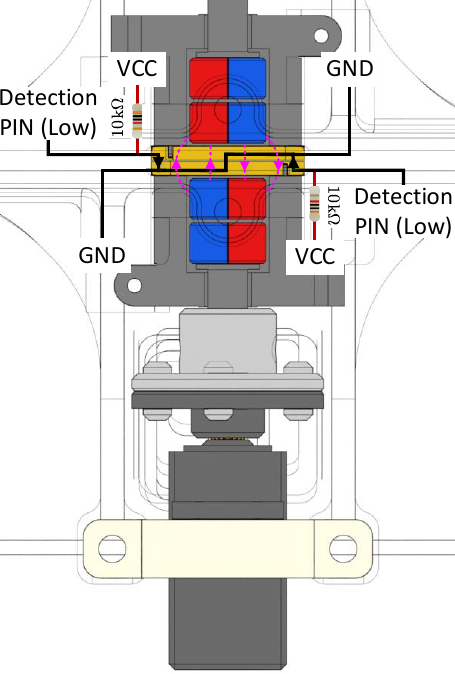}
				\put(-4,40){Docked}
			\end{overpic}
		}
		\subfloat[]{
			\begin{overpic}[width=0.5\textwidth]{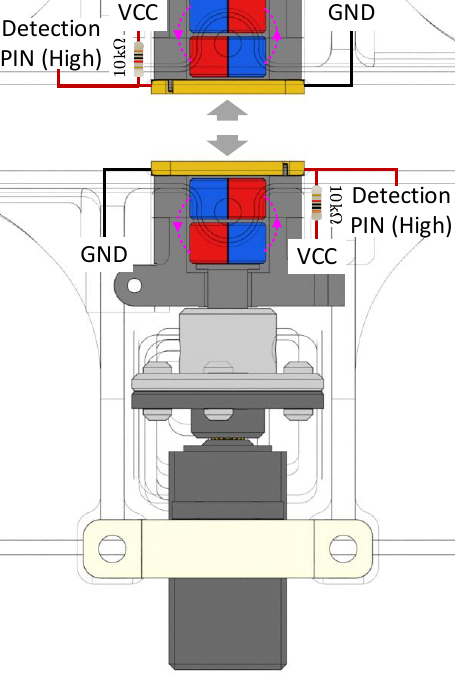}
				\put(-10,40){Undocked}
			\end{overpic}
		}
	\end{minipage}
	\caption{(a) Components of the docking subsystem. (b) The docked state and (c) undocked state when the magnet is switched on and off. The contact detection mechanism is presented.}
	\label{fig:DockingSystem}
\end{figure}

To leave no gap after the assembly, the CuBoat is designed as a cube-shaped boat with holonomic propulsion produced by four thrusters.
The electronics and the four docking systems are installed inside the cubic hull, while the switch, the charging port, and the data ports are reserved on the lid.

Four thrusters for the omnidirectional movements are installed in the "+" shaped configuration, which is more efficient than other holonomic configurations such as the "X" shaped actuator configuration \cite{NAD2015172} or the three-thruster configuration \cite{vallegra2018gradual}.
The comparison among the three main configurations is shown in Fig. \ref {fig:propulsion_configuration}, where
$f$ and $f_e$ are the overall and the effective propulsion forces of the functional thruster, respectively, and  
\begin{equation}
f_e = f \cos \theta
\end{equation}
with $\theta$ being the angle between the overall propulsion and the moving direction.
We can see that only the "+" shaped configuration can provide $100 \%$ propulsion efficiency.

\begin{figure} [htbp] 
	\centering
	\includegraphics[width=0.6\linewidth]{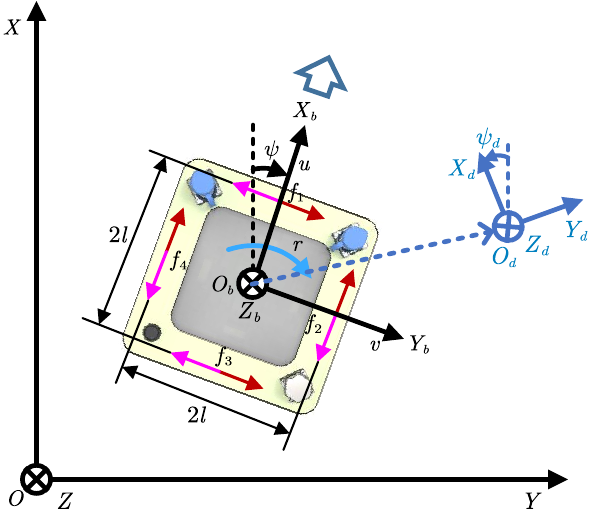}
	\caption{Coordinate system for the movement of the CuBoat: inertial coordinate $O \text{-} X Y Z$, body coordinate $O_b \text{-} X_b Y_b Z_b$ and target coordinate $O_d \text{-} X_d Y_d Z_d$. Red arrows stand for positive propulsion forces and pink arrows stand for negative forces.}
	\label{fig:motion_control}
\end{figure}

The body coordinate system and the thruster layout are shown in Fig. \ref{fig:motion_control}, of which all thrusters are capable of generating both forward and backward forces. 
Then, by using $f_i \ (i=1,2,3,4)$ to denote the propulsion forces generated by four thrusters and $l$ to denote the distance from each thruster to the body center, the applied force and moment vector  $\bm{\tau}$ in the plane can be represented as
\begin{equation} \label{eq:forces}
\bm{\tau } =\left[ \begin{matrix}
0&		1&		0&		1\\
1&		0&		1&		0\\
l&	   -l&	   -l&	    l\\
\end{matrix} \right] \left[ \begin{array}{c}
f_1\\
f_2\\
f_3\\
f_4\\ 
\end{array} \right].
\end{equation}

\subsubsection{Electronics} 
The main electronic components of the CuBoat include the sensors, the processors, the power supply, the docking subsystem, and the propulsion subsystem.
Specifically, two types of sensors are adopted here.
One is the OptiTrack motion capture system connected with the CuBoats through a local area network and a virtual-reality peripheral network (VRPN) interface to provide millimeter-precision positions, and the other is the contact detection circuit of the docking subsystem to detect whether two USVs are connected.
The processing unit is a Raspberry Pi 4B board (1.5 GHz ARM Cortex-A72, 2G LPDDR4) running the robot operating system (ROS) to execute the navigation and control algorithms. 
An Arduino board, which directly controls the four servo motors of the docking subsystems and the four thrusters of the propulsion subsystem, receives and conducts the control commands from the Raspberry Pi so that the processor can focus on communication and control.
A LiPo battery of 9800 mAh capacity with a buck module can supply both 5 V and 12 V power, which endows the CuBoat with 2 hours duration.

\subsubsection{Docking Subsystem}

Fig. \ref{fig:DockingSystem} (a) shows the main components of the isomorphic docking system, where the switchable magnet is actuated by a servo motor. 
When two magnets are switched on and attract each other, the longitudinal connection force is up to 226 N, and the lateral force is nearly 59 N. 
Even if only one-side magnet is on and attracting, the longitudinal and lateral forces are 113 N and 26 N, respectively.
Therefore, the magnetic force is strong enough to build a firm connection.
To monitor whether two CuBoats are successfully connected, a contact detection circuit is designed, of which the schematic diagram is sketched in Fig. \ref{fig:DockingSystem} (b) and (c).
Segmented by oblique cutting, two pieces of copper foil connecting to the detection pin and the ground, respectively, ensure that the foils on the two sides are mutually closed during the connection.
After docking, the magnetic force will press the two-side foils together, so that both the detection pins connect with the ground.
The voltage of the detection pin is high when two USVs are separated, and low when two USVs are docked.

\begin{figure} [tbp] 
	\centering
	\includegraphics[width=0.7\linewidth]{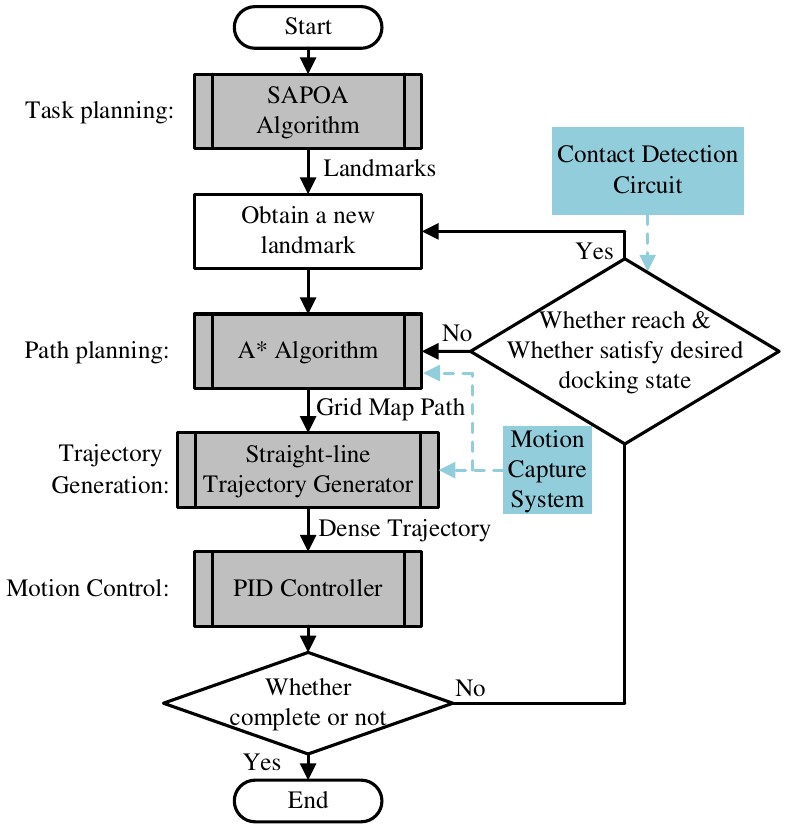}
	\caption{Architecture of the control system. }
	\vspace{-10pt}
	\label{fig:control_frame}
\end{figure}

\begin{figure} [htbp] 
	\centering
	\includegraphics[width=0.7\linewidth]{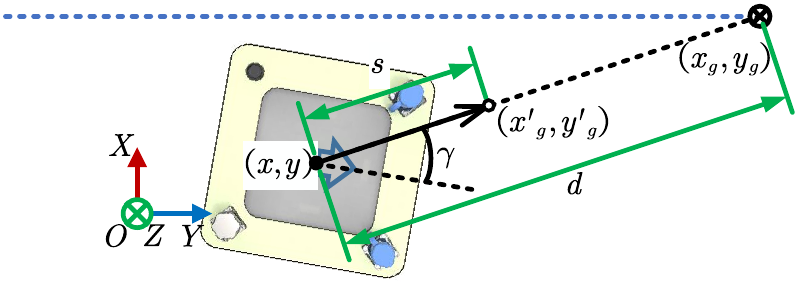}
	\caption{Trajectory Generator. The blue dotted line denotes the referenced grid map path. The solid, hollow circles and the circled X are the current position, the next trajectory point, and the target position, respectively. }
	\vspace{-10pt}
	\label{fig:traj_generator}
\end{figure}

\subsection{Navigation and Control System}  
\label{sect:control}

\begin{figure*}[htbp]      
	\centering
	\begin{minipage}[b]{.3\linewidth}
		\centering
		\subfloat[]{
			\includegraphics[width=1\linewidth]{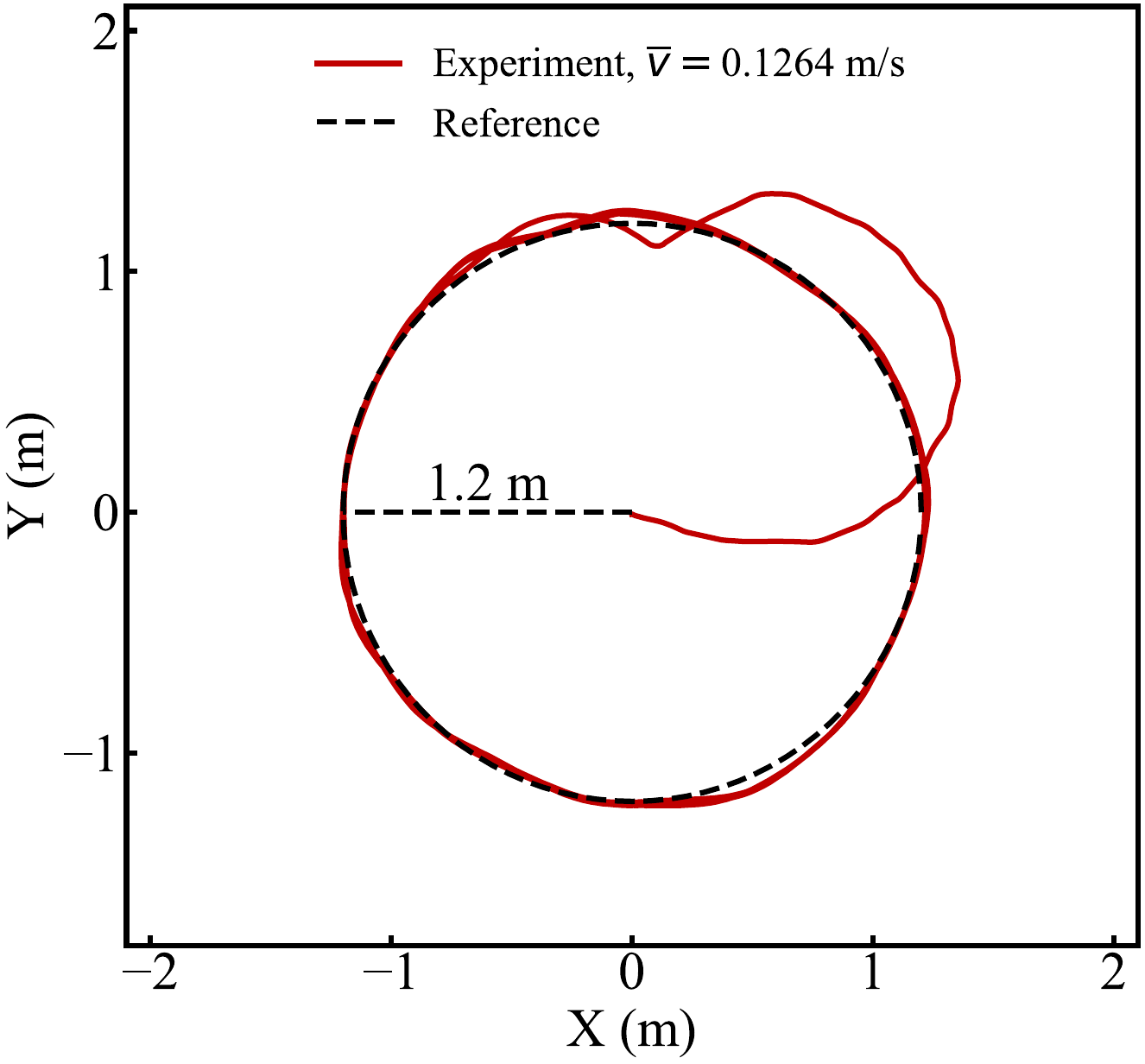}
		}
	\end{minipage}
	\begin{minipage}[b]{.3\linewidth}
		\centering
		\subfloat[]{
			\includegraphics[width=1\linewidth]{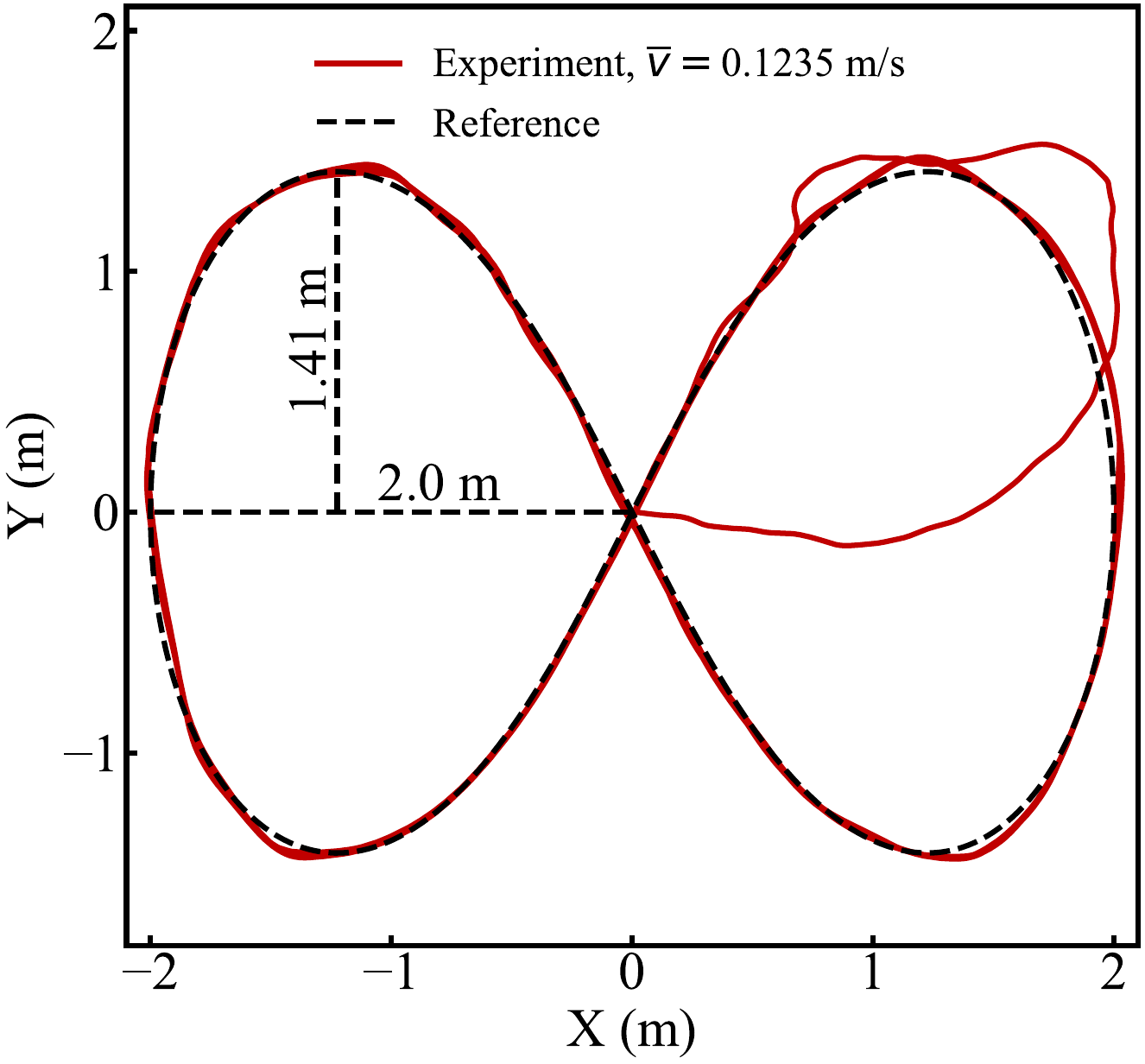}
		}
	\end{minipage}
	\begin{minipage}[b]{.27\linewidth}
		\centering
		\subfloat[]{
			\includegraphics[width=1\textwidth]{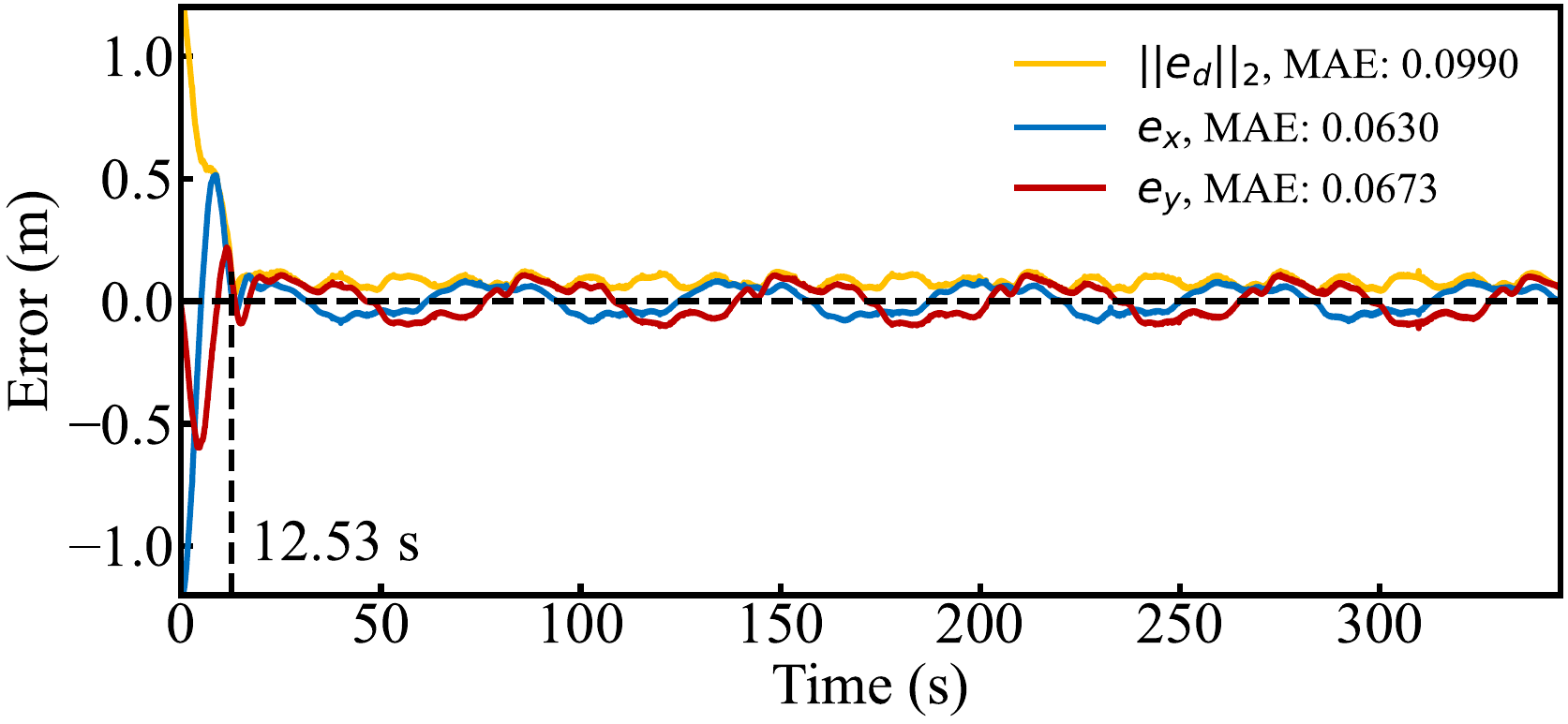}
		}
		\vspace{0pt}
		\subfloat[]{
			\includegraphics[width=1\textwidth]{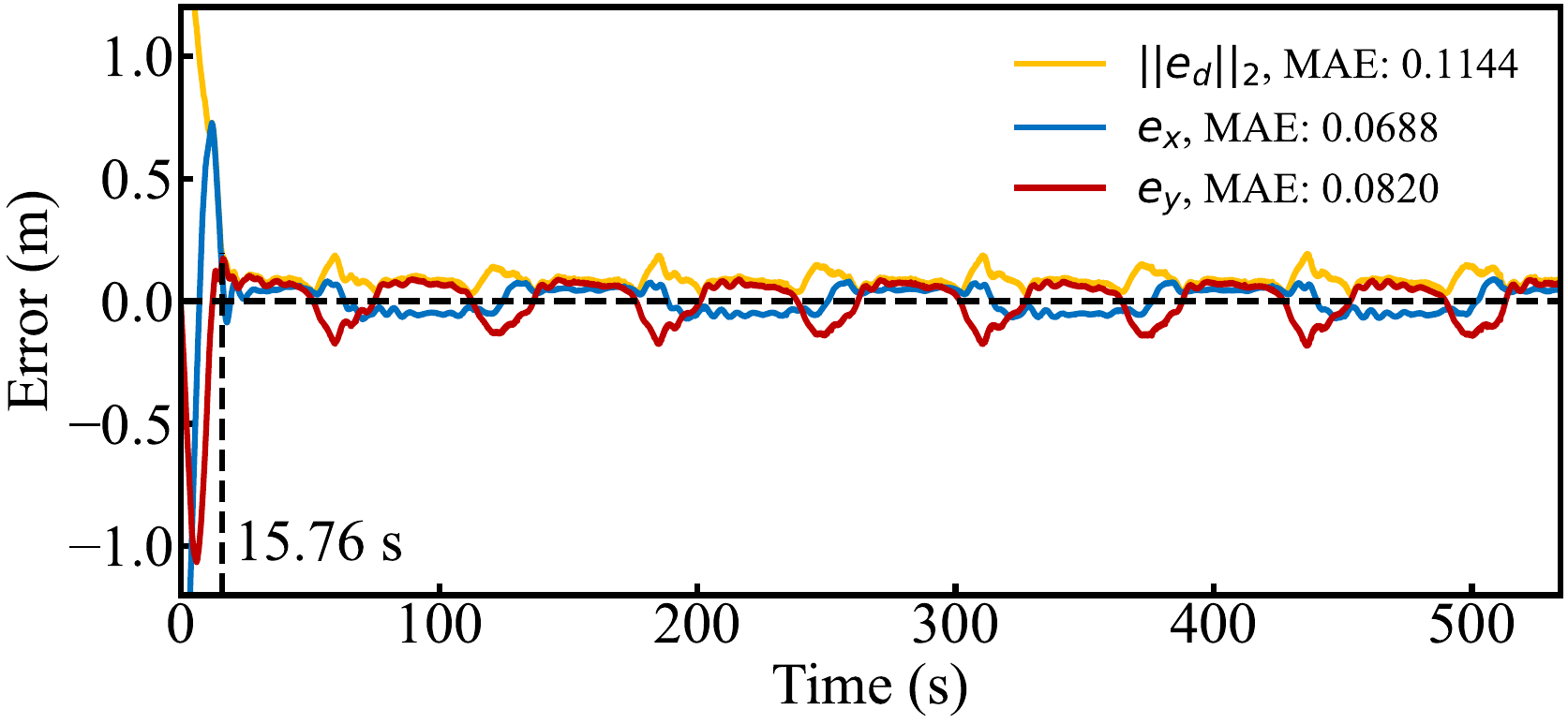}
		}
	\end{minipage}   
	\caption{Experimental paths and errors in tracking  (a)(c) the circular and (b)(d) the eight-shape trajectories. The average velocities $\bar{v}$ and the mean absolute errors (MAEs) are calculated. $e_x$ and $e_y$ are the errors in the X and Y directions, respectively, and $e_d = (e_x, e_y)$ stands for the overall error vector.}   
	\vspace{-10pt}
	\label{fig:tracking}  
\end{figure*}

Fig. \ref{fig:control_frame} illustrates the deployment of the SAPOA algorithm on the CuBoat Testbed.
First, the algorithm running on the central on-shore computer generates the landmark points, and sends them to the dispatched robots, respectively.
It is worth mentioning that although the docking system on the CuBoat is an active one, we still utilized the SAPOA algorithm rather than the SAPOAads, since the former, with a slightly higher moving cost, can reserve space to avoid collision among robots, which is a relatively safer method.
After receiving the landmarks, each CuBoat successively selects the landmark points as the current target, and real-timely plans the path from its present position to the target based on the A* algorithm, as described in Section \ref{sect:navigation}.
Following the rules in the robot navigation stage, all the CuBoats can only access the position information of the neighbor robots, so that they are moving in a decentralized manner.

To facilitate the execution of the A* algorithm, the experimental pool meshes into a grid map with a cell size of $ 0.25 \,\text{m} \times 0.25 \,\text{m}$.
From this, the high-level planner of each boat results in discrete paths whose points are in the grid center with significant gaps between each other.
For the consistency of motions, all CuBoats successively select the path points as the targets of their continuous low-level controllers at the same frequency ($0.25$ Hz).
Two assumptions are made here.
\begin{enumerate}
	\item Within a target-update cycle, each CuBoat can move at least one grid.
	\item Each CuBoat only occupies the desired grids.
\end{enumerate}
During the movement, a straight-line trajectory generator is employed, as shown in Fig. \ref{fig:traj_generator}.
Let  $\bm{x}=\left[ \begin{matrix} x& y\\\end{matrix} \right] ^T$ and  $\bm{x}_g=\left[ \begin{matrix} x_g& y_g\\\end{matrix} \right] ^T$ be the current position of the CuBoat and the target point from the A* algorithm, respectively, and then the next trajectory point to be generated can be calculated by 
\begin{equation} \label{eq:traj_generator}
\left[ \begin{array}{c}
x_g^{\prime}\\
y_g^{\prime}\\
\end{array} \right] =\left[ \begin{array}{c}
x\\
y\\
\end{array} \right] +\min \left\{ s,d \right\} \left[ \begin{array}{c}
\cos \gamma\\
\sin \gamma\\
\end{array} \right] ,
\end{equation}
where $s$ is the step length input by user, $d=\lVert \bm{x}_g-\bm{x} \rVert _2$ measures the distance between the current position and the target, and $\gamma =\arctan \left( \frac{y_g-y}{x_g-x} \right) $ denotes the slope angle of $ \bm{x}_g-\bm{x}$.

Finally, three independent low-level PID controllers are adopted for the CuBoat movement. 
We represent the CuBoats in a two-dimensional inertial coordinate ($O \text{-} X Y Z$), where the coordinates $O_b \text{-} X_b Y_b Z_b$ and $O_d \text{-} X_d Y_d Z_d$ denote the body frame and the target frame, respectively, as shown in Fig. \ref{fig:motion_control}. 
The homogeneous representation of the current position and the target to the global frame $O \text{-} X Y Z$ can be introduced as
\begin{equation} \label{eq:representation}
\bm{\eta }=\left[ \begin{matrix}
x&		y&		\psi&		1\\
\end{matrix} \right] ^T,
\bm{\eta }_g=\left[ \begin{matrix}
x_g&		y_g&		\psi _g&		1\\
\end{matrix} \right] ^T.
\end{equation}
The coordinate values $x_g$, $y_g$, and $\psi_g$ also denote the errors for the motion controller, since the controllers aim to overlap the target point $\bm{\eta }_g$ with the origin $\bm{\eta }$ in the body frame.
Therefore, we need to transform the target point $\bm{\eta }_g$ from the global frame $O \text{-} X Y Z$ to the body frame $O_b \text{-} X_b Y_b Z_b$, which is
\begin{equation} \label{eq:transform}
\bm{\widetilde{\eta}}_g =\mathbf{A}^g \bm{\eta}_g,
\end{equation}
where $\widetilde{\bm{\eta }}_g=\left[ \begin{matrix}
\widetilde{x}_g&		\widetilde{y}_g&		\widetilde{\psi }_g&		1\\
\end{matrix} \right] ^T$ is the transformed target and $\mathbf{A}^g$ is the homogeneous transformation matrix as follows
\begin{equation} \label{eq:homo_transform}
\mathbf{A}^{g}=\left[ \begin{matrix}
\cos \psi&		-\sin \psi&		0&		-x\cos \psi +y\sin \psi\\
\sin \psi&		\cos \psi&		0&		-x\sin \psi -y\cos \psi\\
0&		0&		1&		-\psi\\
0&		0&		0&		1\\
\end{matrix} \right]. 
\end{equation}

Through Eq.(\ref{eq:transform}), the errors are decoupled, and the position and the orientation can be independently controlled.
We employed the Ziegler–Nichols method \cite{654876} to tune the PID controller. Its major procedure is first setting the integral and differential gains to zero, then enlarging the proportional gain till the system oscillation, and finally assigning the PID gains as a function of the proportional gain and the frequency of oscillation at the point of instability.
Three PID controllers are tuned according to the Ziegler–Nichols method and used to control the longitudinal, lateral, and rotational motions, respectively, of which the PID gains ($K_P, K_I, K_D$) are exhibited in Table \ref{tab:pid_gains}. 
Notably, all the CuBoats in the testbed adopt identical parameters, since fabricated according to the unified criterion.

\newcommand{\pathfigwidth}{0.25}
\begin{figure*} [htbp]
	\centering
	\subfloat[]{
		\begin{minipage}[b]{\pathfigwidth\linewidth}      
			\centering           
			\includegraphics[width=1\linewidth]{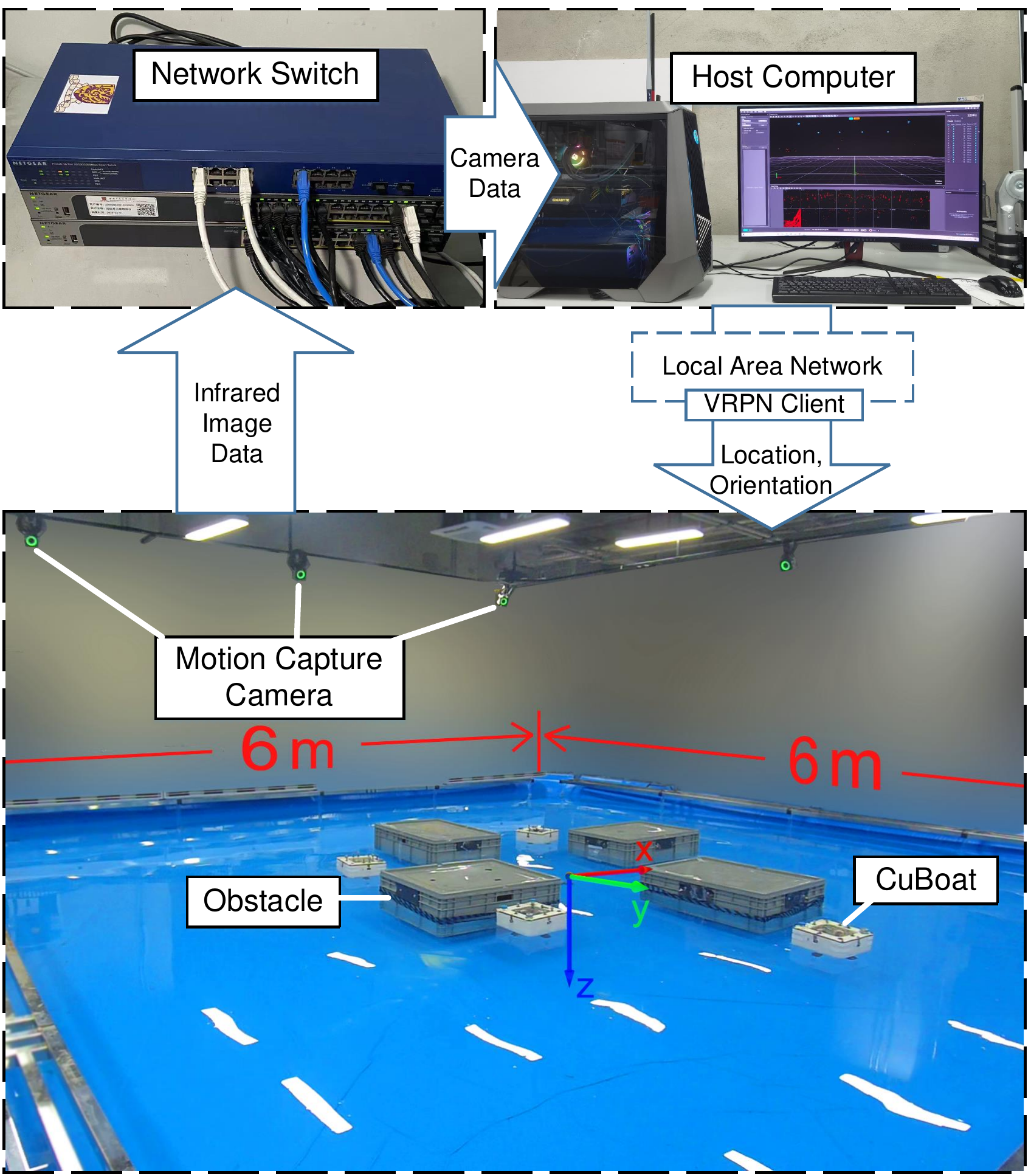}
			\vspace{-5pt}
		\end{minipage}
	}
	\subfloat[]{     
		\begin{minipage}[b]{\pathfigwidth\linewidth}      
			\centering
			\hspace{-15pt}      
			\includegraphics[width=1\linewidth]{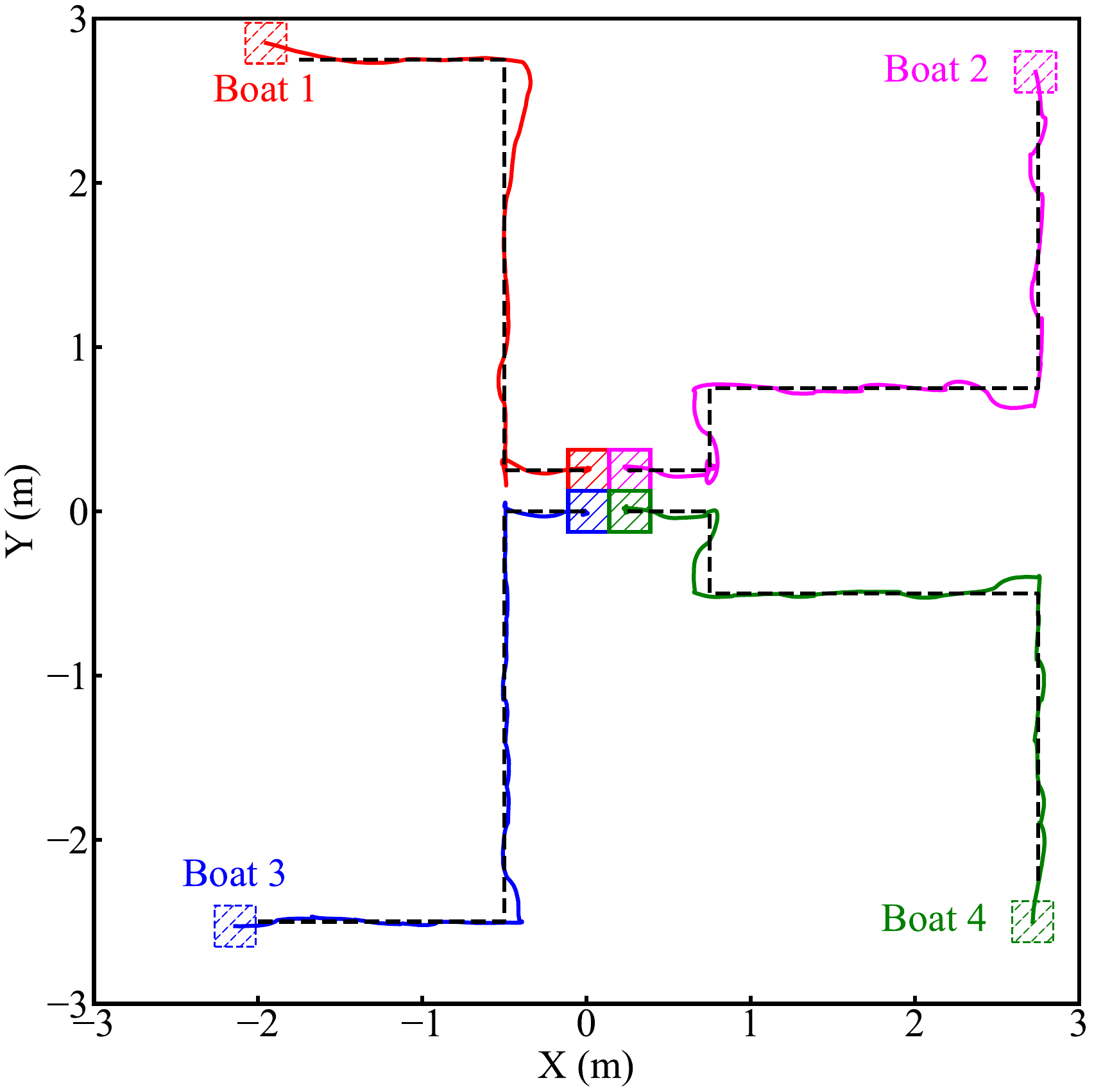}   \\
			\includegraphics[width=1\linewidth]{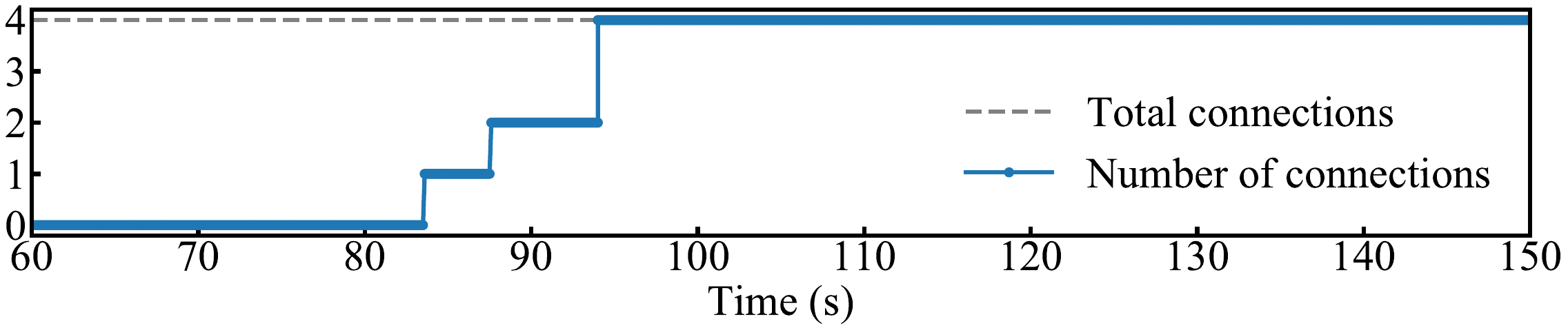}
		\end{minipage} 
	}
	\subfloat[]{
		\begin{minipage}[b]{\pathfigwidth\linewidth}      
			\centering
			\hspace{-15pt}            
			\includegraphics[width=1\linewidth]{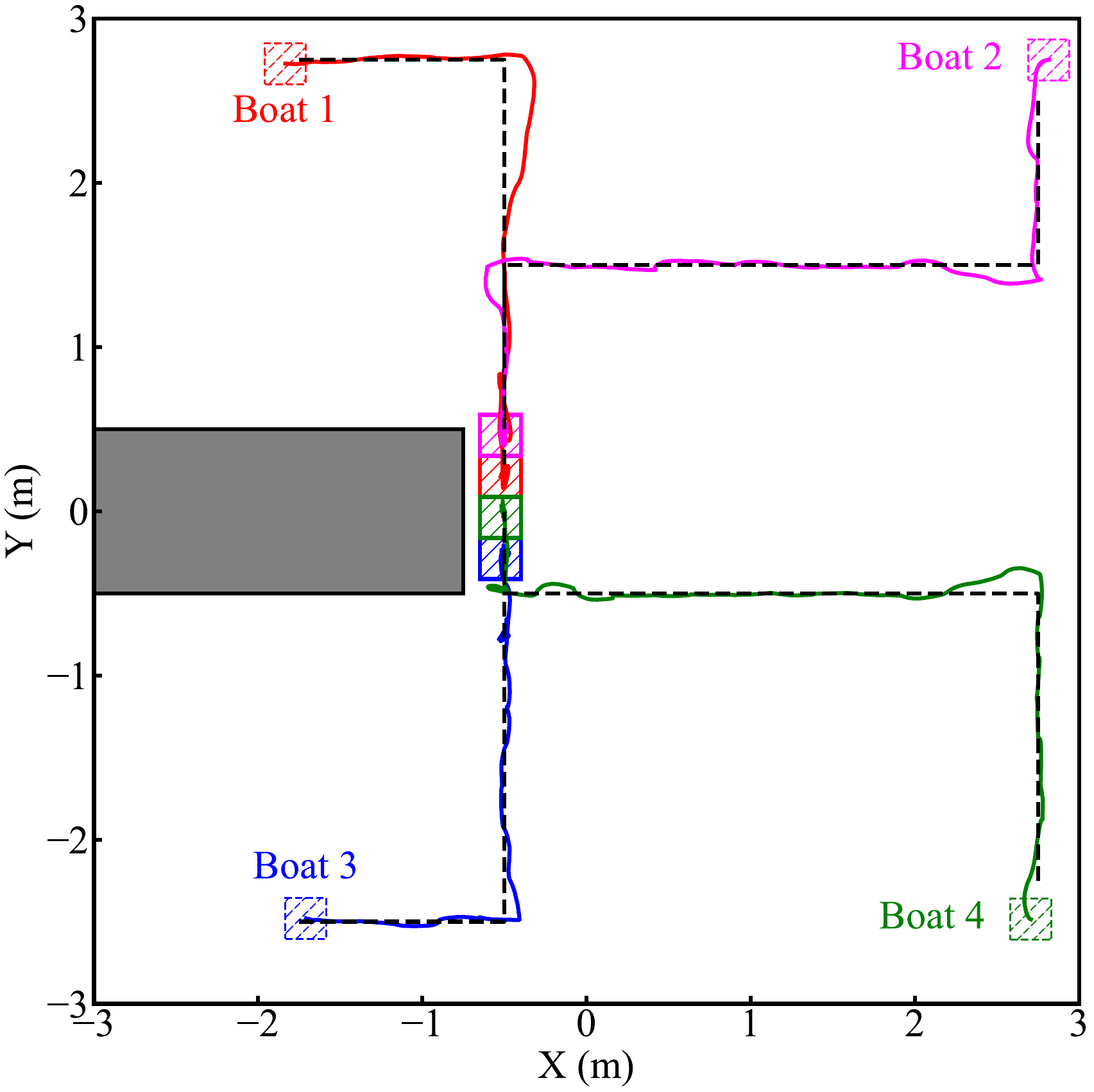}   \\
			\includegraphics[width=1\linewidth]{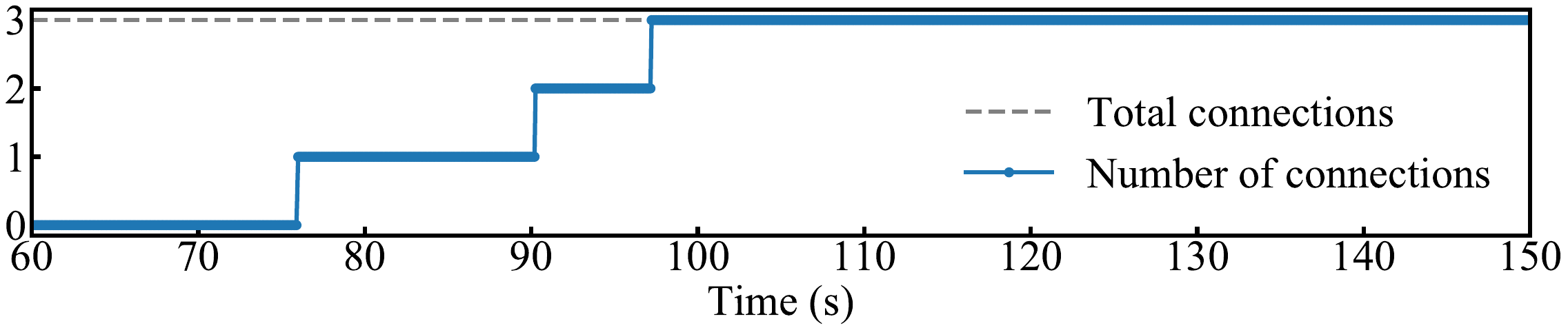}
		\end{minipage}      
	}
	
	\subfloat[]{
		\begin{minipage}{\pathfigwidth\linewidth}      
			\centering
			\hspace{-15pt}             
			\includegraphics[width=1\linewidth]{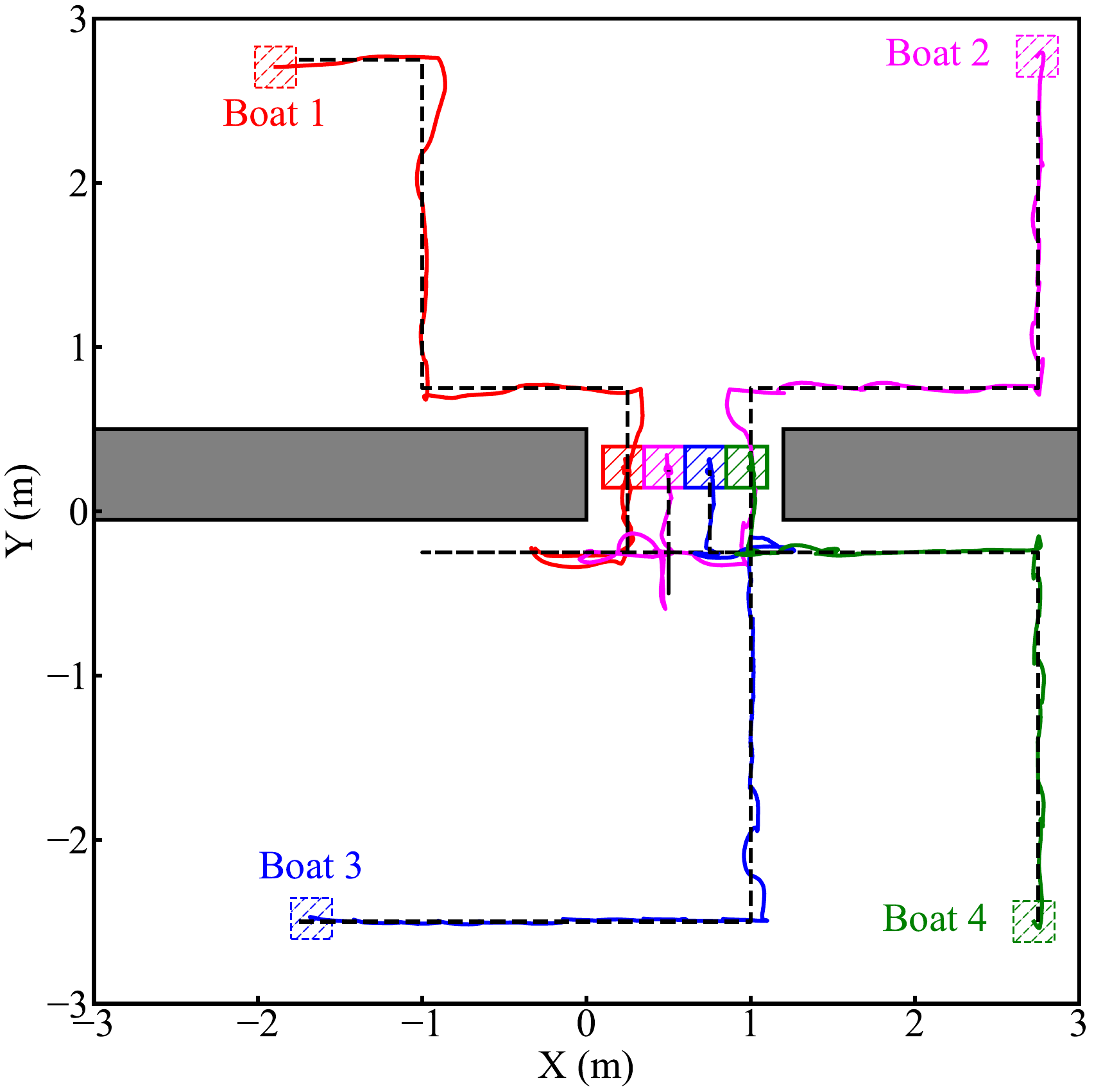}  \\
			\includegraphics[width=1\linewidth]{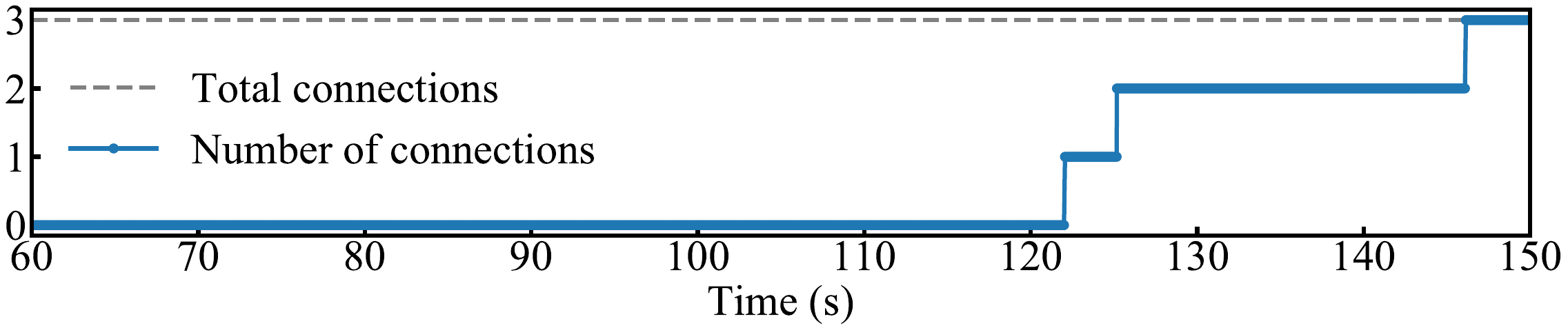}   
		\end{minipage}      
	}
	\subfloat[]{
		\begin{minipage}{\pathfigwidth\linewidth}      
			\centering
			\hspace{-15pt}            
			\includegraphics[width=1\linewidth]{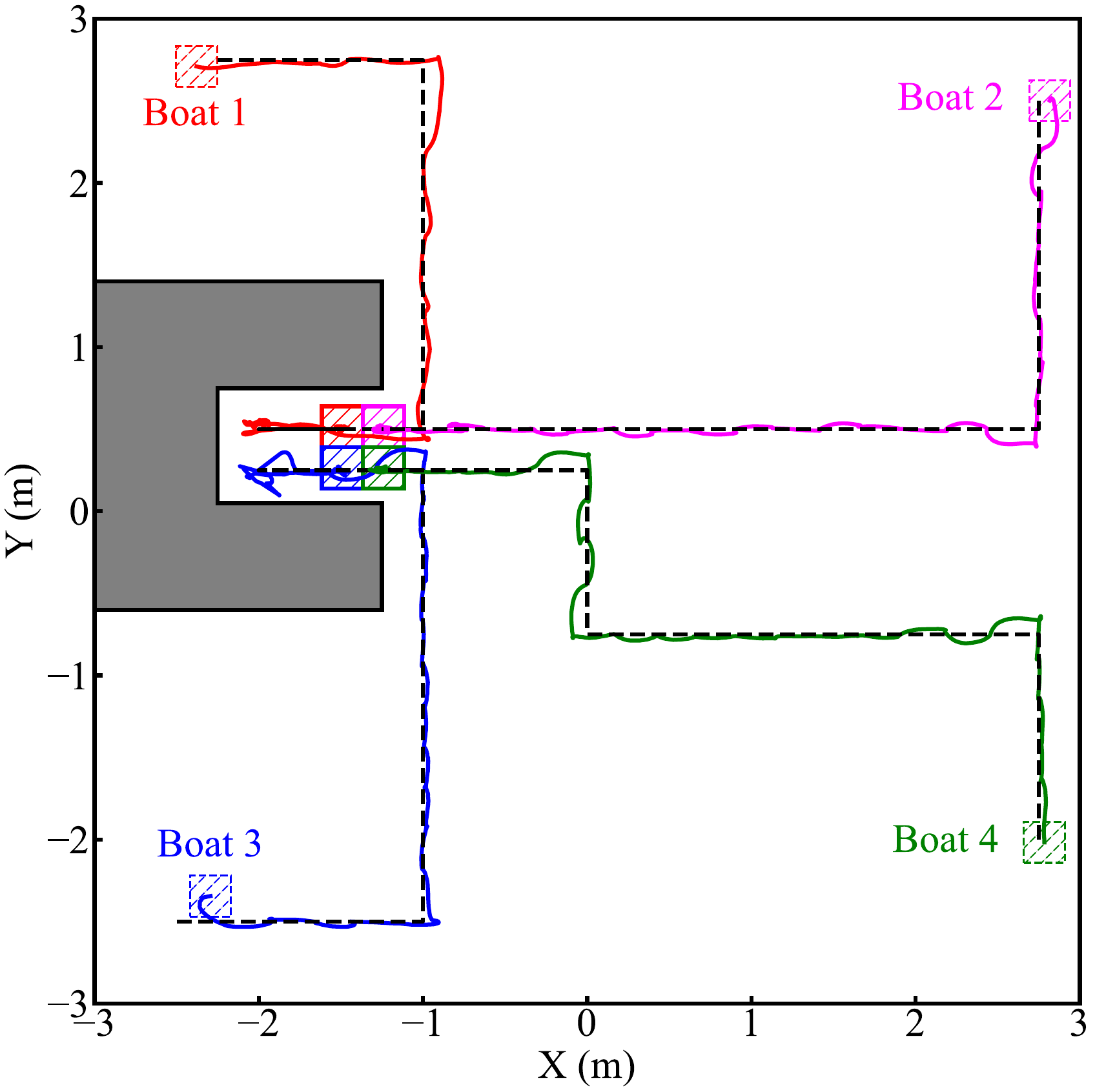}  \\
			\includegraphics[width=1\linewidth]{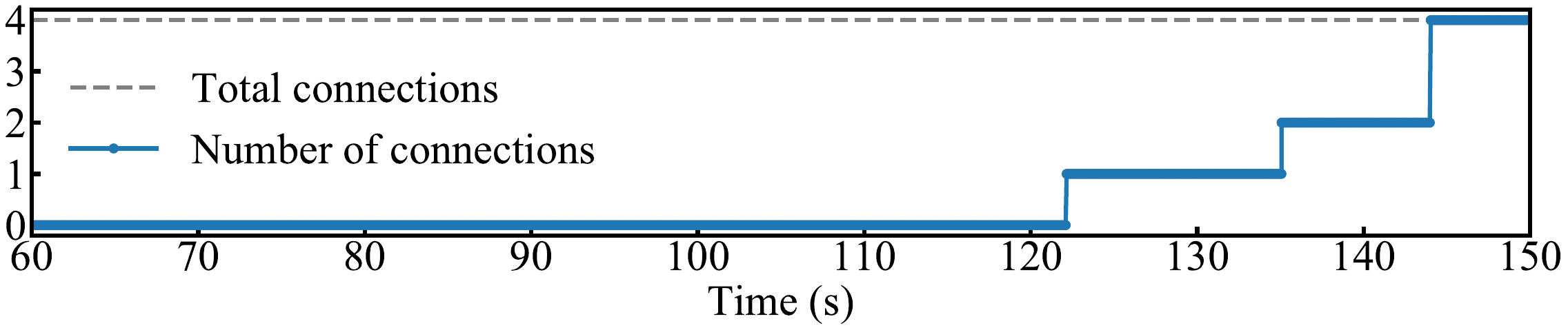}   
		\end{minipage}      
	}
	\subfloat[]{
		\begin{minipage}{\pathfigwidth\linewidth}      
			\centering
			\hspace{-15pt}            
			\includegraphics[width=1\linewidth]{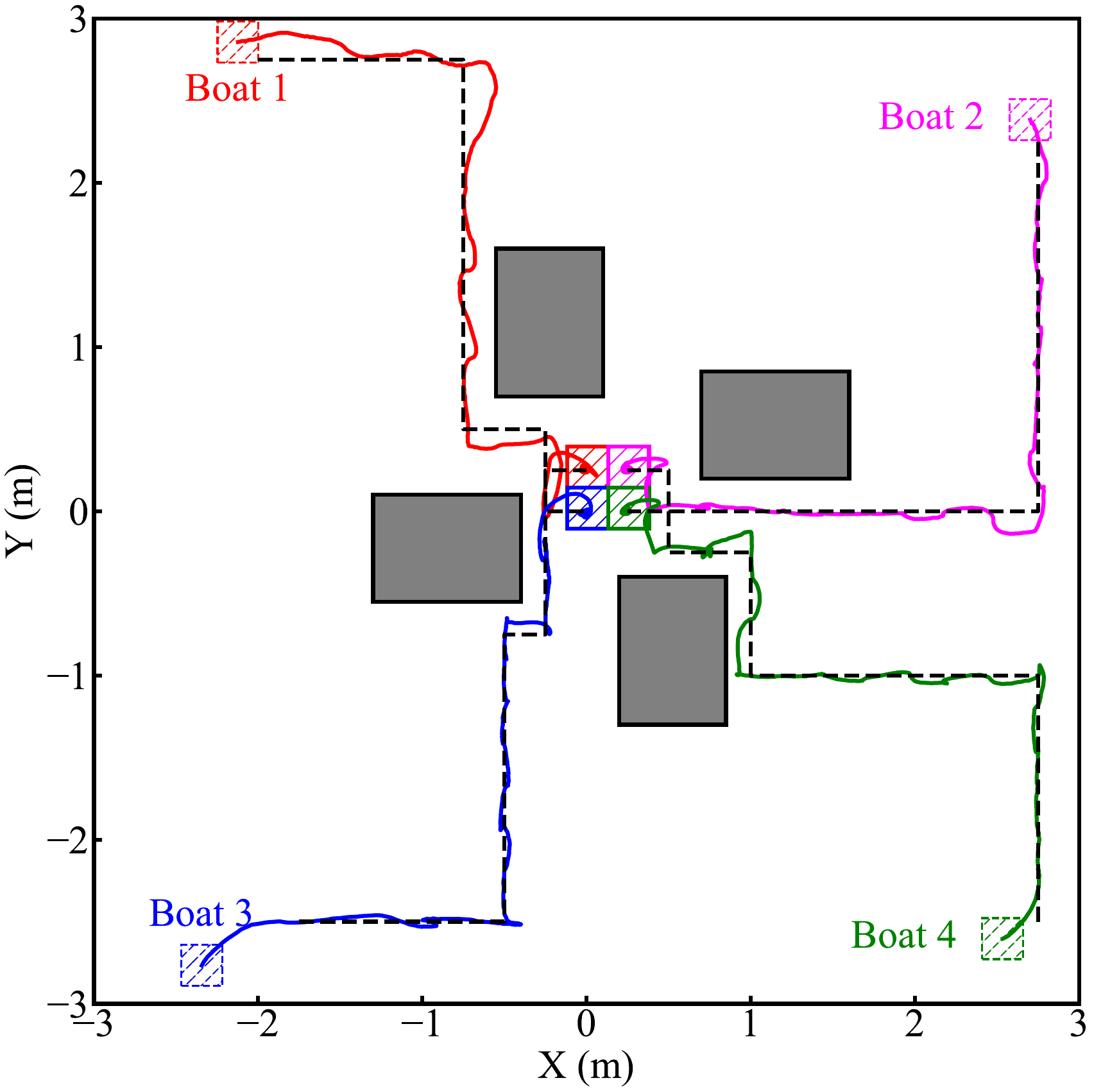}  \\
			\includegraphics[width=1\linewidth]{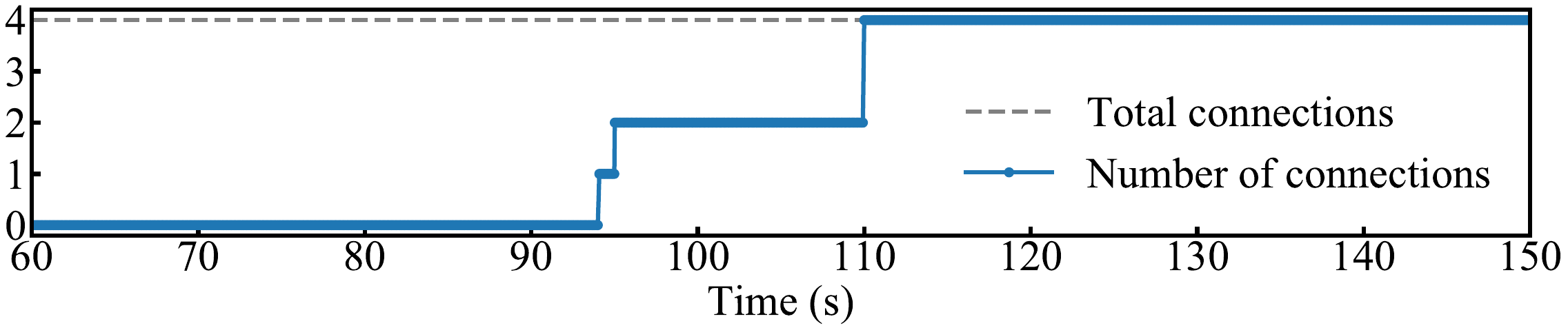}   
		\end{minipage}
	}   
	\caption{Experiments with five different obstacle configurations. Panel (a) pictures the experimental setup. The upper graphs of panels (b)-(f) show the trajectories of the CuBoats. The dashed-line boxes represent the robots in their initial locations, the continuous-line boxes represent their final destinations, the dashed lines represent the reference trajectories generated by the A* algorithm, and the continuous lines represent their experimental trajectories. The lower graphs of panels (f)-(j) show the evolution of completed connections on time. }
	\vspace{-10pt}
	\label{fig:exp_path}
\end{figure*}

\newcommand{\errorfigwidth}{0.19}
\begin{figure*} [!h]
	\centering
	\subfloat[]{
		\begin{minipage}[b]{\errorfigwidth\linewidth}      
			\centering           
			\includegraphics[width=1\linewidth]{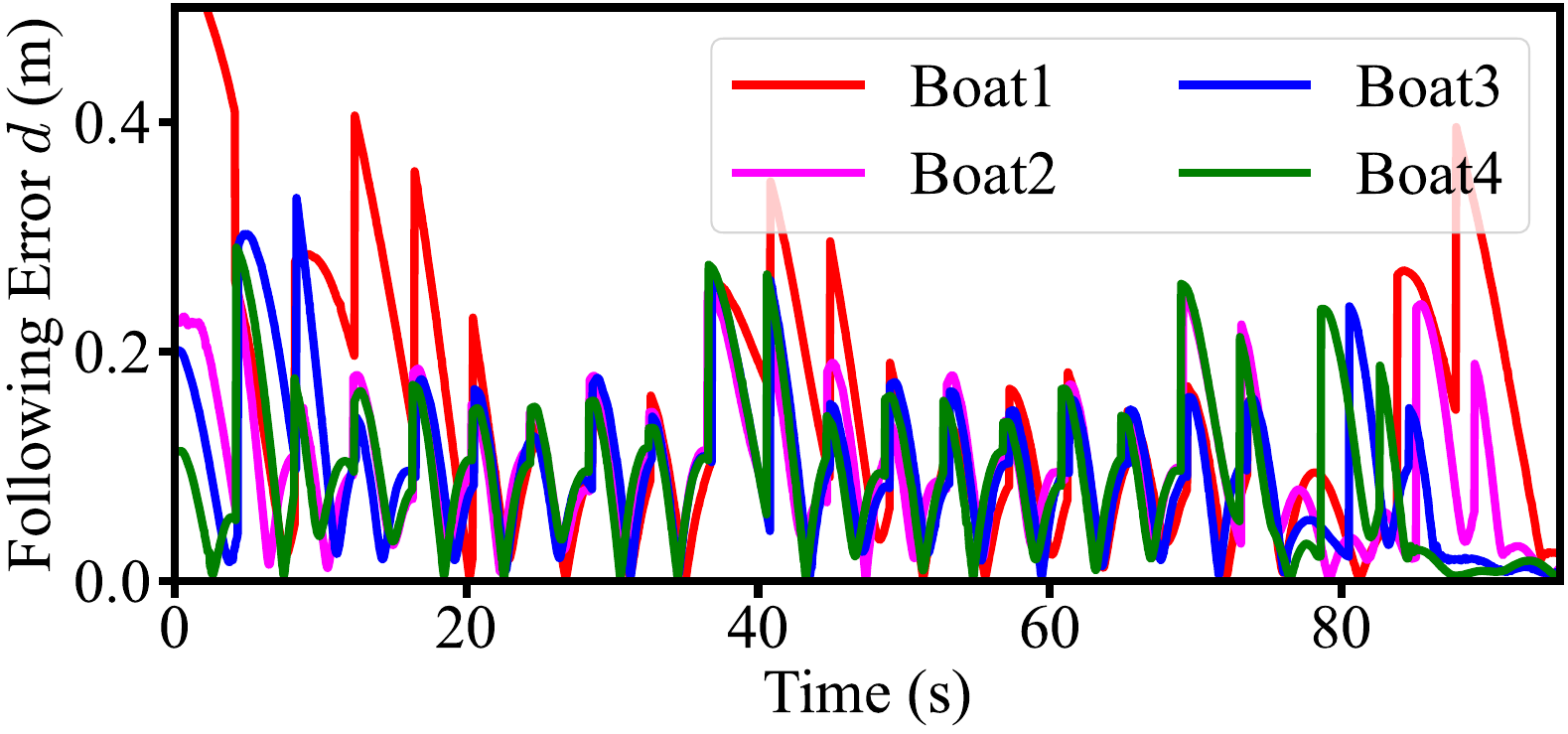} \\
			\includegraphics[width=1\linewidth]{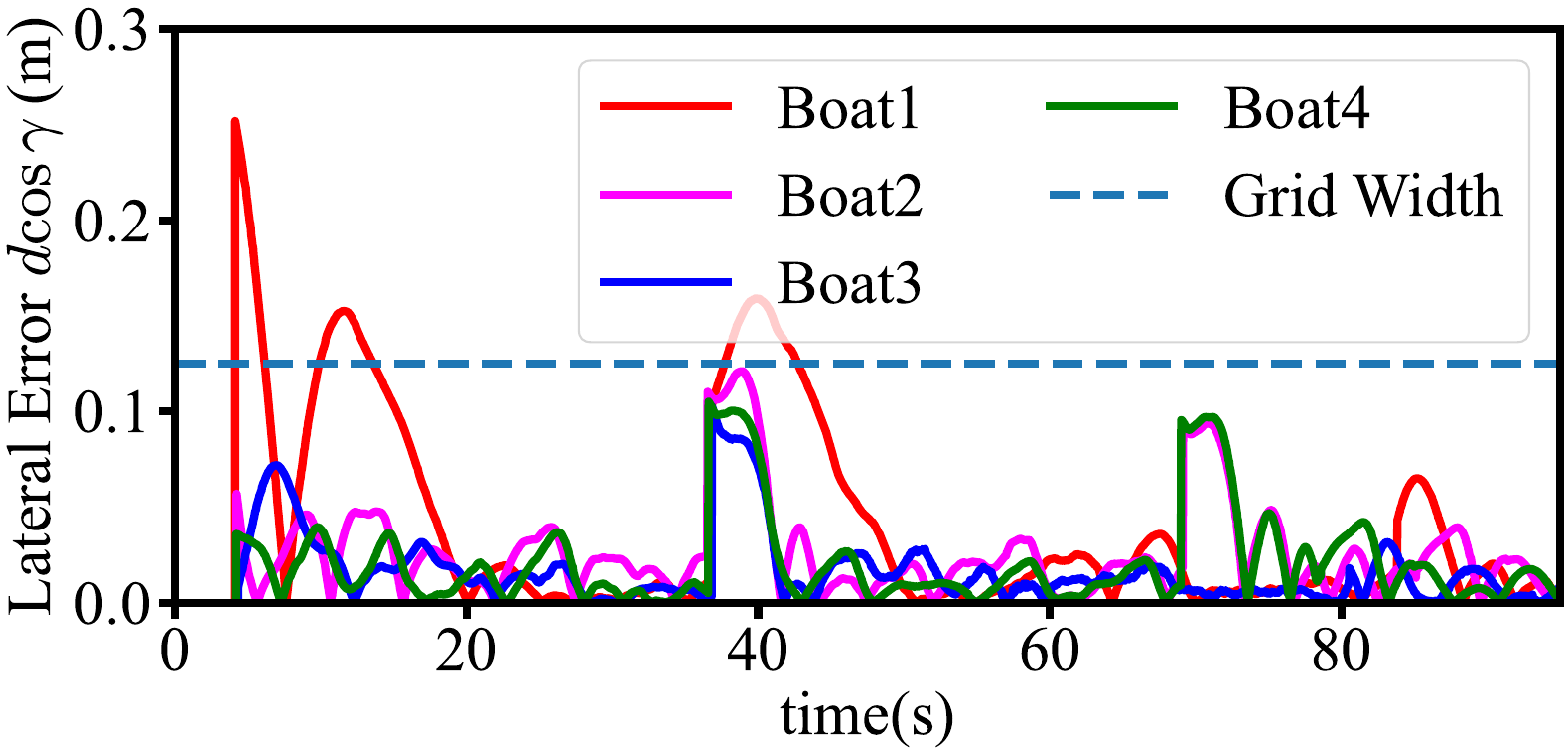}
		\end{minipage}
	}
	\subfloat[]{     
		\begin{minipage}[b]{\errorfigwidth\linewidth}      
			\centering
			\includegraphics[width=1\linewidth]{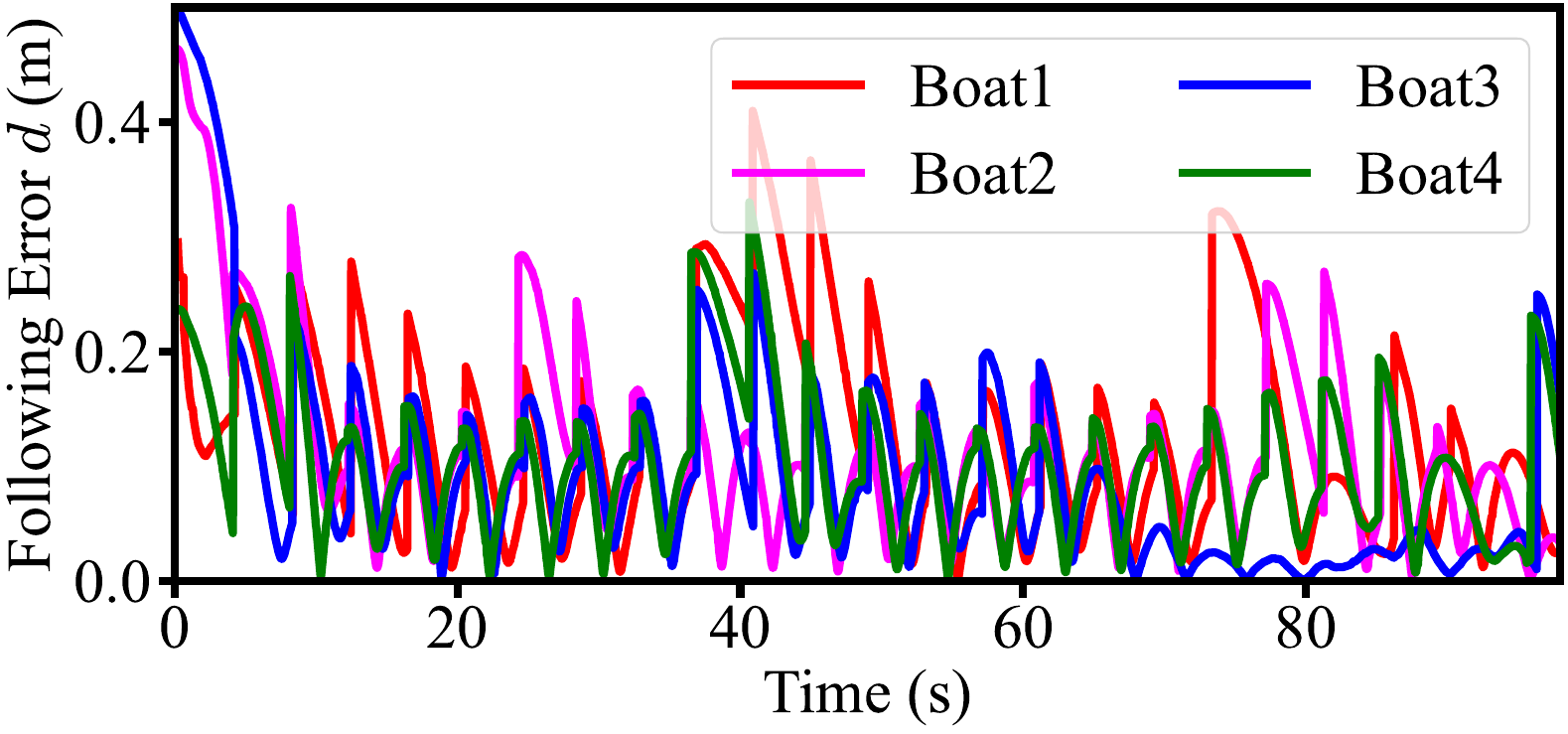} \\
			\includegraphics[width=1\linewidth]{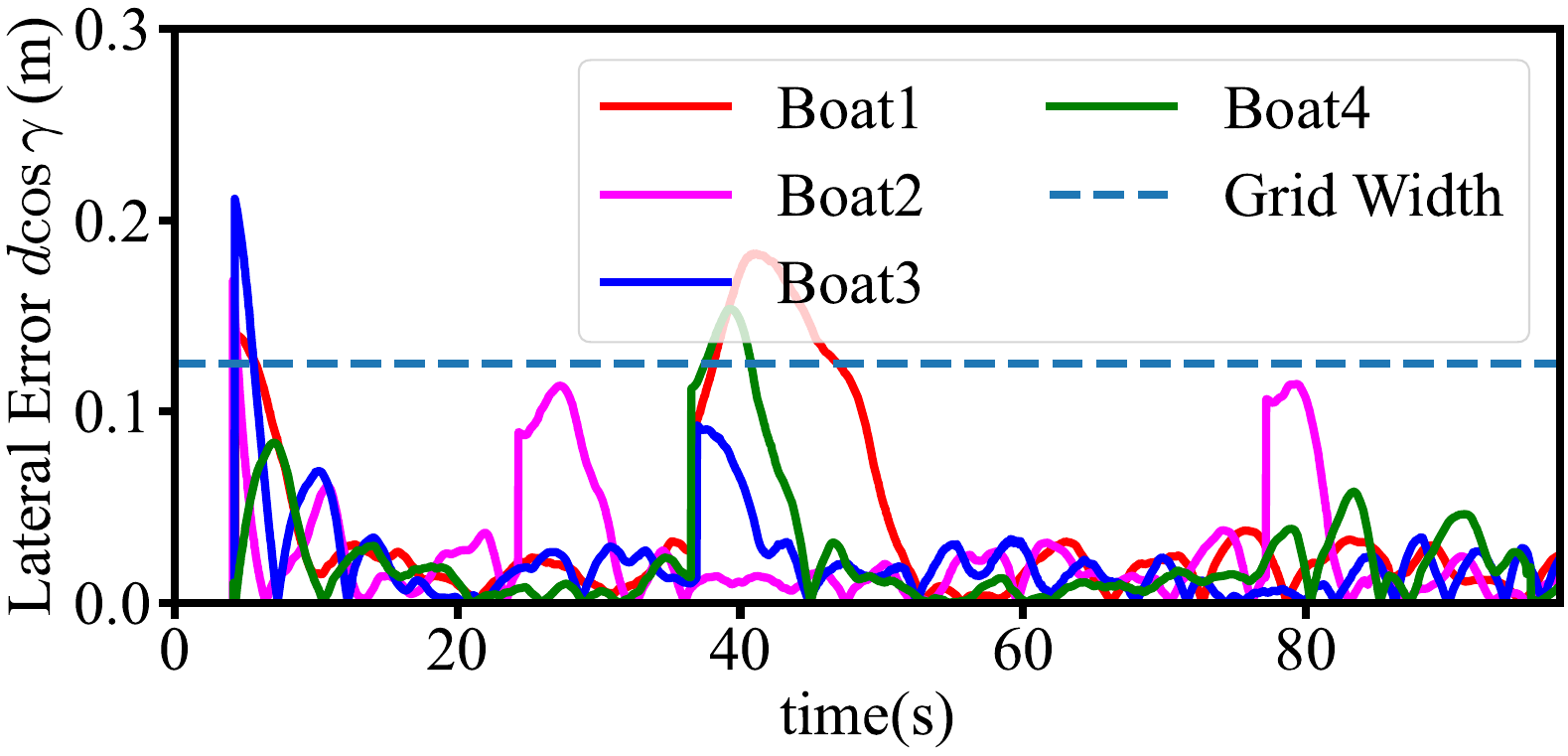}
		\end{minipage} 
	}
	\subfloat[]{
		\begin{minipage}[b]{\errorfigwidth\linewidth}      
			\centering
			\includegraphics[width=1\linewidth]{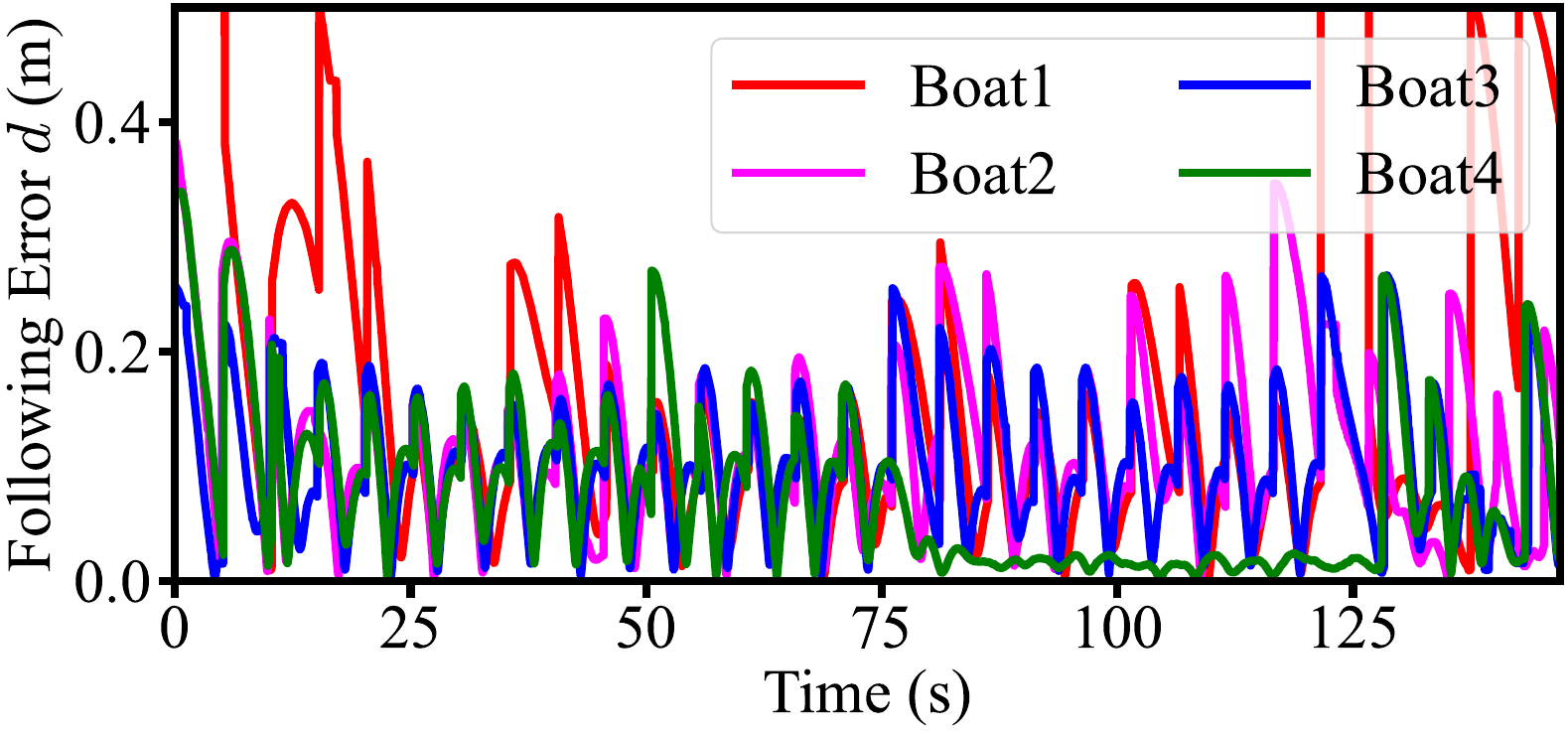} \\
			\includegraphics[width=1\linewidth]{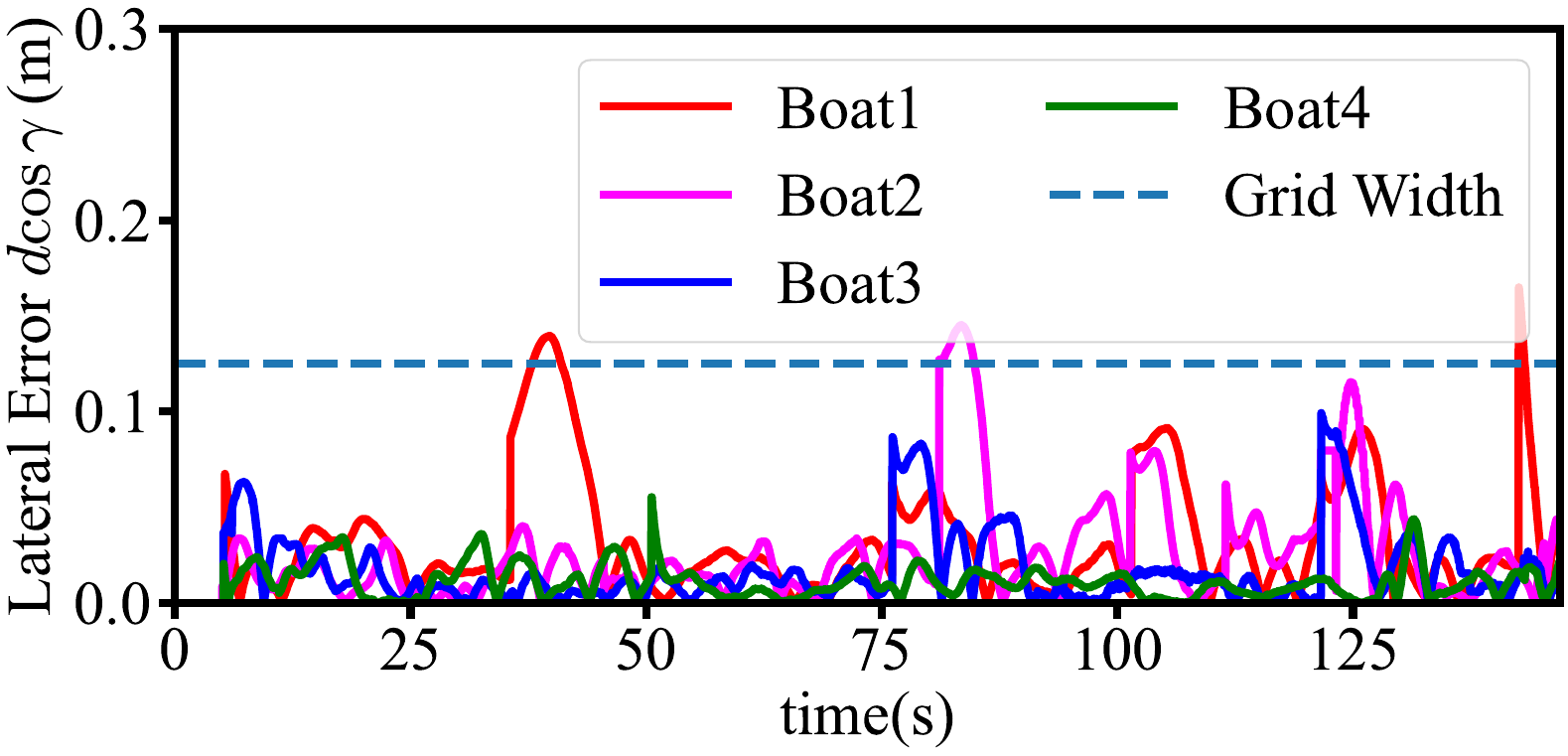}
		\end{minipage}      
	}
	\subfloat[]{
		\begin{minipage}[b]{\errorfigwidth\linewidth}      
			\centering
			\includegraphics[width=1\linewidth]{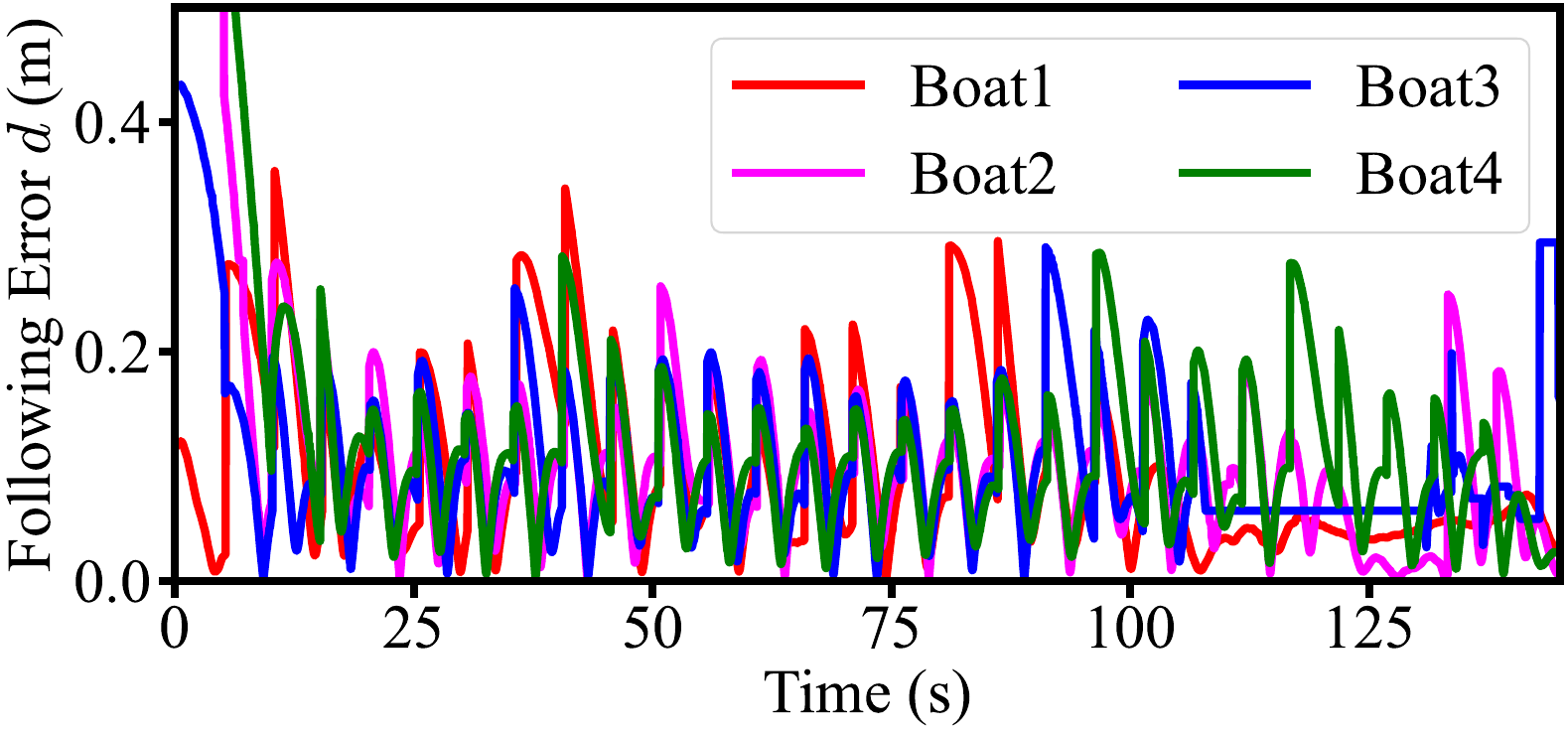} \\
			\includegraphics[width=1\linewidth]{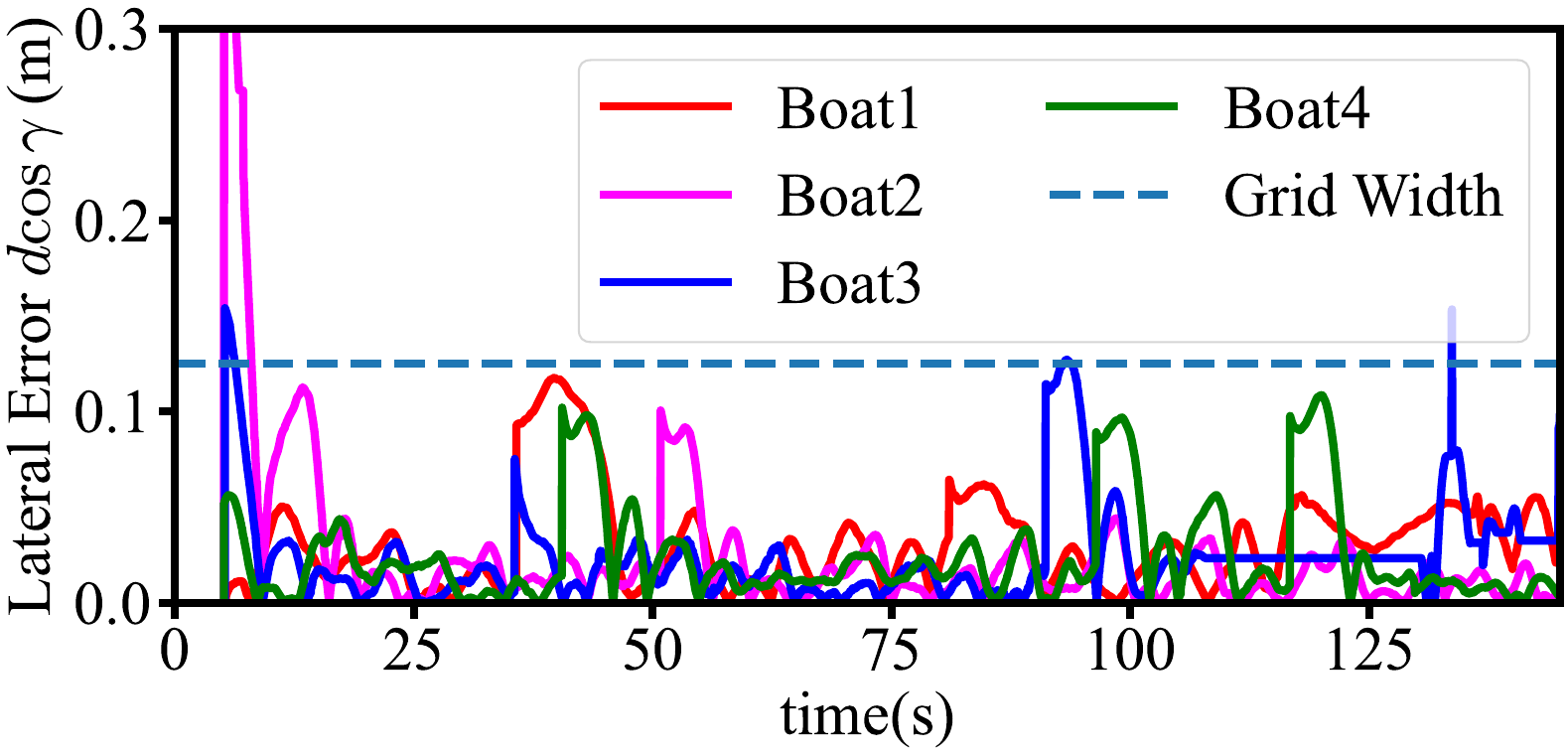}   
		\end{minipage}      
	}
	\subfloat[]{
		\begin{minipage}[b]{\errorfigwidth\linewidth}      
			\centering
			\includegraphics[width=1\linewidth]{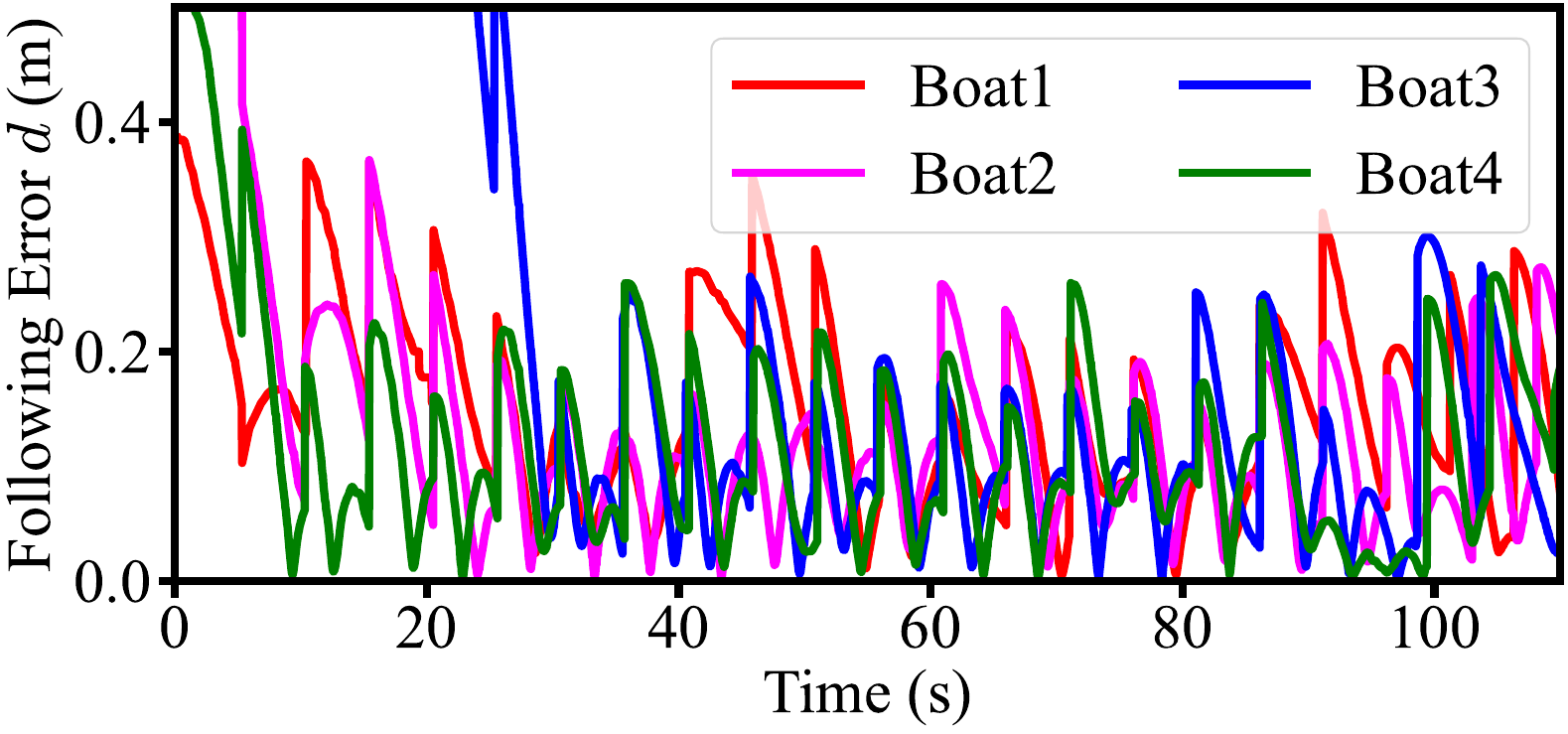} \\
			\includegraphics[width=1\linewidth]{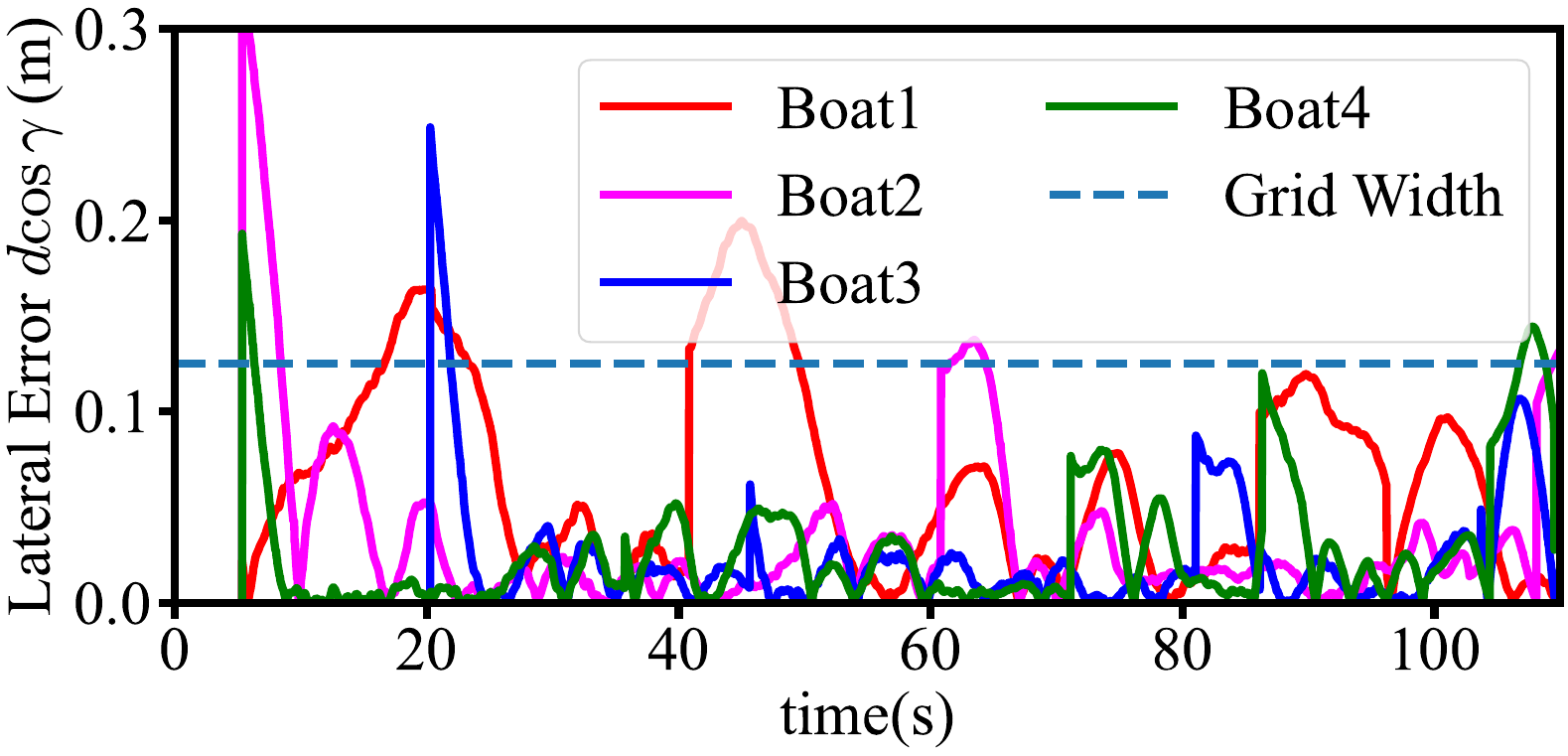}  
		\end{minipage}      
	}  
	\caption{The upper and lower graphs display the following errors $d$ and the lateral errors $d\cos \gamma$, respectively.}
	\vspace{-10pt}
	\label{fig:exp_error}
\end{figure*}

\begin{table}[tbp]
	\centering
	\caption{PID CONTROLLER GAINS}
	\label{tab:pid_gains}
	\setlength{\tabcolsep}{5mm}{
		\begin{tabular}{cccc}
			\toprule[1.5pt]
			& {Longitudinal} & {Lateral} & {Rotational} \\ 
			\hline
			{$K_P$}          & 2210     & 1500       & 153  \\
			{$K_I$}          & 7        & 10.5       & 2.4   \\
			{$K_D$}          & 4.67     & 7          & 1.6 \\  
			\bottomrule[1.5pt]
	\end{tabular}}
\end{table}

\section{Experimental Results and Analysis}
\label{sect:experiment}

\subsection{Trajectory Tracking}

To authenticate the PID motion controllers, two trajectory tracking tests are conducted, in which the trajectories embody a circle and a figure-eight shape.
Fig. \ref{fig:tracking} portrays the experimental paths and the tracking errors, from which we can draw two conclusions.
First, the tracking errors of both tests are small enough for the precise movement of one CuBoat and the alignment between the two.
Specifically, the MAE in one direction (at most $0.082 \, \text{m}$) and the overall MAE (no larger than $0.114 \, \text{m}$) are both much less than the length of a boat side $0.25 \, \text{m}$, which guarantees that two CuBoats can easily align to each other.
It is also noteworthy that the error of tracking a complicated trajectory (the figure-eight curve) is superior to that of a simple trajectory (circle) ($0.114 \, \text{m}$).
Second, the convergence rates of the controllers are sufficiently fast for the rapid docking tasks. 
In each test, the initial points of both tests are $(0,0)$, that is, the initial errors are greater than $1 \, \text{m}$, nearly $5$ times the size of the robot.
Nevertheless, neither of the settling time exceeds $16 \, \text{s}$. 
Therefore, the PID controllers are qualified for the self-assembly task in this paper.

\newcommand{\figwidth}{0.15}
\begin{figure*} [tbp]
	\centering
	\vspace{-5pt}
	\subfloat[]{
		\centering
		\begin{overpic}[width=\figwidth\textwidth]{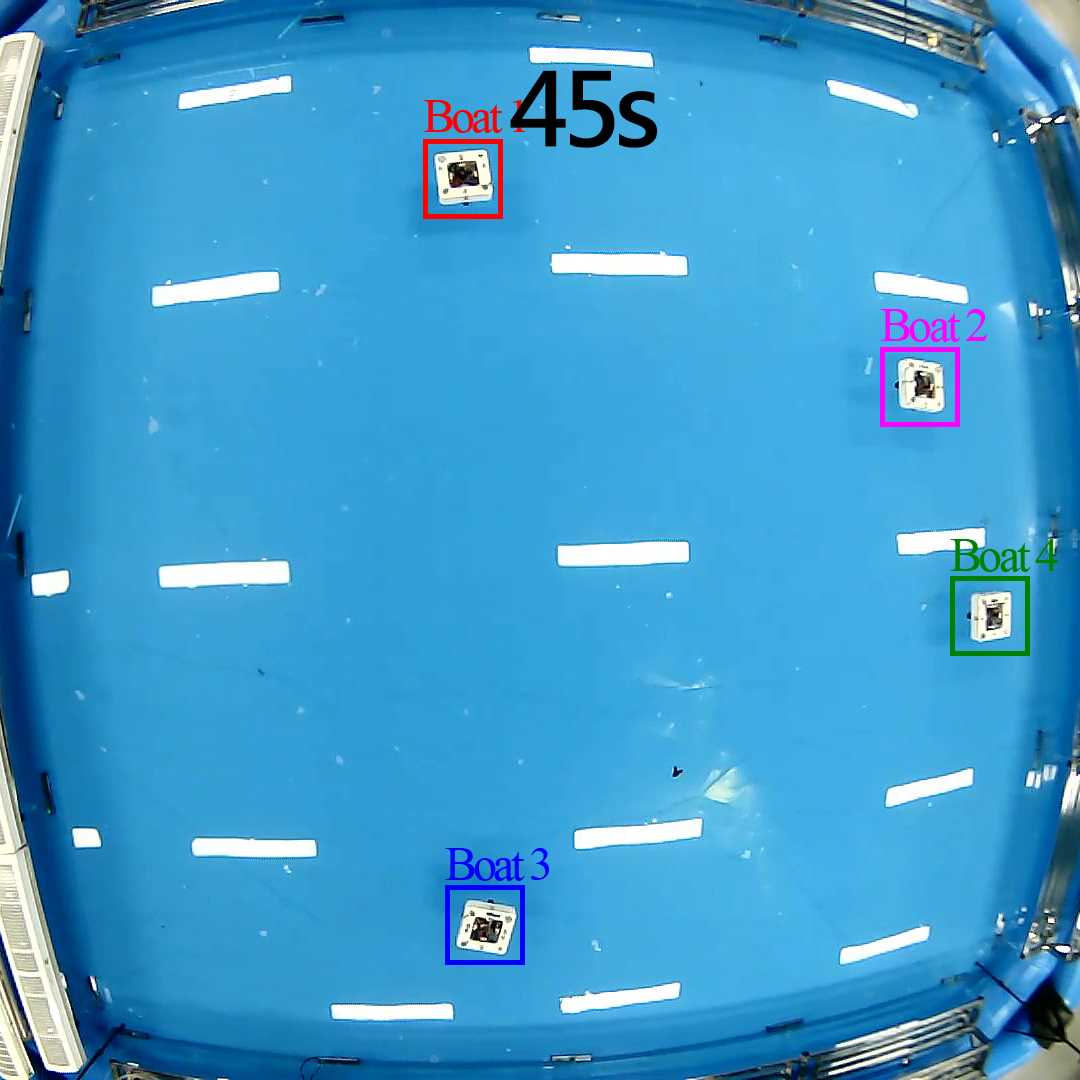}
			\put(-45,45){Moment}
			\put(-25,30){I}
			\put(20,105){No Obstacle}
		\end{overpic}
	}
	\subfloat[]{
		\centering
		\begin{overpic}[width=\figwidth\textwidth]{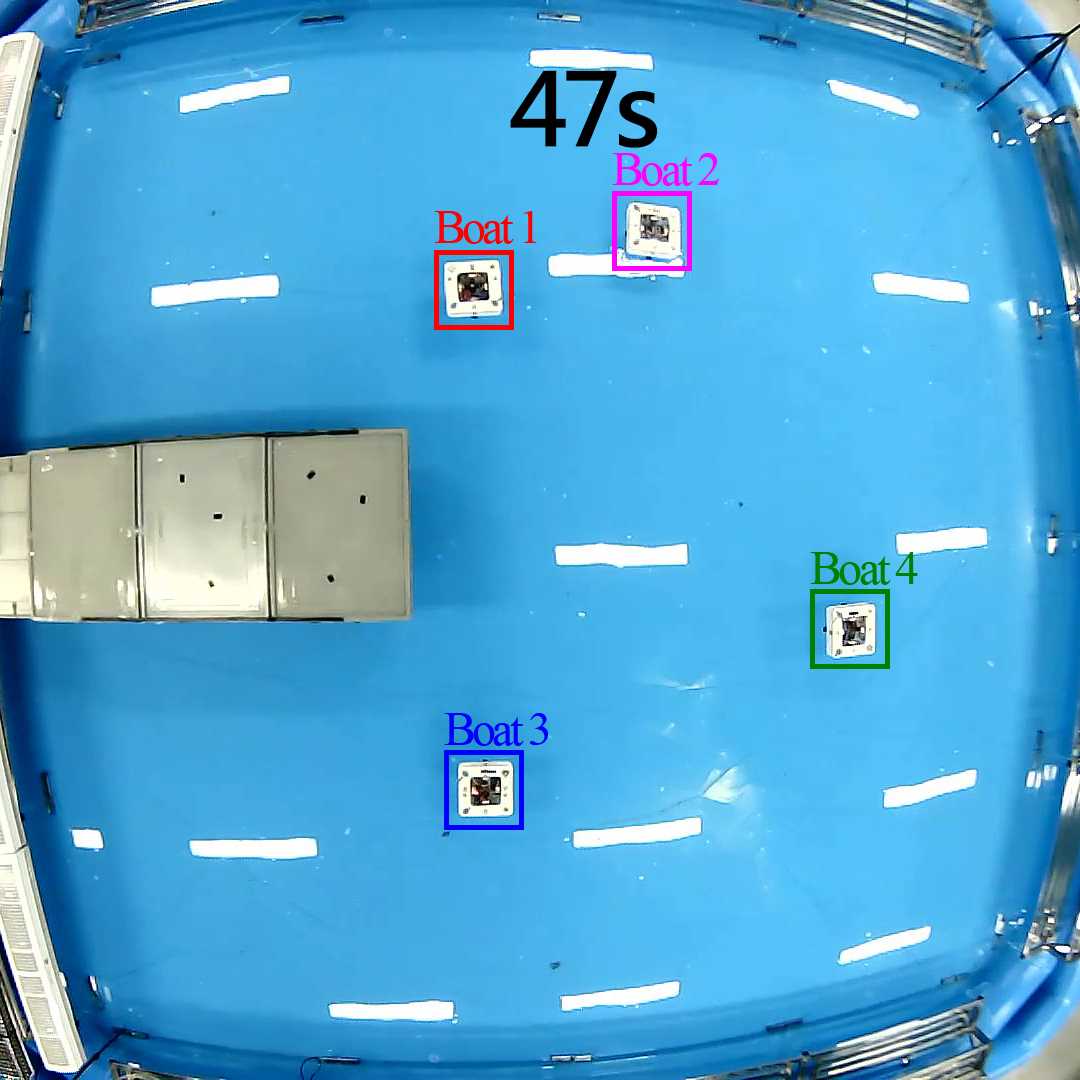}
			\put(10,105){1-side Obstacles}
		\end{overpic}
	}
	\subfloat[]{
		\centering
		\begin{overpic}[width=\figwidth\textwidth]{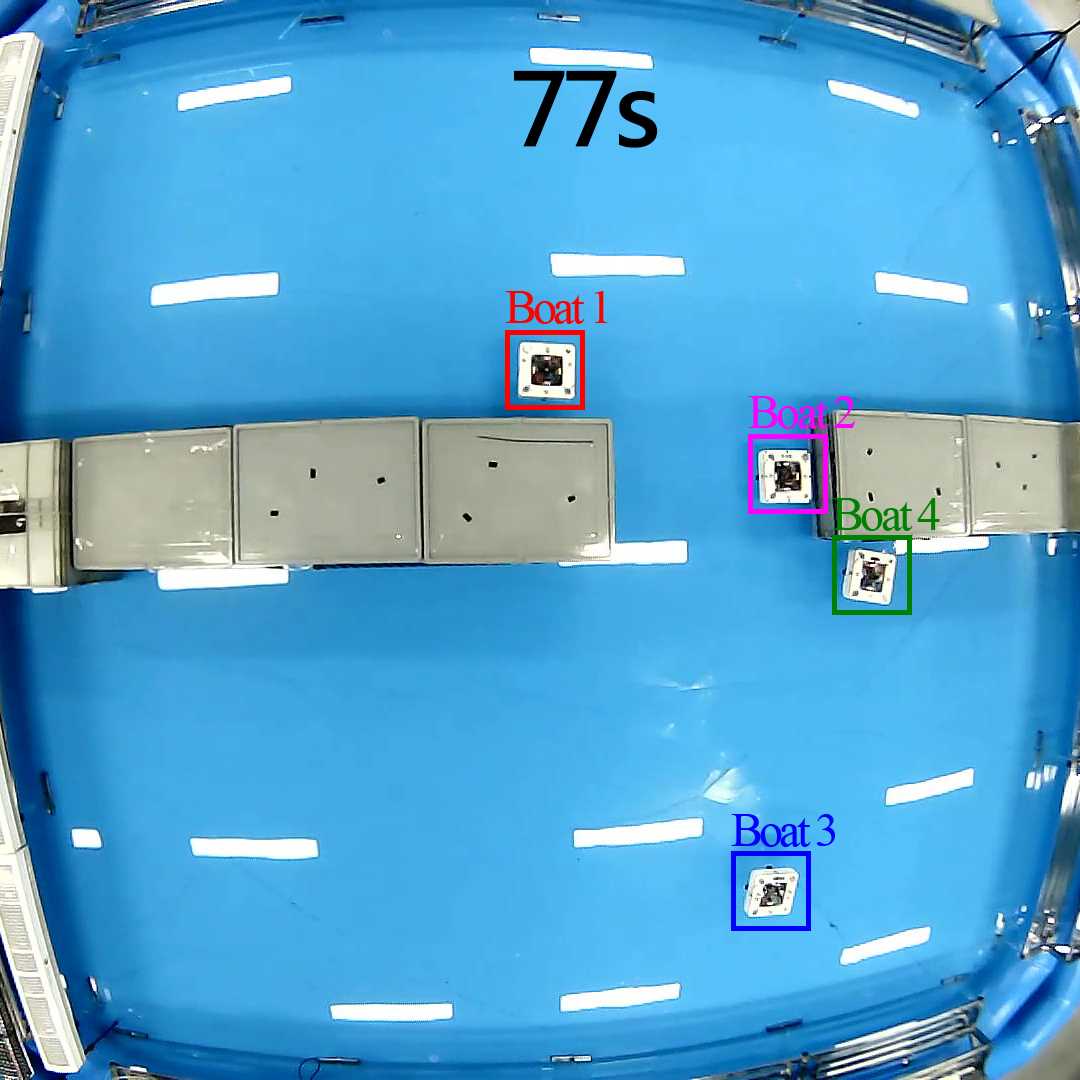}
			\put(10,105){2-side Obstacles}
		\end{overpic}
	}
	\subfloat[]{
		\centering
		\begin{overpic}[width=\figwidth\textwidth]{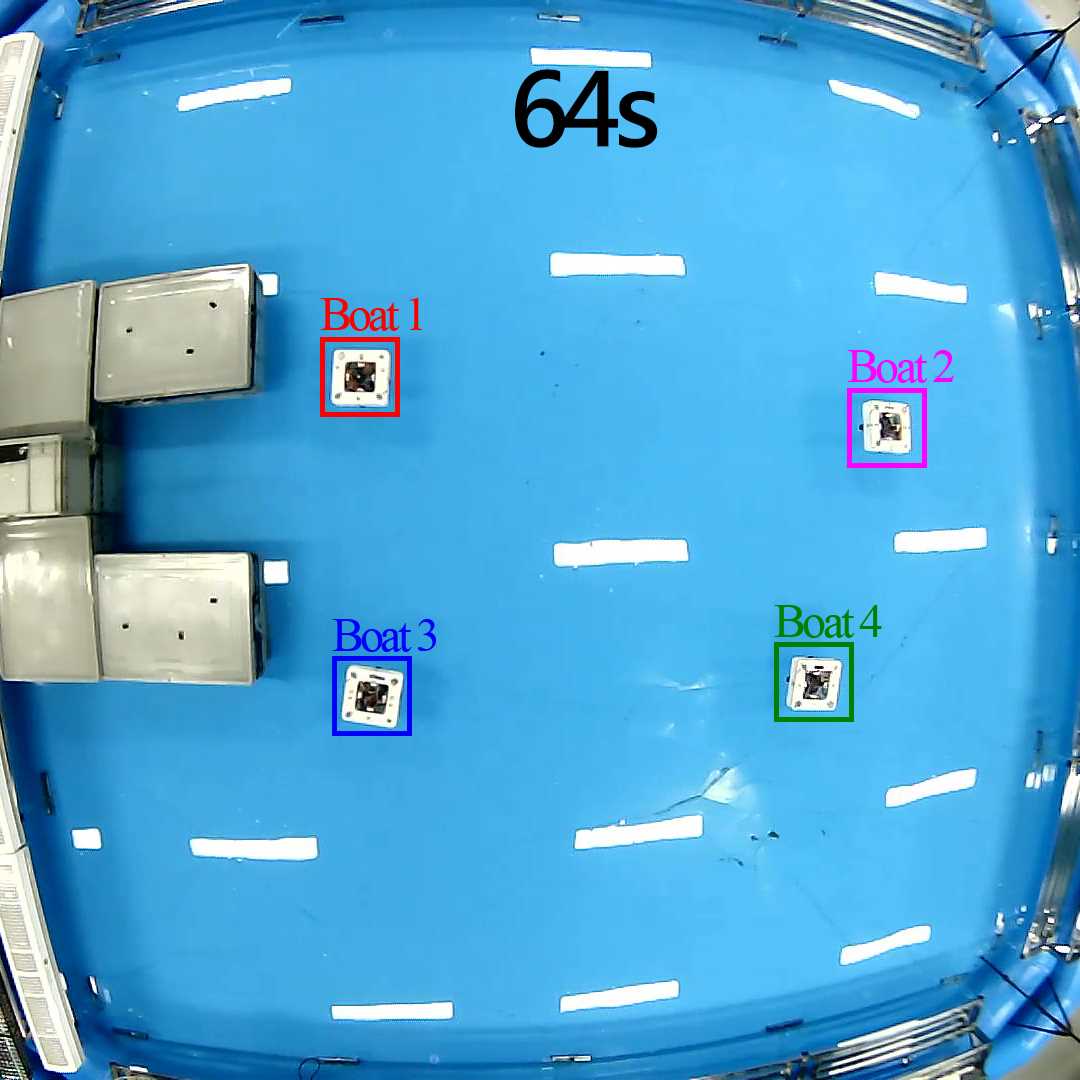}
			\put(10,105){3-side Obstacles}
		\end{overpic}
	}
	\subfloat[]{
		\centering
		\begin{overpic}[width=\figwidth\textwidth]{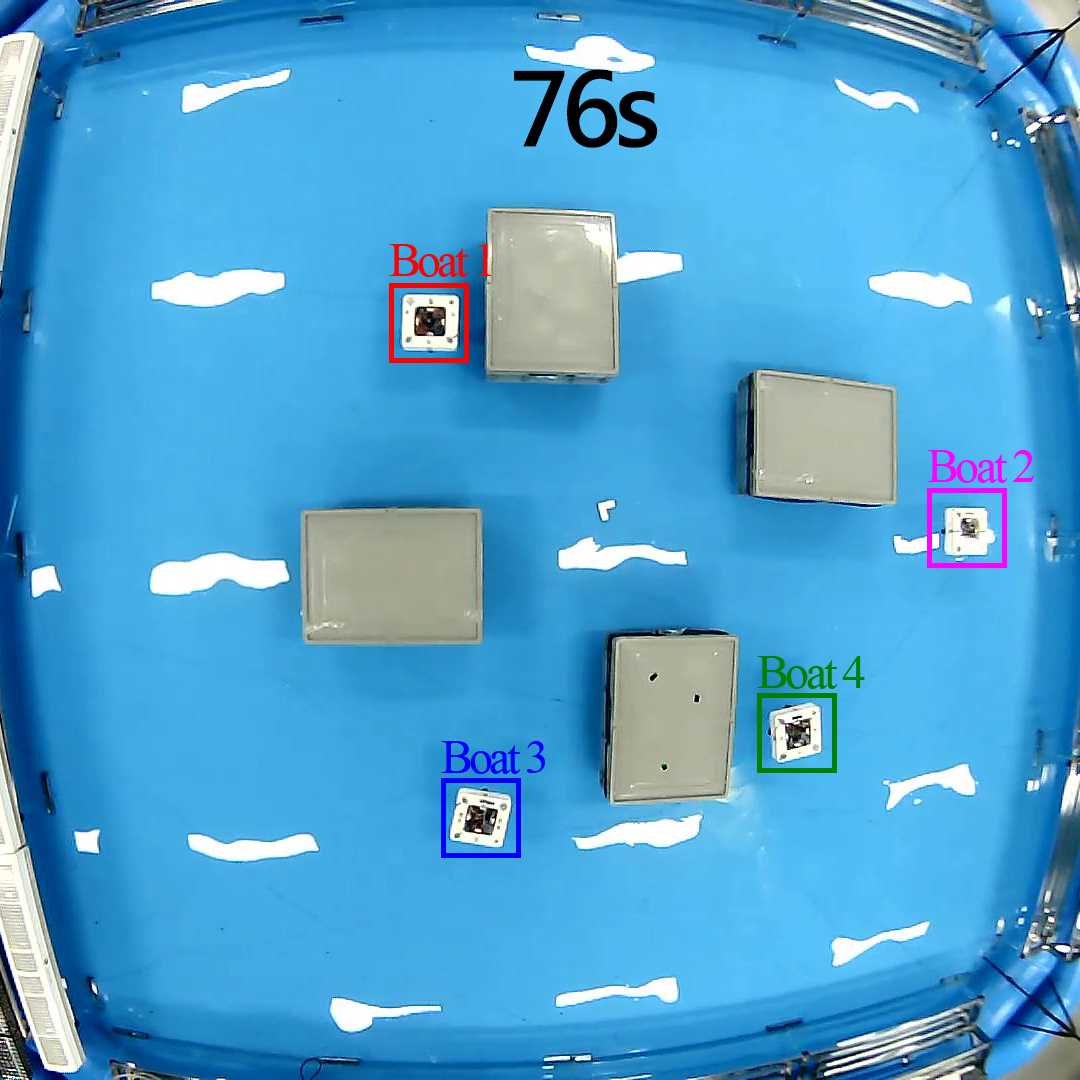}
			\put(10,105){4-side Obstacles}
		\end{overpic}
	}
	\hfill
	\subfloat[]{
		\centering
		\begin{overpic}[width=\figwidth\textwidth]{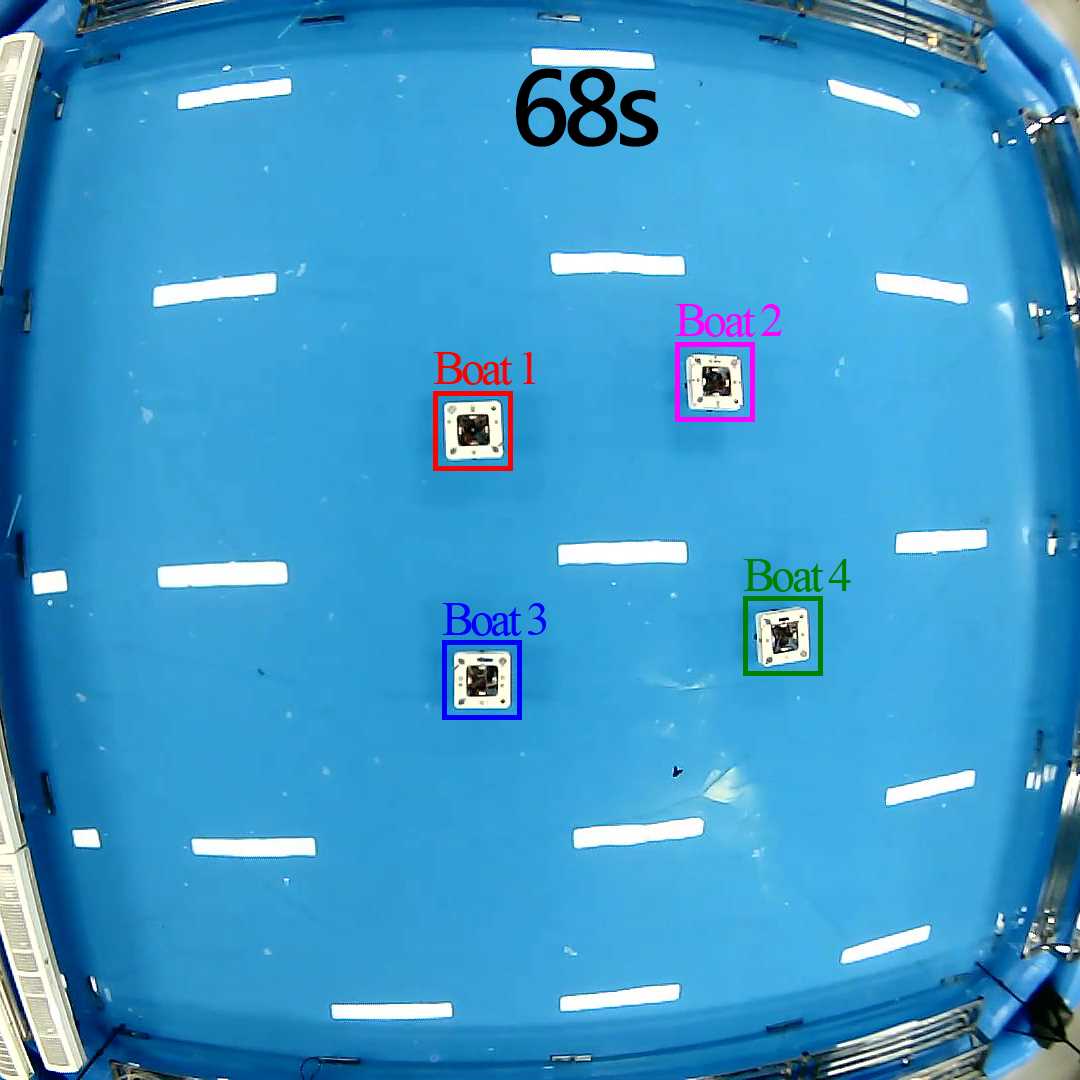}
			\put(-45,45){Moment}
			\put(-27,30){II}
		\end{overpic}
	}
	\subfloat[]{
		\centering
		\includegraphics[width=\figwidth\textwidth]{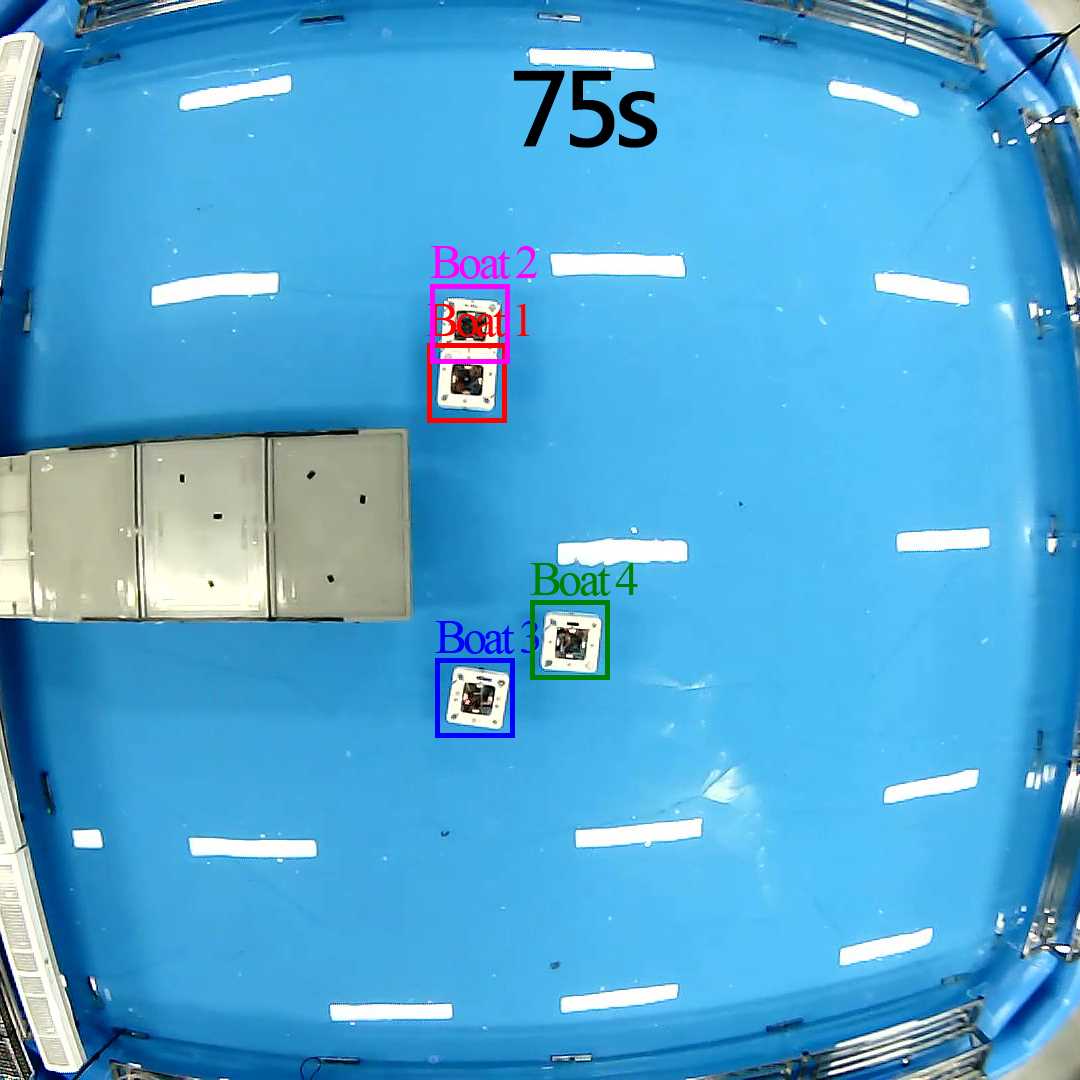}
	}
	\subfloat[]{
		\centering
		\includegraphics[width=\figwidth\textwidth]{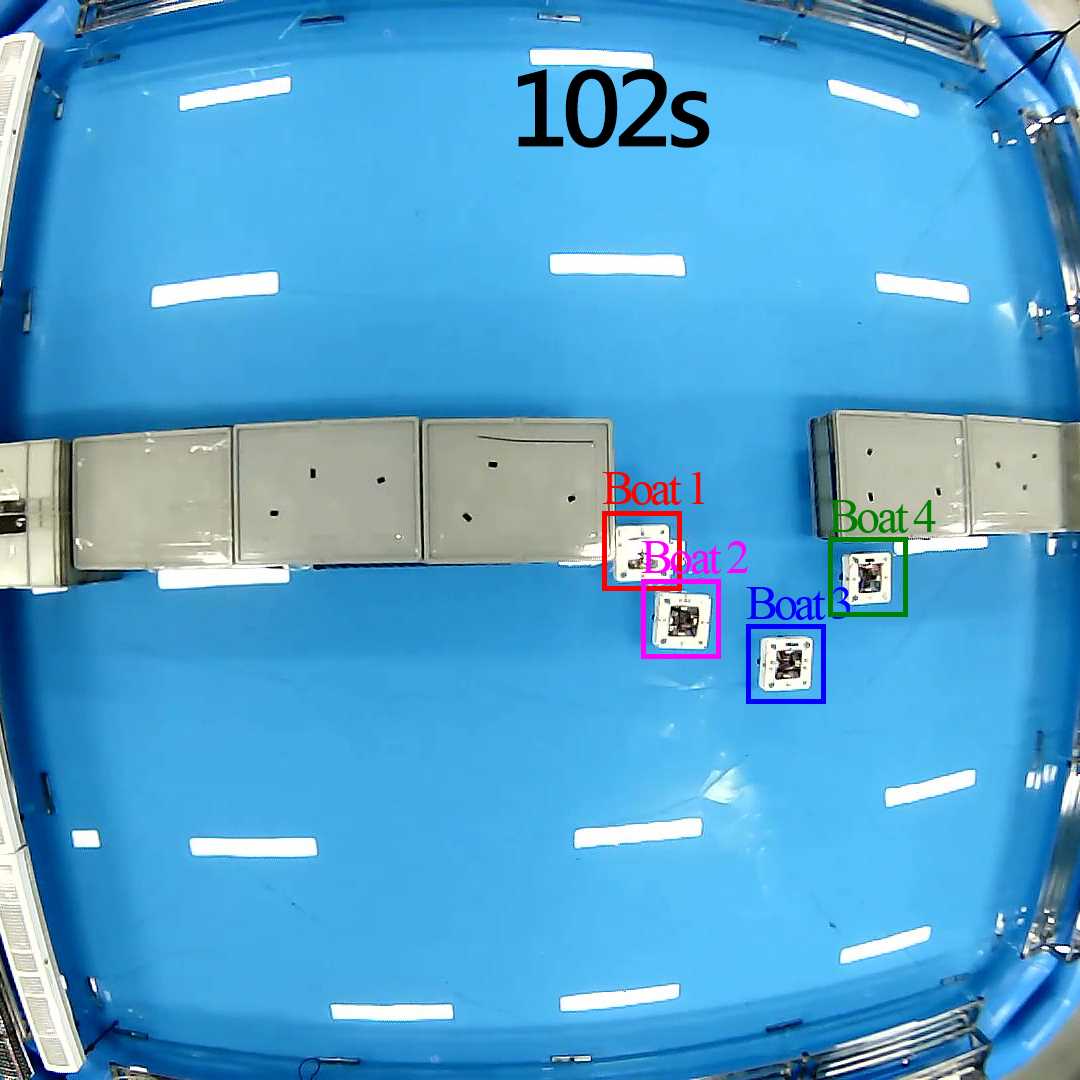}
	}
	\subfloat[]{
		\centering
		\includegraphics[width=\figwidth\textwidth]{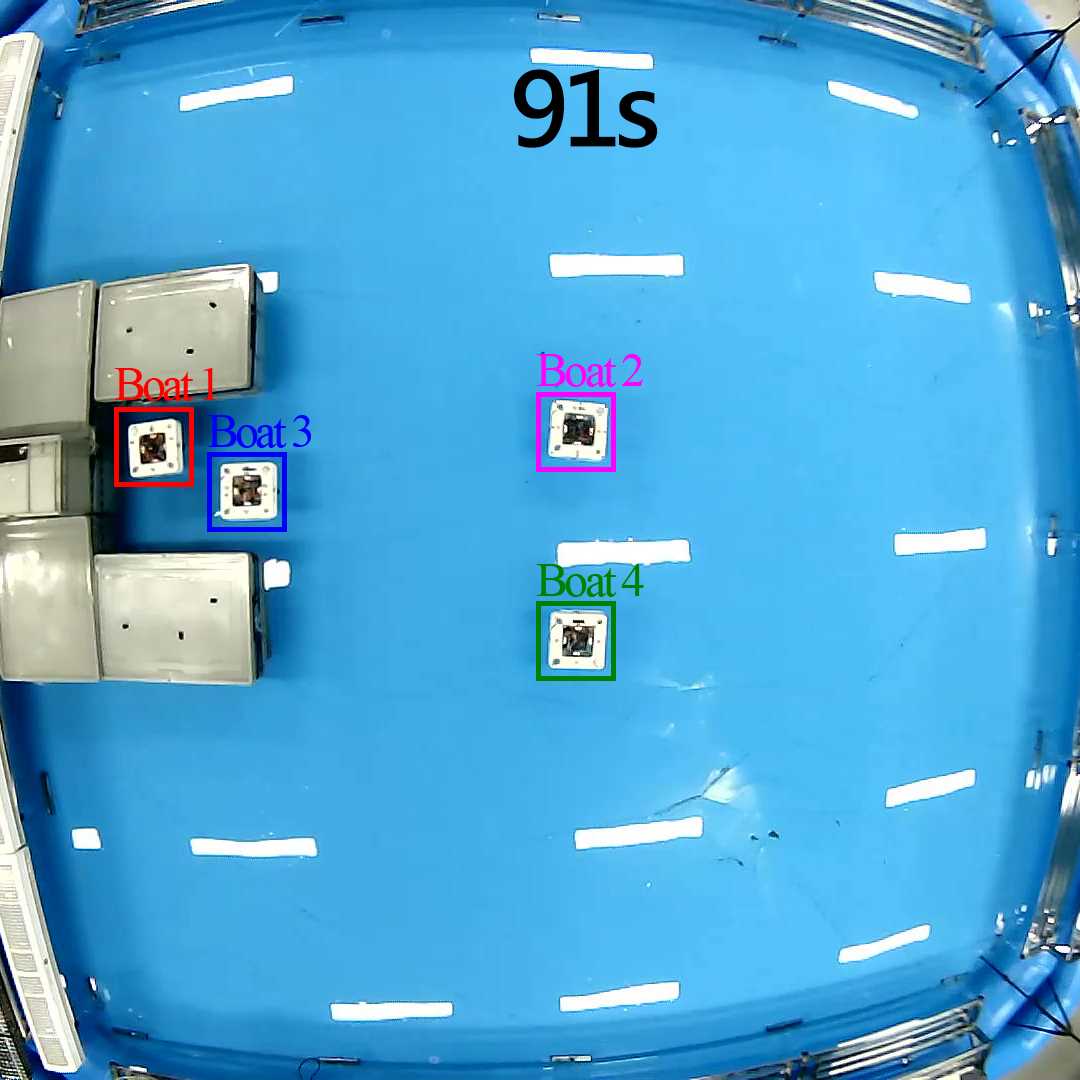}
	}
	\subfloat[]{
		\centering
		\includegraphics[width=\figwidth\textwidth]{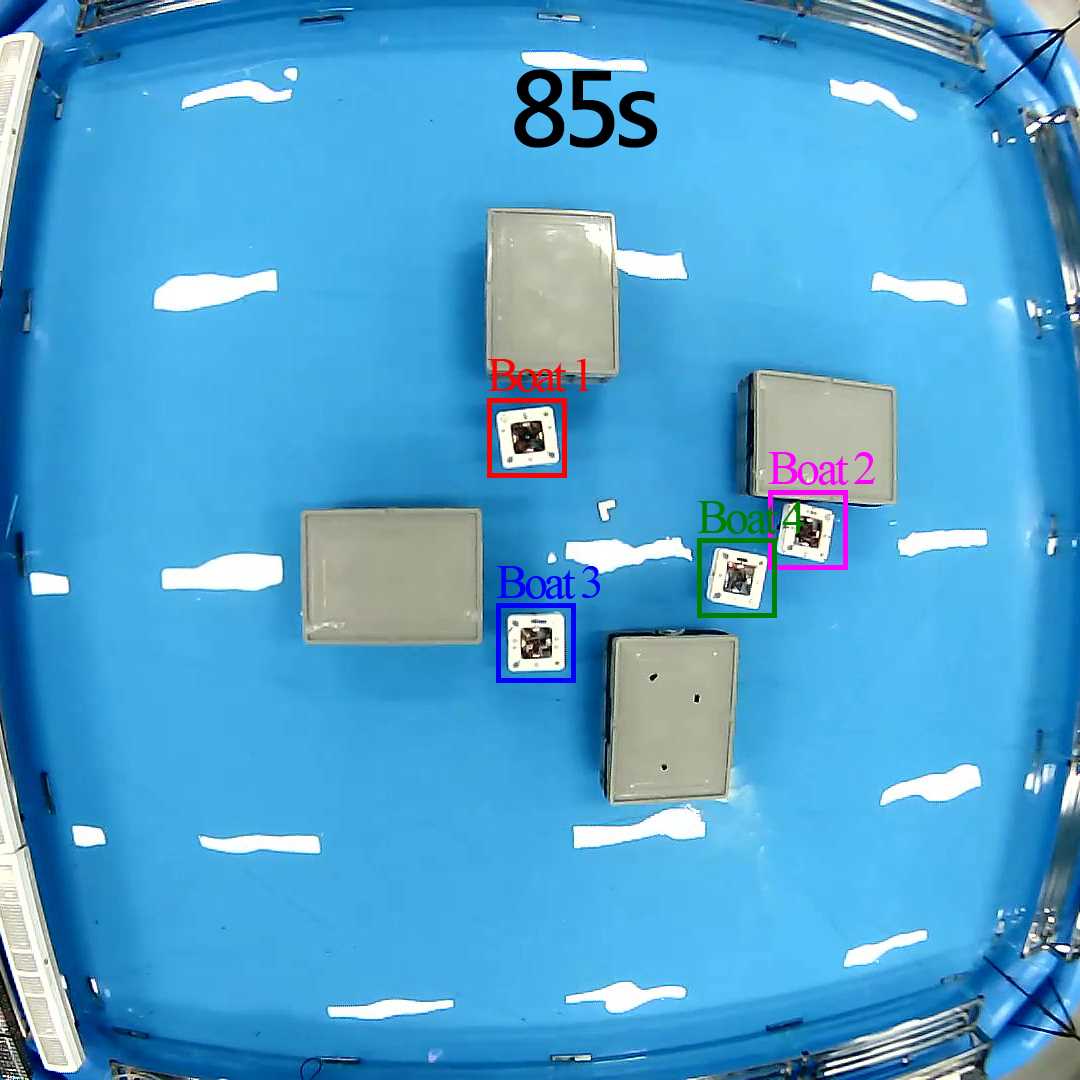}
	}
	\hfill
	\subfloat[]{
		\centering
		\begin{overpic}[width=\figwidth\textwidth]{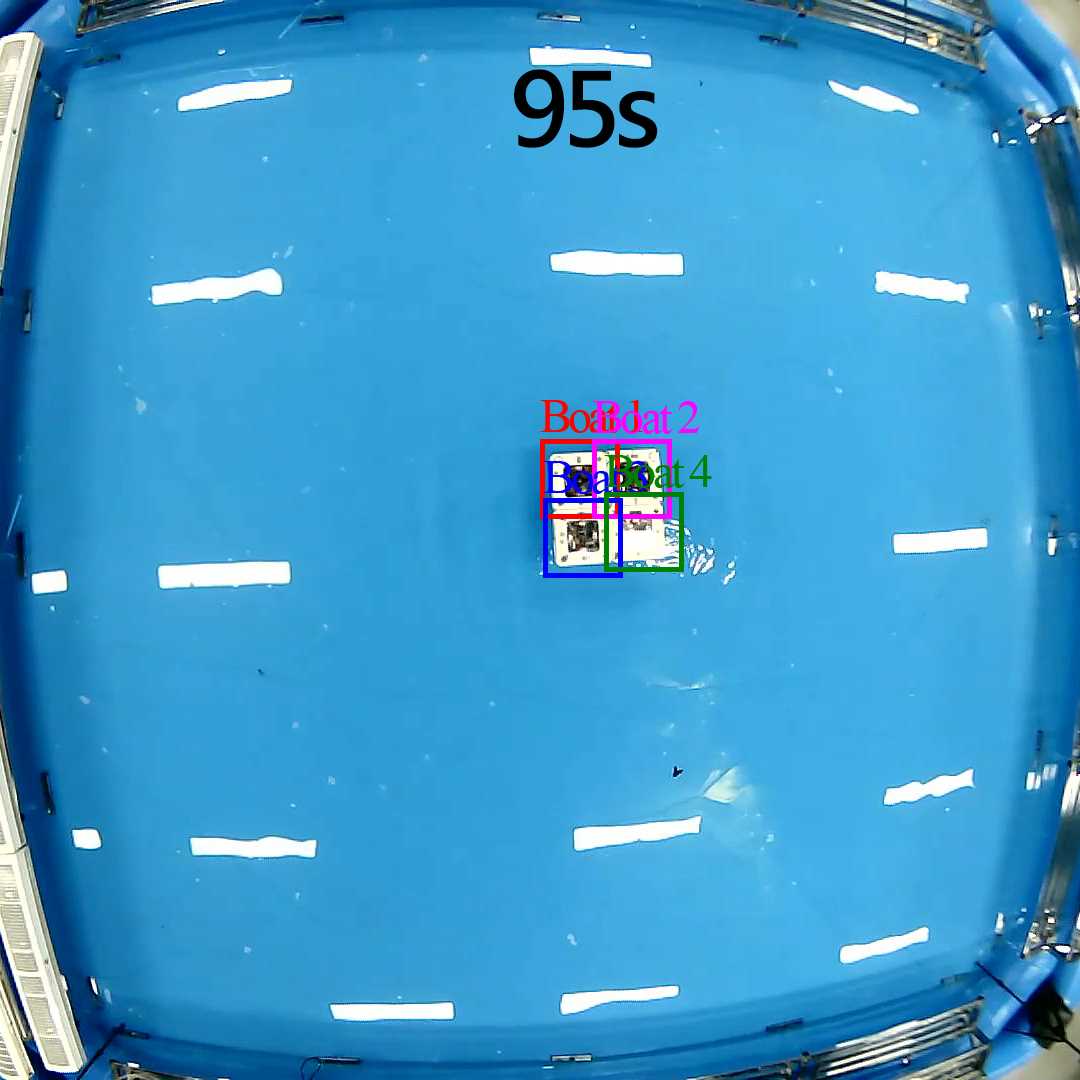}
			\put(-45,45){Moment}
			\put(-30,30){III}
		\end{overpic}
	}
	\subfloat[]{
		\centering
		\includegraphics[width=\figwidth\textwidth]{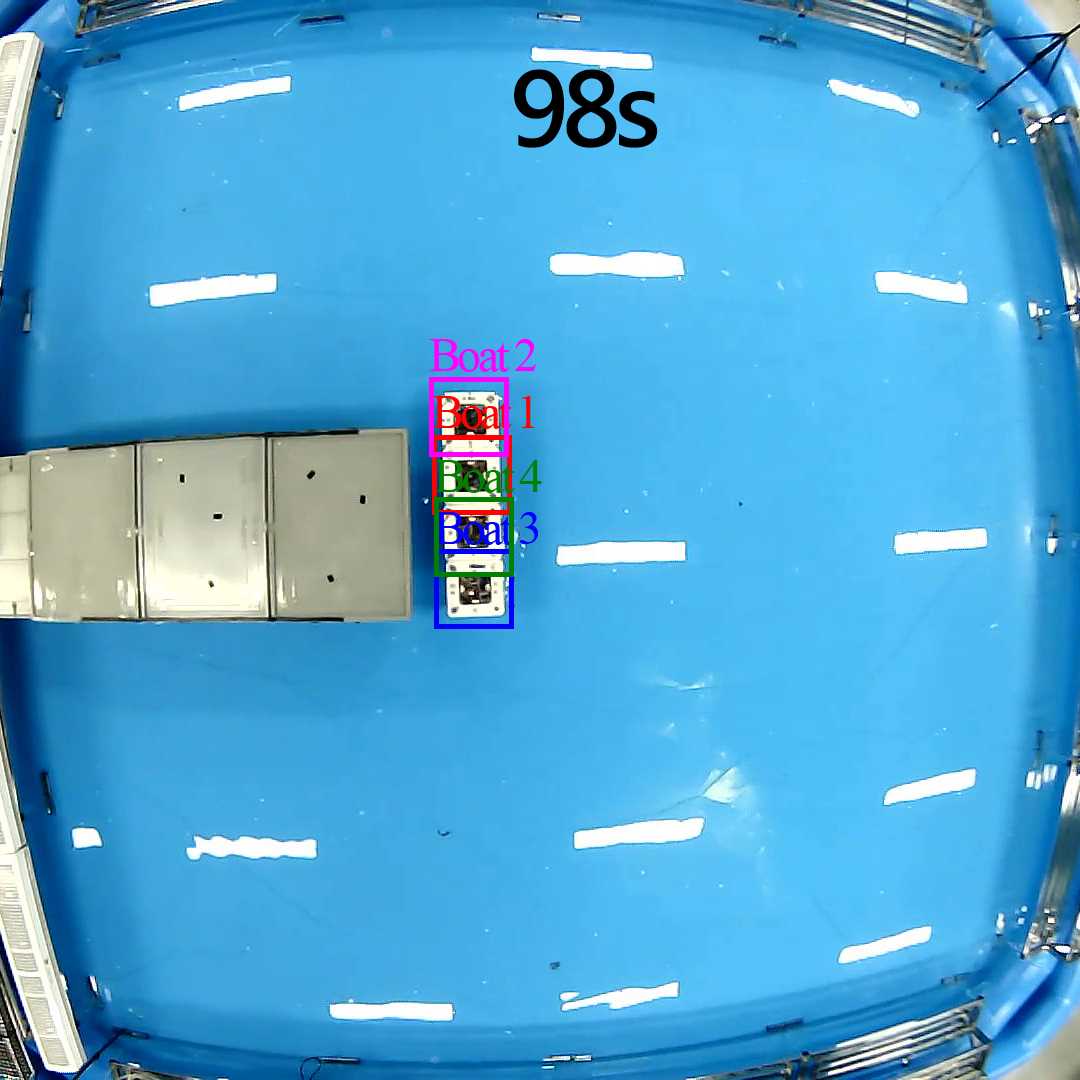}
	}
	\subfloat[]{
		\centering
		\includegraphics[width=\figwidth\textwidth]{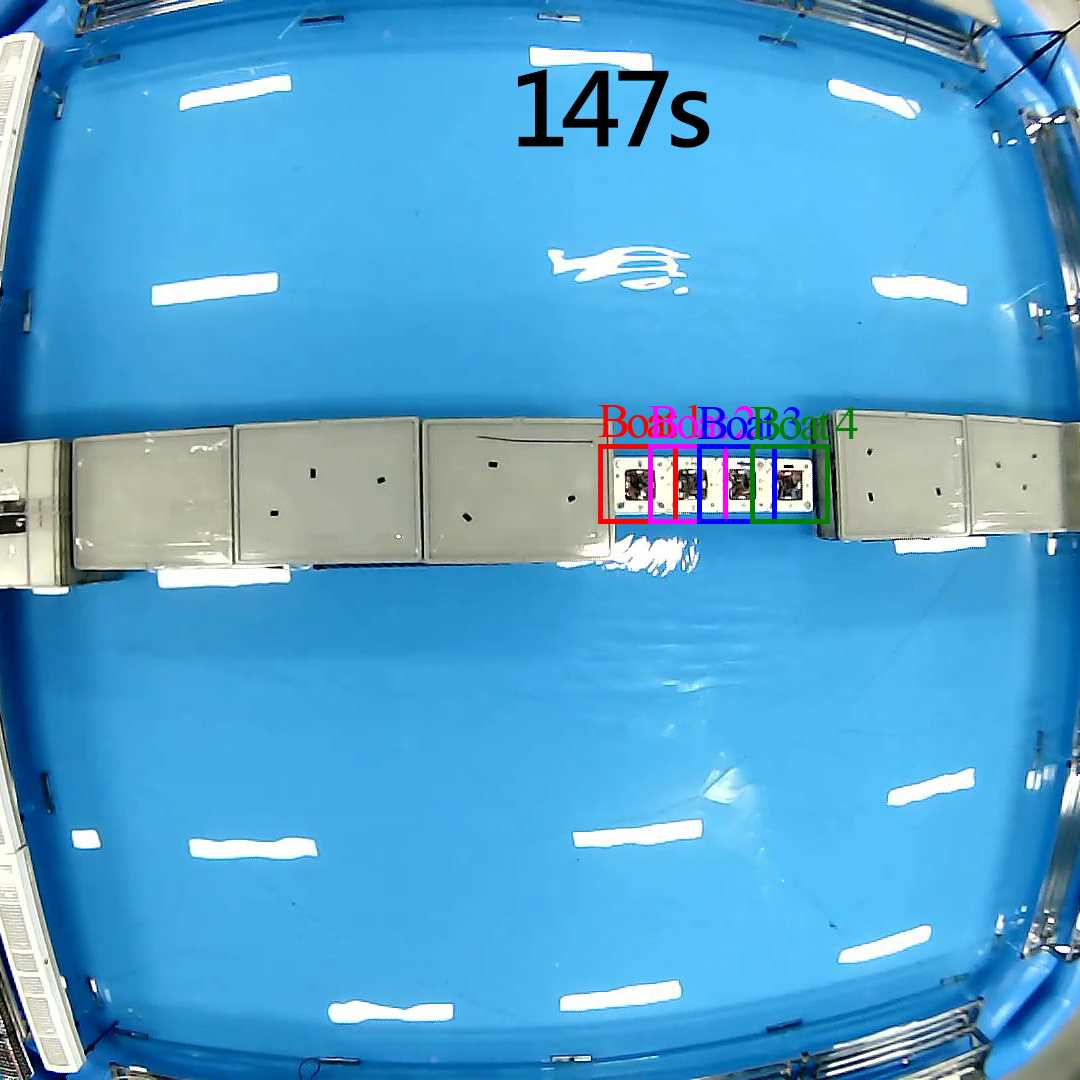}
	}
	\subfloat[]{
		\centering
		\includegraphics[width=\figwidth\textwidth]{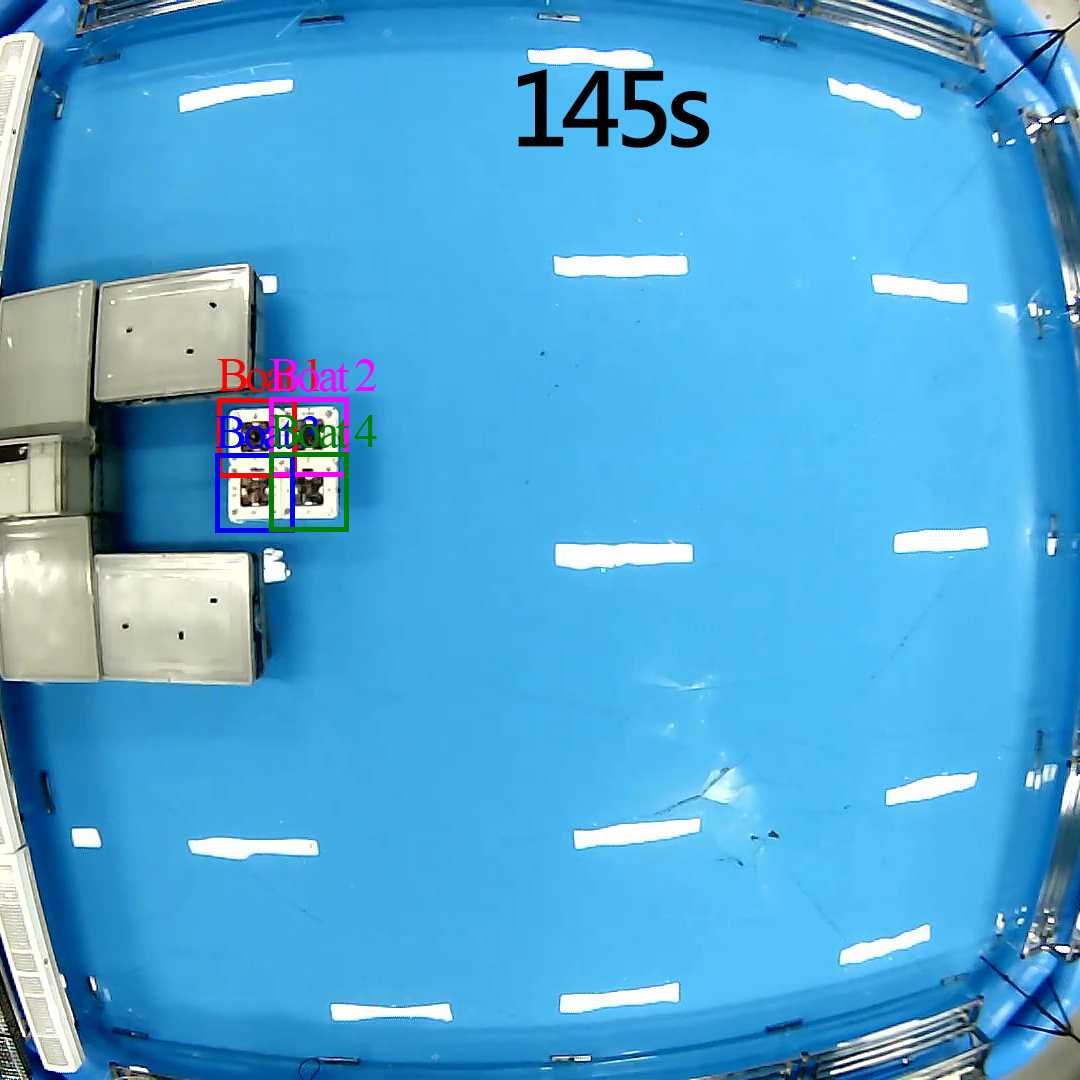}
	}
	\subfloat[]{
		\centering
		\includegraphics[width=\figwidth\textwidth]{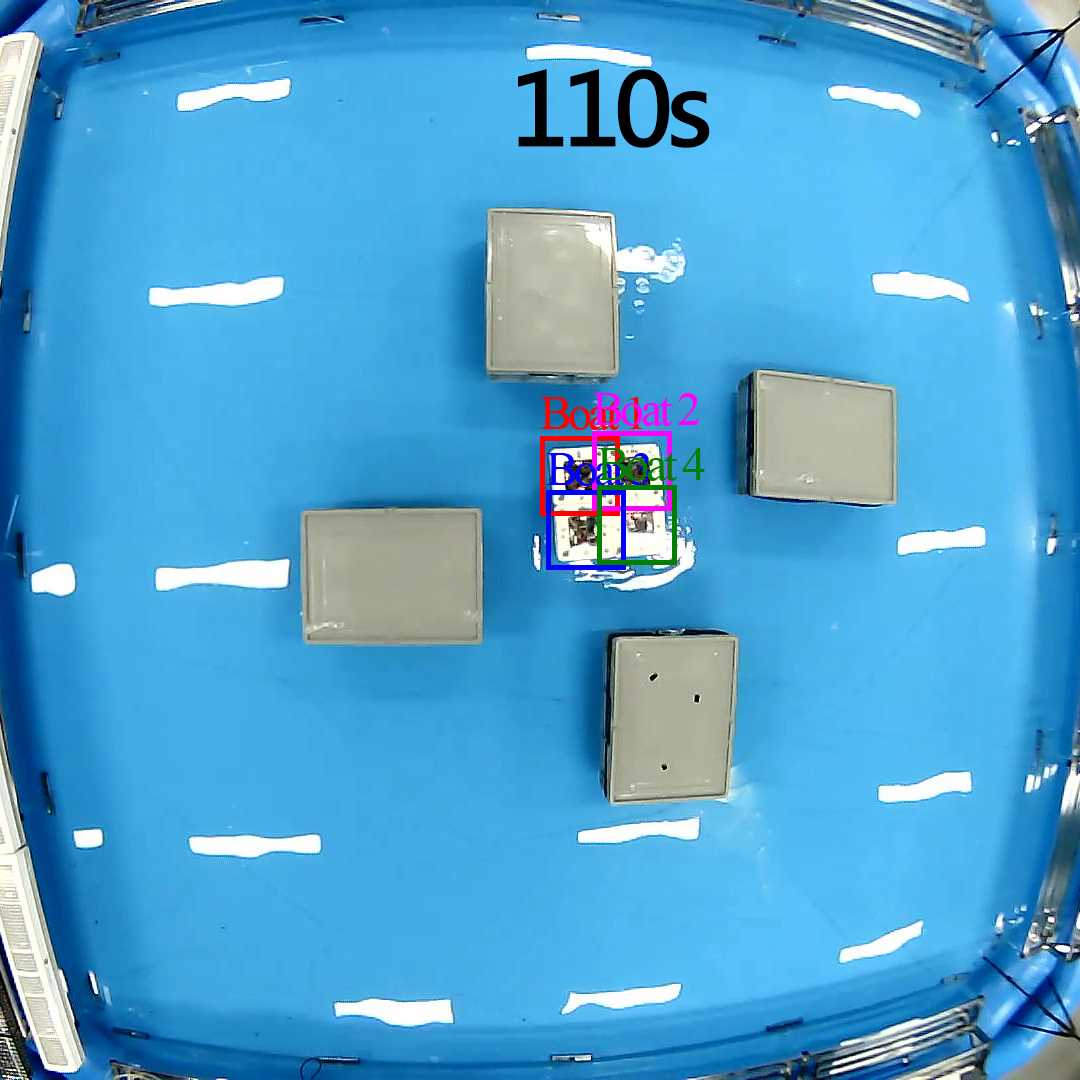}
	}
	\hfill
	\caption{Top view of the self-assembly process. Three moments are selected: moving, approaching, and assembly completion.}
	\vspace{-10pt}
	\label{fig:bridging}
\end{figure*}

\subsection{Self-assembly}
We conducted the self-assembly experiments of the CuBoats in five maps with different obstacle configurations. 
According to the classification in Section \ref{sect:evaluation}, maps 1-5 corresponding to categories 1-5 (from no obstacle to 4-direction obstacles) present a single robot platform, a long dock, a broken bridge, a boat dock, and four reefs, as shown in Fig. \ref{fig:exp_path}.
All desired structures are successfully assembled as expected.

Fig. \ref{fig:exp_path} (a) depicts the experimental setup.
The upper graphs of Fig. \ref{fig:exp_path} (b)-(f) portray the trajectories of all CuBoats.
It is possible to see how the robots bypass the obstacles and assemble,
during which the CuBoats first dock in pairs, and then assemble into the target structures.
The lower graphs of Figures \ref{fig:exp_path} (b)-(f) plot the evolution of the completed connections versus time.
We can see that the CuBoats consume a roughly increasing time from Cat. 1 to 5,
since the difficulty of the maps grows with the number of directions where the obstacles appear.
The completion times in sequence are $94 \, \text{s}$, $97 \, \text{s}$, $146 \, \text{s}$, $144 \, \text{s}$, and $110 \, \text{s}$.
As a consequence of applying the SAPOA algorithm, the CuBoats can successfully assemble without collision with obstacles and mistaken docking.

Figures \ref{fig:exp_error} (a)-(e) show the following errors $d$ and the lateral errors $d\cos \gamma$ in equation (\ref{eq:traj_generator}) for all experiments, which echo the assumptions in the navigation system.
First, all the following errors $d$ are reduced to nearly 0 in every target-update cycle, i.e., the CuBoats can move one grid. 
Second, during the movement, the lateral errors hardly exceed half a grid width ($0.125 \, \text{m}$), i.e., the boats occupy the desired grids.
The overall processes of the five experiments are shown in Fig. \ref{fig:bridging}.

\section{Conclusion}
\label{sect:conclusion}

In this paper, we present an SAP algorithm for modular robots, which expands the parallel self-assembly process to environments with obstacles. 
This algorithm has been extensively described in our previous work \cite{zhang2021efficient},
which contains four stages: (1) determining the (dis)assembly sequence by an assembly tree, (2) expanding the target locations and recording the landmarks, (3) dispatching robots to targets, and (4) navigating all robots and assembling in parallel.
A series of comparative simulations indicate that our algorithm as well as its variants have a higher success rate ($\geq 80\%$) than the existing methods when running in an environment with obstacles.
To validate the practicality, a multi-USV hardware testbed system is proposed including four omnidirectional USVs, named CuBoat, with active docking systems.
The CuBoats can execute the high-level self-assembly plan by maneuvering on the water surface and forming the desired structures.
The hardware experiments of 4 robots on 5 maps with different obstacles verify the effectiveness and generality.

Our algorithm and multi-USV system offer numerous potential applications, including the provision of temporary bridges, harbors, or airstrips for water surface monitoring and exploration.
In the future, water fluctuations and more robots can be involved in the experiments to investigate the effect of disturbances (waves, tidal influence, wind, etc.) and robot numbers.
Besides, conducting theoretical proofs for the correctness of the algorithm is one of the future research directions.


\bibliographystyle{IEEEtran}
\bibliography{selfassembly.bib}

%
\vspace{-10 mm} 
\begin{IEEEbiography}[{\includegraphics[width=1in,height=1.25in,clip,keepaspectratio]{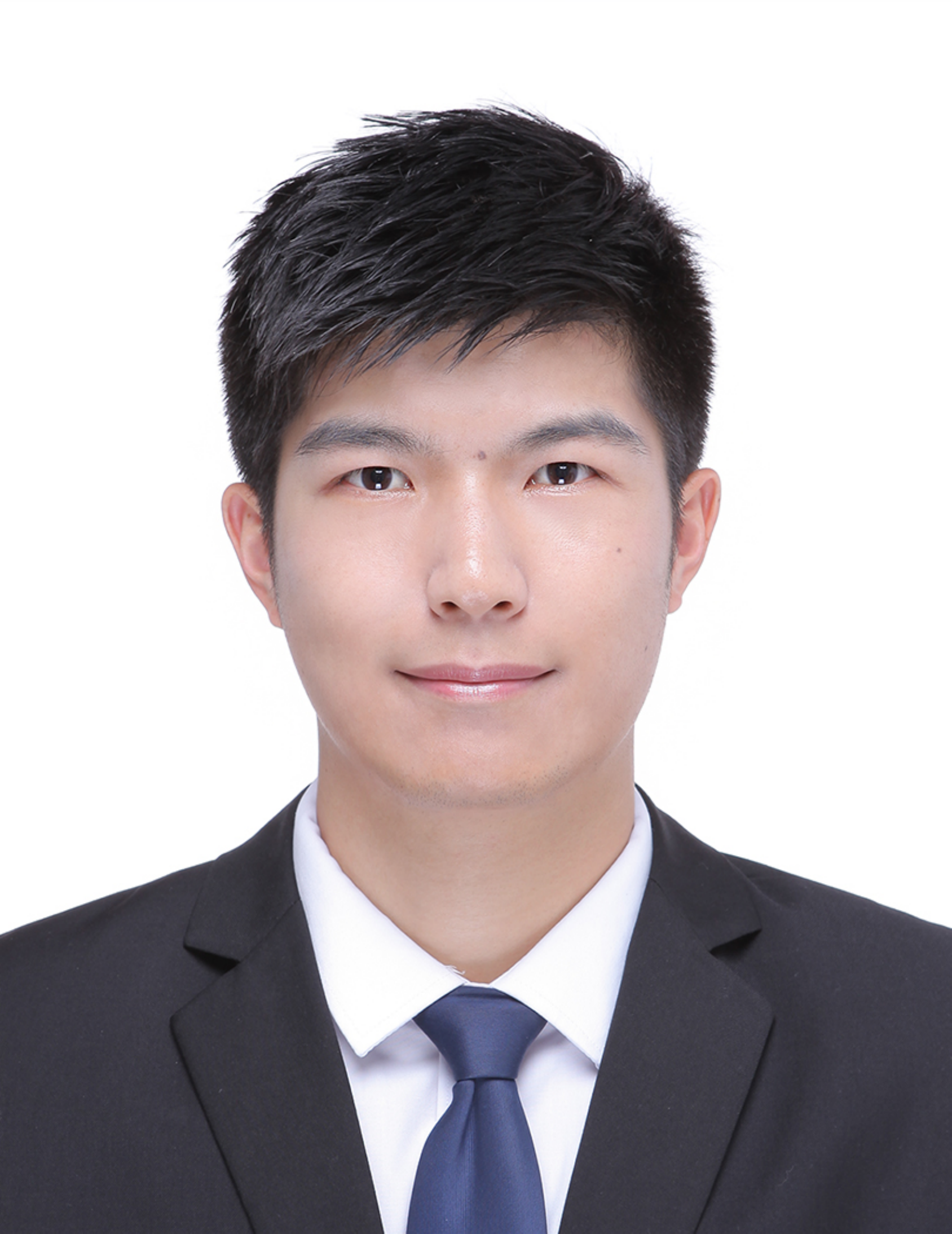}}]{Lianxin Zhang}
	received the B.E. degree in mechanical engineering and automation from Nanjing University of Science and Technology, Nanjing, China, in 2015, and the M.S. degree in mechanics from Tongji University, Shanghai, China, in 2018. He is currently pursuing the Ph.D. degree at The Chinese University of Hong Kong, Shenzhen, Guangdong, China, where he specializes in the design and control of novel unmanned surface vehicles.
\end{IEEEbiography}
\vspace{-10 mm} 
\begin{IEEEbiography}[{\includegraphics[width=1in,height=1.25in,clip,keepaspectratio]{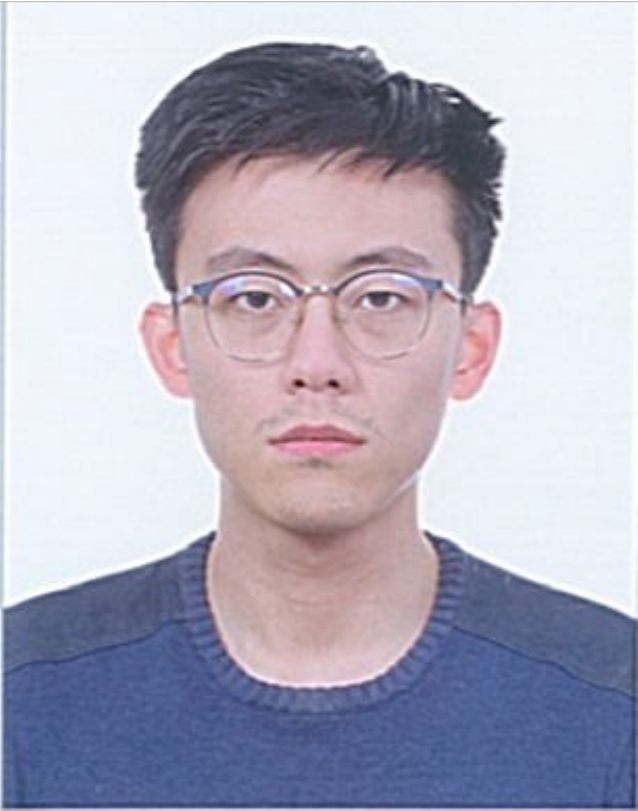}}]{Yihan Huang}
	received the B.E. degree in electronic information engineering from the Chinese University of Hong Kong, Shenzhen, China, in 2021. 
	He is now working in the Robotics and Artificial Intelligence Laboratory of the Chinese University of Hong Kong, Shenzhen, China. 
\end{IEEEbiography}
\vspace{-10 mm}
\begin{IEEEbiography}[{\includegraphics[width=1in,height=1.25in,clip,keepaspectratio]{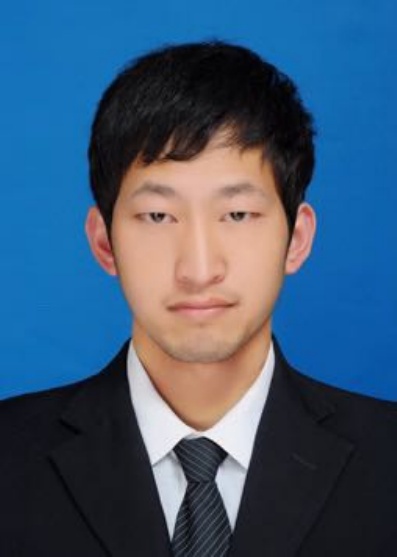}}]{Zhongzhang Cao}
	received his B.S.degree from Xi’an Jiaotong University, Xian, China. He currently works at the Robotics and Artificial Intelligence Laboratory of the Chinese University of Hong Kong (Shenzhen). His main research interests are structural design and optimization analysis of robots.
\end{IEEEbiography}
\vspace{-10 mm} 
\begin{IEEEbiography}[{\includegraphics[width=1in,height=1.25in,clip,keepaspectratio]{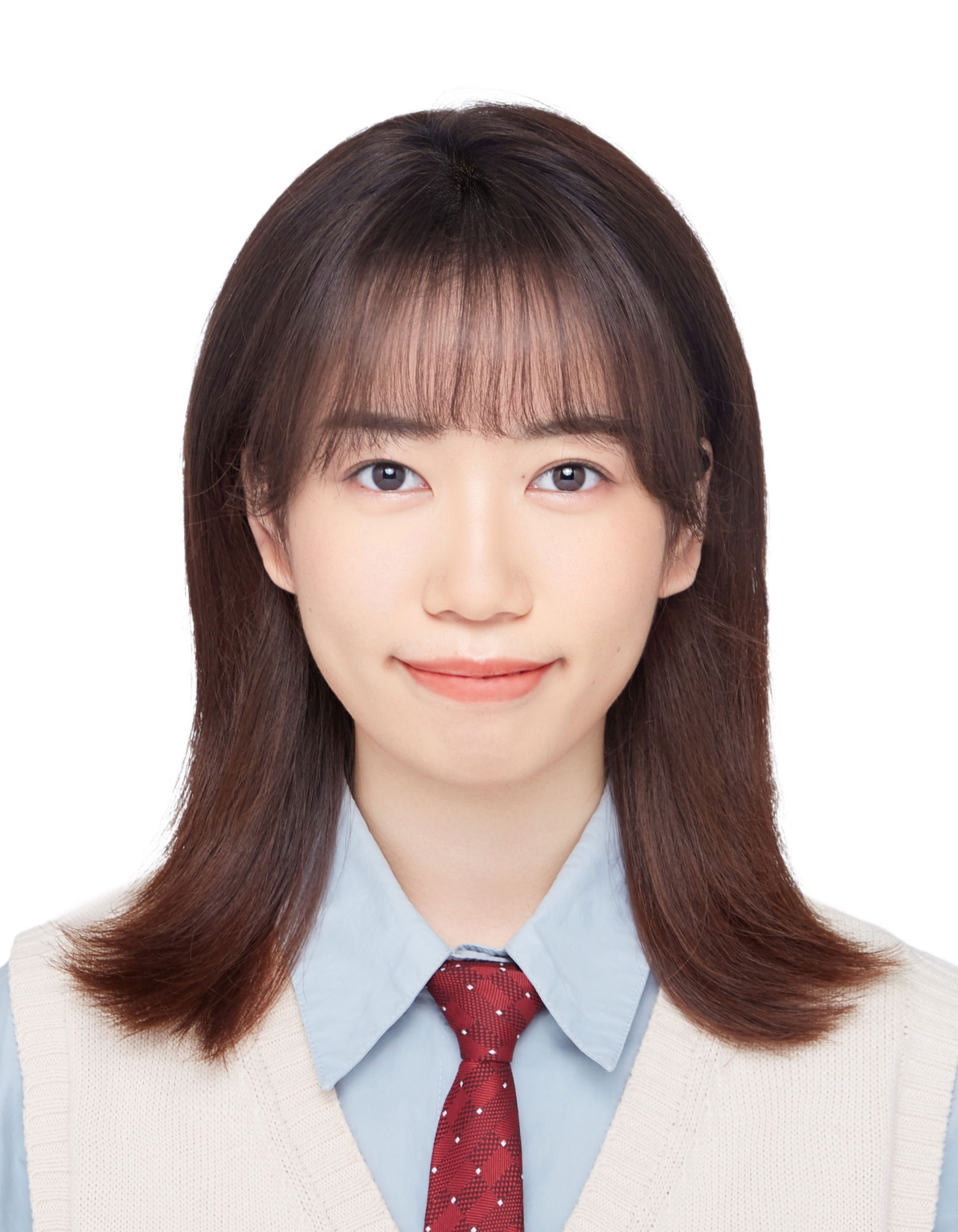}}]{Yang Jiao}
	received the B.E. degree in electronic information engineering from The Chinese University of Hong Kong, Shenzhen, Guangdong, China, in 2022. 
	She is currently pursuing the M.S. degree in intelligent systems, robotics \& control with the Department of Electrical and Computer Engineering, University of California San Diego, San Diego, CA, USA.
\end{IEEEbiography}
\vspace{-10 mm} 
\begin{IEEEbiography}[{\includegraphics[width=1in,height=1.25in,clip,keepaspectratio]{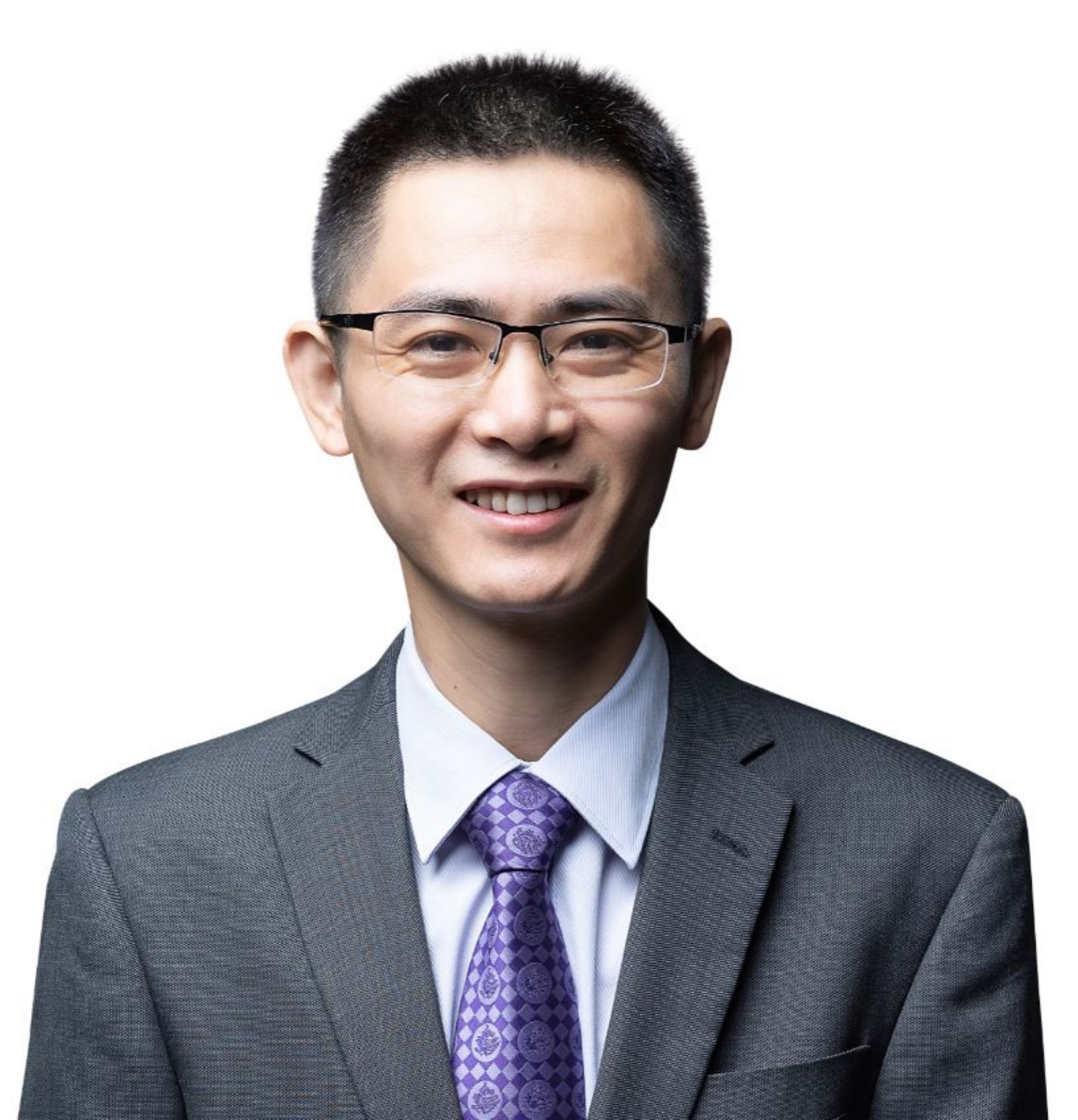}}]{Huihuan Qian}
	(Member, IEEE) received the B.E. degree from the Department of Automation, University of Science and Technology of China, Hefei, China, in 2004, and the Ph.D. degree from the Department of Mechanical and Automation Engineering, The Chinese University of Hong Kong, Hong Kong, SAR, China, in 2010. He is currently an Assistant Professor with the School of Science and Engineering, The Chinese University of Hong Kong, Shenzhen, Guangdong, China, and the Associate Director of the Shenzhen Institute of Artificial Intelligence and Robotics for Society, Shenzhen. His current research interests include robotics and intelligent systems, especially in marine environment.
\end{IEEEbiography}

\end{document}